\documentclass{article}

\usepackage[final]{neurips_2024}

\usepackage[utf8]{inputenc} % allow utf-8 input
\usepackage[T1]{fontenc}    % use 8-bit T1 fonts
\usepackage{hyperref}       % hyperlinks
\usepackage{url}            % simple URL typesetting
\usepackage{booktabs}       % professional-quality tables
\usepackage{amsfonts}       % blackboard math symbols
\usepackage{nicefrac}       % compact symbols for 1/2, etc.
\usepackage{microtype}      % microtypography
\usepackage{xcolor}         % colors
\usepackage{bm}

\usepackage{amsmath}
\usepackage{algorithm}

\usepackage{algorithmic}

\usepackage{pifont}
\usepackage{makecell}
\usepackage{booktabs}
\usepackage{multicol}
\usepackage{bbm}
\usepackage{graphicx}
\usepackage{multirow}
\usepackage{caption}
\usepackage{subcaption}
\usepackage{xcolor}
\definecolor{morandi1}{RGB}{100, 130, 180}   % 柔和蓝（Step 1）
\definecolor{morandi2}{RGB}{200, 145, 100}   % 柔和橙（Step 2）
\definecolor{morandi3}{RGB}{115, 160, 115}   % 柔和绿（Step 3）
\title{Optimizing Distributional Geometry Alignment with Optimal Transport for Generative Dataset Distillation}

\author{Xiao Cui$^{1,2}$~~~ Yulei Qin$^3$~~~ Wengang Zhou$^1$ ~~~ Hongsheng Li$^{2}$~~~ Houqiang Li$^1$  \\
$^1$ University of Science and Technology of China 
\\ $^2$ CUHK MMLab~~~~$^3$ %Tencent Youtu Lab \\
Independent Researcher \\
{\tt\small cuixiao2001@mail.ustc.edu.cn, 
yuleichin@126.com} \\ 
{\tt\small \{zhwg,lihq\}@ustc.edu.cn, hsli@ee.cuhk.edu.hk}}

\begin{document}

\maketitle
\makeatletter{\renewcommand*{\@makefnmark}{}
\footnotetext{Corresponding authors: Wengang Zhou and Houqiang Li.}\makeatother}

\begin{abstract}
%Dataset distillation aims to generate a compact distilled dataset from a large-scale real dataset such that models trained on the distilled data perform comparably to those trained on the full data. %While recent generative and model inversion methods improve scalability, they often rely solely on global statistics (e.g., mean and variance), neglecting the instance-level and geometric structure within the real data distribution—leading to suboptimal alignment and limited generalization. 
% Existing scalable approaches typically focus only on matching global distributional statistics (e.g., mean and variance), but neglect crucial instance-level characteristics and intra-class variations inherent within the real data distribution. 
% Dataset distillation aims to synthesize a small distilled dataset from the real dataset, enabling models trained on it to perform comparably to those trained on the original data.
% OT offers a geometrically faithful framework for distribution matching, making it particularly effective at capturing subclass structures and local variations that are often weakened via the averaging of intraclass objectives.
% Existing large-scale approaches typically 
% Dataset distillation seeks to synthesize a compact distilled dataset that enables student models to achieve performance comparable to training on the full real dataset.
Dataset distillation seeks to synthesize a compact distilled dataset, enabling models trained on it to achieve performance comparable to models trained on the full dataset. Recent methods for large-scale datasets focus on matching global distributional statistics (e.g., mean and variance), but overlook critical instance-level characteristics and intraclass variations, leading to suboptimal generalization. We address this limitation by reformulating dataset distillation as an Optimal Transport (OT) distance minimization problem, enabling fine-grained alignment at both global and instance levels throughout the pipeline. OT offers a geometrically faithful framework for distribution matching. 
It effectively preserves local modes, intra-class patterns, and fine-grained variations that characterize the geometry of complex, high-dimensional distributions.
% It effectively captures intra-class structures and local variations that are often smoothed out by objectives based on averaging samples in each class. 
Our method comprises three components tailored for preserving distributional geometry: (1) OT-guided diffusion sampling, which aligns latent distributions of real and distilled images; (2) label-image-aligned soft relabeling, which adapts label distributions based on the complexity of distilled image distributions; and (3) OT-based logit matching, which aligns the output of student models with soft-label distributions. Extensive experiments across diverse architectures and large-scale datasets demonstrate that our method consistently outperforms state-of-the-art approaches in an efficient manner, achieving at least 4\% accuracy improvement under IPC=10 settings for each architecture on ImageNet-1K.

\end{abstract}

\section{Introduction}
The expansion of data has fueled advances in deep learning, 
% it 
but also introduced prohibitive costs in storage, computation, and energy~\cite{dd_comprehensive_review,survey,layoutenc}.
To address these challenges, dataset distillation aims to synthesize a small set of training samples
% that enables 
to expedite model training while maintaining comparable performance~\cite{dd_begin}.
% models to achieve performance comparable to training on the full dataset
% This
Such distillation not only improves accessibility and cost-efficiency, but also facilitates practical applications such as knowledge transfer~\cite{transfer}, federated learning~\cite{federate,federate2}, and continual learning~\cite{continual,continual2}.
Moreover, it provides a valuable lens to investigate the theoretical principles underlying training efficiency and representation capacity in deep learning~\cite{insight,insight2}.

% Conventional dataset distillation methodologies generally fall into two categories: optimization-based~\cite{dc,idc,mtt,mttnew} and distribution-matching-based approaches~\cite{dm,idm,m3d,ncf}.
% 引出问题是：无论怎么归类，这些方法都没有办法scale up 数据集很小 分辨率也很低
%Conventional
Traditional dataset distillation methods can be broadly categorized into optimization-based~\cite{dc,idc,mtt,mttnew} and distribution-matching-based approaches~\cite{dm,idm,m3d,ncf}.
Despite their effectiveness, these methods remain largely restricted to small-scale, low-resolution datasets such as MNIST~\cite{mnist}, CIFAR~\cite{cifar}, %TinyImageNet~\cite{tinyimage}, 
or downsampled ImageNet subsets~\cite{imagenet}. 
This limitation stems from %two main issues: 
the prohibitive computational cost of alternating optimization between the distilled data and the condensation model~\cite{edc}, and the reliance on integrated image representations that demand costly pixel-level refinement~\cite{igd}.
% This limitation arises primarily from the computational burden of alternate optimization for the distilled set and observer model~\cite{edc}, which requires a massive number of unrolled iterations, and from the rigid treatment of data as fixed entities, necessitating exhaustive pixel-level refinements~\cite{igd}.
% 要不另起一段单独写为什么需要生成式方法来解决scale up问题
% 论点1：限制来源于BILEVEL OPTIMIZATION；需要大量计算来优化对齐
% Recent alternative approaches, including model inversion techniques and generative models, attempt to mitigate these scalability issues.
Recent efforts have explored generative and model-inversion-based techniques to overcome these scalability bottlenecks.
% 转折点1：有方法来简化计算-->生成式是个好思路来scaleup
%Among these, 
% Model-inversion-based methods~\cite{cda,sre2l,rded,edc}, primarily derived from data-free knowledge distillation frameworks, are completely based on batch normalization statistics extracted from condensed models.
% This restrictive approach limits their ability to capture detailed geometric distributional structures.
Model-inversion-based methods~\cite{cvdd,cda,sre2l,rded}, originally proposed under data-free distillation framework~\cite{yin2020dreaming,smith2021always}, rely entirely on global batch-normalization statistics extracted from pretrained models. While simple,
%and efficient, 
this design imposes an inherent limitation: it fundamentally lacks the ability to recover or preserve instance-level, local distributional structures.
In contrast, generative-model-based methods~\cite{igd,d3m,d4m,tdsdm} 
leverage real image samples during the sampling process, showing potential to approximate the true data distribution more faithfully.%to better align with the real data distribution.

However,
% despite their %theoretical potential,
existing generative approaches have yet to fully realize this promise, as they still solely focus on matching global gradient statistics.
Besides, the fine-grained distributional structures are not properly captured by cosine-similarity-based diversity guidance, resulting in local mode collapse and distributional mismatch in the distilled set.
% Besides,
% the fine-grained 
% distributional structures
% are not properly quantified via the existing cosine-similarity based diversity guidance,
% restricting the representation capacity of the distilled set.
% and the diversity of distilled images are 
% introduce diversity through random noise, 
% without explicitly modeling instance-specific semantic variations.
To address this limitation, we propose a principled reformulation grounded in Optimal Transport (OT), which enables fine-grained 
distributional geometry alignment between real distribution and model output distribution.
%Specifically, we define \emph{distributional geometry alignment} as preserving global statistical properties alongside instance-specific local structures and subclass distributions within each class.
% Specifically, We define \emph{distributional geometry alignment} as preserving both global and local distributional structures. This includes global statistics, local patterns, and subclass distributions within each class.
Specifically, we define \emph{distributional geometry alignment} as preserving distribution-level global and local structures (e.g., coarse-grained patterns and subclass densities), rather than image-level features.
% Specifically, we define \emph{distributional geometry alignment} as the preservation of not only representative  global properties but also the local neighborhood structure and subclass properties within each class.
% However, despite their theoretical potential, existing generative approaches have yet to fully realize this promise. They often prioritize matching global statistical profiles or introduce diversity through random noise, which fails to explicitly preserve instance-specific semantic variations.
% To overcome this limitation, we propose a principled reformulation grounded in Optimal Transport (OT), which enables fine-grained 
% distributional geometry alignment between real data distributions and model output distributions.
% Specifically,
% we define \emph{distributional geometry alignment} as the process of preserving not only global distributional information but also the local and relational structure between individual samples in the dataset. 
Our key insight is that each real data point encapsulates rich intra-class semantic variation, such as the distinctive traits of different subclasses or local modes within the same class. 
% OT offers a theoretically grounded mechanism to retain and transfer this fine-grained structure during distillation~\cite{sink,sink2,sinkd}, enabling more faithful alignment than traditional class-averaged objectives.
OT inherently provides a geometrically faithful and perceptually aligned measure of distributional differences~\cite{sink}, making it especially 
% effective 
promising for preserving and transferring these fine-grained semantic structures during distillation. %Consequently, OT achieves more accurate alignment than traditional class-level objectives.
% Our key insight is that every real data point carries valuable intraclass information, and OT enables us to explicitly integrate this fine-grained information (e.g. the unique characteristics of different types of subclasses within a class) during distillation.

\begin{figure}
    \centering
    \includegraphics[width=\linewidth]{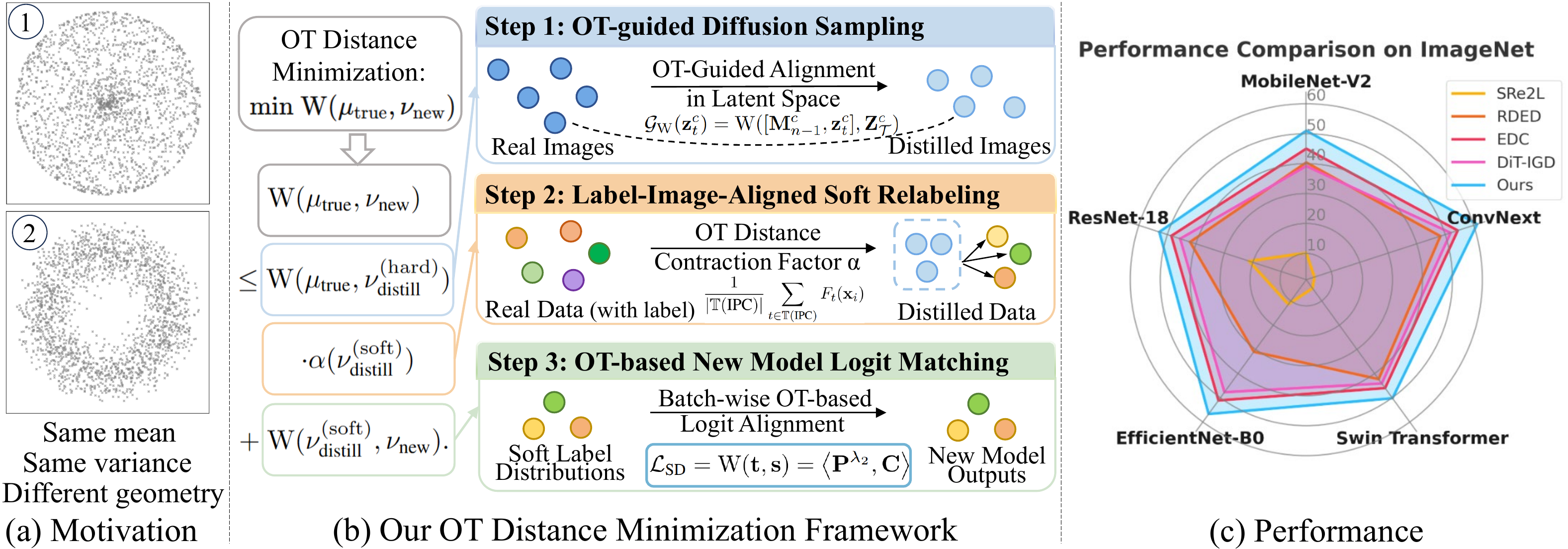}
    %\vspace{-0.3em}
    \caption{(a) %Matching global statistics (mean/variance) fails to capture distributional differences. 
    %Distributions with identical means and variances can exhibit vastly different geometric structures, indicating that relying solely on global statistics for optimization may introduce significant bias.
    %Distributions with identical global statistics may differ geometrically, making global-statistics-based optimization biased.
    Distributions with identical mean or variance may differ geometrically, causing biases in global-statistics-based optimization.
(b) We decompose the OT objective into three stages: OT-guided diffusion sampling, label-image-aligned soft relabeling, and OT-based logit matching.
% (c) Consistent improvements are achieved over prior methods on ImageNet-1K (IPC=10) across %various 
% architectures.
(c) %Our method achieves consistent gains over prior approaches on ImageNet-1K (IPC=10) across %diverse 
%architectures.
Our method consistently outperforms prior approaches across architectures on ImageNet-1K (IPC = 10).
}
    \label{framework}
\end{figure}

Building upon this, we formulate dataset distillation as an OT distance minimization problem.
As shown in Figure~\ref{framework}, to make the alignment process tractable and optimization-friendly, we decompose the total OT distance into three complementary objectives that altogether contribute to its minimization:
(1) instance-level transport in image latent space,
(2) label-image alignment in label space, and
(3)
% final output 
batch-wise
logit alignment between new model predictions and soft targets.
Such decomposition ensures alignment through the sequential stages of dataset distillation, ranging from image generation and label assignment to student model training.
% and ensures alignment at each stage.
% This decomposition enables each stage of the distillation pipeline to minimize a well-defined objective, resulting in more faithful and generalizable distributional alignment.
In the first stage, latent space transport is achieved by continuously computing the OT distance between the accumulated synthetic images (including the newly generated samples) and the real image batches at each sampling step. The gradients from this computation are used to guide the diffusion sampling process. In the second stage, we align the
complexity of the synthetic image distribution with that of the soft label distribution,
% complex distribution characteristics of synthetic images with those of soft labels,
thereby narrowing the OT distance between the distilled data and the real data.
In the final stage, we transfer the rich distributional geometric information embedded in the distilled set to new student models by minimizing the batch-wise OT distance between the student outputs and the soft-label distributions.

% We evaluate our method across diverse architectures such as ResNet, MobileNet, EfficientNet, Swin Transformer, ConvNet, and ConvNeXt. Our approach consistently outperforms state-of-the-art distillation methods in all datasets, architectures and IPCs, with particularly strong results in challenging IPC=10 scenarios. Our contributions are threefold: 
We evaluate our method across a diverse range of architectures, including ResNet, MobileNet, EfficientNet, Swin Transformer, ConvNet, and ConvNeXt. Our approach consistently outperforms state-of-the-art methods across all datasets, architectures, and IPCs, achieving particularly strong results on ImageNet-1K~\cite{imagenet} under the challenging IPC=10 scenario. Our contributions are threefold: 
\begin{itemize}
\item We propose a novel perspective of dataset distillation by formulating the task as an OT distance minimization problem.
We decompose 
% the total OT distance 
the objective into three tractable components.
\item 
We systematically enhance distributional geometry alignment through key stages of the pipeline, including image sampling, soft label relabeling, and student model logit matching.
\item We demonstrate the effectiveness and generalizability of our method across a broad range of datasets and model architectures, which significantly surpass existing techniques.
%We demonstrate the scalability and effectiveness of our approach across diverse datasets and model architectures, significantly outperforming existing methods.%, especially in challenging IPC=10 scenarios.
\end{itemize}

\section{Related Works}
%\subsection{Dataset Distillation}
% There exist many studies on dataset distillation. Early researchers use low-resolution oriented methods for small-scale datasets, while recent works are typically designed for large-scale datasets.
Numerous studies have investigated dataset distillation: initial works target low-resolution, small-scale datasets, while more recent methods address large-scale, higher-resolution scenarios.

\subsection{Small-scale distillation methods}
Traditional dataset distillation methods can be broadly classified into optimization-based and distribution-matching (DM)-based approaches.
Optimization-based methods~\cite{dc,idc,dream,mtt,mttnew} adopt a bi-level optimization framework, where model parameters are updated in the outer loop while synthetic data are refined in the inner loop to match gradients or trajectories.
% Although effective, these methods suffer from high computational costs due to the alternating optimization of models and data until convergence.
In contrast, DM-based methods~\cite{dm,idm,m3d,dance,optical,ncf} directly align the feature distributions of real and synthetic data, thereby avoiding costly nested optimization. However, all these methods 
%rely on the observer model for matching, 
exhibit high model dependence on the condensation model,
which limits the
versatility of the distilled datasets in generalizing across different architectures~\cite{glad,agnostic}. 
Also, they incur significant time and memory costs due to three factors: (1) treating synthetic data as fixed entities, (2) requiring exhaustive pixel-level refinements, and (3) relying on real data for image refinement. As a result, traditional dataset distillation approaches are predominantly applied to small-scale datasets.

\subsection{Large-scale distillation methods}
%To address the scalability limitations of traditional methods, 
Recent methods for large-scale, high-resolution datasets fall into two main categories: model-inversion-based and generative-model-based methods.
Model-inversion-based methods~\cite{sre2l,rded,edc,gvbsm,scdd,wmdd,cui2025rethinking} compress the real dataset into a compact model representation, eliminating the need for real data during image refinement. This reduces memory overhead and allows scaling to large datasets such as ImageNet-1K~\cite{imagenet}. However, the lack of real data in the reconstruction process results in the loss of fine-grained instance-level information, which hinders the distilled dataset from accurately capturing the structural properties and instance-specific characteristics of the real distribution.
Generative-model-based methods~\cite{d4m,dit,minimax,DDPS,igd} leverage pretrained generative models to avoid pixel-level refinements. 
% While it generate one sample at a time reduces memory overhead and avoids treating data as fixed entities, the independent synthesis of each sample fails to create a consistent and coherent overall distribution, limiting the ability of the distilled dataset to capture the full diversity and structural relationships inherent in the real data distribution. 
While generating one sample at a time reduces memory overhead and avoids treating data as fixed entities, independently synthesizing each sample prevents the distilled dataset from maintaining a coherent overall distribution, thereby limiting its ability to capture the full diversity and structural relationships of the real data distribution.
% Due to the intrinsic limitations of model-inversion-based approaches, particularly their inability to preserve the fine-grained structure of the real distribution, we adopt the generative-model-based paradigm as our starting point.
Due to their inherent exclusion of real data during reconstruction, model-inversion-based methods fail to preserve fine-grained structures of the real distribution; accordingly, we adopt the generative-model-based paradigm as our starting point.
We address the shortcomings of both families by proposing an OT framework that ensures distributional geometry alignment throughout the distillation process.
% In this work, we propose a novel approach that gradually approximates the distributional geometry of real datasets. By computing optimal transport distances in the latent space, we guide diffusion-based sampling to ensure a structured and distribution-preserving distillation process.

\section{Preliminaries}
% Diffusion models generate data by learning a parameterized reverse process that denoises a progressively perturbed input.
%Given a data distribution \( q(x) \), the forward process gradually corrupts a clean sample \( \mathbf{z}_0 \sim q(x) \) through Gaussian noise:
Given images $\mathbf{x}\sim q(\mathbf{x})$, define the induced latent distribution $q_Z$ by
$
  \mathbf{z}_0 = E(\mathbf{x})$,  $\mathbf{z}_0\sim q_Z(\mathbf{z_0}),
$ where $E$ is the encoder mapping images into latent space.
A latent diffusion model learns
$
  p_{\phi}(\mathbf{z}_0) \approx q_Z(\mathbf{z}_0)\,,
$
from which we can efficiently sample.  Let $D$ %be a decoder satisfying
be a decoder that reconstructs images via
$
  \hat{\mathbf{x}} = D(\mathbf{z}_0).
$
The forward noising process corrupts the clean latent~$\mathbf{z}_0$ via Gaussian perturbations:
% Given samples from the image distribution \( q(\mathbf{x}) \), diffusion models learn a parameterized distribution \( p_{\phi}(\mathbf{x}) \) that approximates \( q(\mathbf{x}) \) and can easily sample from it. An encoder \( E \) transforms the image into the latent space, \( \mathbf{z}_0 = E(\mathbf{x}) \), and a decoder \( D \) reconstructs the latent code back to the image space, \( \hat{\mathbf{x}} = D(\mathbf{z}_0) \).  
% The forward process corrupts the latent code \( \mathbf{z}_0 \) with Gaussian noise $\bm{\epsilon}$:
\begin{equation}
    \mathbf{z}_t = \sqrt{\alpha_t} \mathbf{z}_0 + \sqrt{1 - \alpha_t} \bm{\epsilon}, \quad \bm{\epsilon} \sim \mathcal{N}(0, \mathbf{I}),
\end{equation}
where \( \alpha_t \) controls the noise schedule. The reverse process reconstructs clean samples via a parameterized denoising function \( \epsilon_{\phi}(\mathbf{z}_t, t) \), iteratively refining noisy inputs with update function $s$:
\begin{equation}
    \mathbf{z}_{t-1} = s(\mathbf{z}_t, t, \epsilon_{\phi}(\mathbf{z}_t, t)).
\end{equation}
Guided diffusion~\cite{yu2023freedom} modifies this process by introducing an auxiliary guidance function \( \mathcal{G}(\mathbf{z}_t,t) \) that adjusts the sampling trajectory. This allows generation to be conditioned on labels, structural priors, or more abstract objectives by modifying \( s(\mathbf{z}_t, t, \epsilon_{\phi}) \) to optimize for a task-specific constraint.

Influence-Guided Diffusion (IGD)~\cite{igd} leverages guided diffusion for dataset distillation by modifying the reverse sampling process to generate training-optimal data. Instead of passively sampling from \( p_{\phi}(\mathbf{x}) \), IGD introduces trajectory influence function and diversity function into the diffusion process to prioritize samples. %that maximize model generalization. 
Given a guided diffusion framework, IGD modifies the sampling update as:
\begin{equation}
    \mathbf{z}_{t-1} =s(\mathbf{z}_t, t, \epsilon_{\phi}) - \rho_t \nabla_{\mathbf{z}_t} \mathcal{G}_I(\mathbf{z_t},t) - \gamma_t \nabla_{\mathbf{z}_t} \mathcal{G}_D(\mathbf{z}_t),
\end{equation}
where \( \mathcal{G}_I(\mathbf{z}_t,t)\) represents the influence function for global distributional trajectory matching, and \( \mathcal{G}_D(\mathbf{z}_t) \) enforces diversity to prevent redundancy in the distilled dataset. 

\section{Methods}
% \subsection{Problem Statement}
% Dataset distillation
% aims to create a small synthetic dataset $\mathcal{S}\equiv\nu_{\text{distill}}$  from a large-scale full dataset $\mathcal{T} \equiv\nu_{\text{true}}$, using far less examples $|\mathcal{S}| \ll |\mathcal{T}|$.
% A student model trained on \(\mathcal{S}\) should mimic training on
% \(\mathcal{T}\), meaning its output distribution,
% \(\nu_{\text{new}}(\mathbf{x},\mathbf{y})\), should remain close to
% the ground-truth distribution \(\mu_{\text{true}}(\mathbf{x},\mathbf{y})\).
% Formally, we measure this via a Wasserstein distance
% \(\mathrm{W}\bigl(\nu_{\text{new}}, \mu_{\text{true}}\bigr)\),
% aiming to make \(\mathrm{W}\bigl(\nu_{\text{new}}, \mu_{\text{true}}\bigr)\)
% as small as possible despite only accessing a compact synthetic set. We give a more detailed explanation on distributions used later in Table~\ref{tab:distributions}.

\subsection{Problem Statement}
Dataset distillation aims to construct a compact distilled dataset \(\mathcal{S} \equiv \nu_{\text{distill}}(\mathbf{x}, \mathbf{y})\) (this means that $\nu_{\text{distill}}$ denotes the empirical distribution over dataset $\mathcal{S}$) from a real, full dataset \(\mathcal{T} \equiv \mu_{\text{true}}(\mathbf{x}, \mathbf{y})\), such that \(|\mathcal{S}| \ll |\mathcal{T}|\). A student model trained on \(\mathcal{S}\) should mimic the performance of training on \(\mathcal{T}\), i.e., the output distribution \(\nu_{\text{new}}(\mathbf{x}, \mathbf{y})\) of the student model should remain close to the ground-truth distribution \(\mu_{\text{true}}(\mathbf{x}, \mathbf{y})\).
% We formulate the dataset distillation objective as an OT minimization problem
%:$
% \text{min} \ \mathrm{W}(\mu_{\text{true}},\nu_{\text{new}})
% $, where \(\mathrm{W}(\cdot,\cdot)\) represents Wasserstein distance. 
We formulate the dataset distillation objective as an OT minimization problem, where we minimize the Wasserstein distance $\mathrm{W}(\mu_{\text{true}},\nu_{\text{new}})$ between the ground-truth and student-induced distributions.
The key distributions involved in the formulations are summarized in Table~\ref{tab:distributions}. We provide a detailed description of symbols in Appendix~\ref{a3}.

\begin{table}[t]
\centering
\footnotesize
\caption{Distributions involved in dataset distillation. All are defined over the joint space \(\mathcal{X} \times \mathcal{Y}\), where \(\mathcal{X}\) denotes the image space and \(\mathcal{Y}\) denotes the label space.}
\label{tab:distributions}
\resizebox{\textwidth}{!}{
\begin{tabular}{@{}lccc@{}}
\toprule
\textbf{Distribution} & \textbf{Image Source} & \textbf{Label Source} & \textbf{Description} \\ \midrule
\(\mu_{\text{true}}(\mathbf{x}, \mathbf{y})\) & Real (full) Images & Ground-truth \(y(\mathbf{x})\) & True data-label distribution \\
\(\nu_{\text{distill}}^{(\mathrm{soft})}(\mathbf{x}, \mathbf{y})\) & Distilled (generated) Images & Teacher soft label \(\mathbf{t}(\mathbf{x})\) & Distilled data with soft label \\
\(\nu_{\text{distill}}^{(\mathrm{hard})}(\mathbf{x}, \mathbf{y})\) & Distilled (generated) Images & One-hot label \(y_{\text{onehot}}(\mathbf{x})\) & Distilled data with hard label \\
\(\nu_{\text{new}}(\mathbf{x}, \mathbf{y})\) & Distilled (generated) Images & Student logit output \(\mathbf{s}(\mathbf{x})\) & Output of the model trained on \(\mathcal{S}\) \\ \bottomrule
\end{tabular}}
\end{table}

\subsection{Reconstructing the Optimal Transport Distance}
We now provide a theoretical decomposition of our objective,
\(\mathrm{W}( \mu_{\text{true}},\nu_{\text{new}})\), by introducing two key principles:
(1) the triangle inequality partitioning the discrepancy introduced before and after distilled set construction, and
(2) a multiplicative contraction term reflecting the benefit of soft labels over hard labels.
Unlike the commonly used measures such as KL divergence and cosine similarity, which do not satisfy the triangle inequality, the Wasserstein distance  \(\mathrm{W}(\cdot,\cdot)\)  is a
% true
%rigorous 
proper metric on the space of distributions. This property allows us to decompose the total discrepancy as:
\begin{equation}
\mathrm{W}(\mu_{\text{true}},\nu_{\text{new}})
\leq
\mathrm{W}(\mu_{\text{true}}, \nu_{\text{distill}}^{(\mathrm{hard})})
+
\mathrm{W}(\nu_{\text{distill}}^{(\mathrm{hard})}, \nu_{\text{new}}).
\end{equation}
As soft labels better approximate the true label distribution than one-hot assignments, recent works adopt soft labels instead of hard one-hot assignments, leading to the following relaxed upper bound:
\begin{equation}
\mathrm{W}(\mu_{\text{true}},\nu_{\text{new}})
\leq
\underbrace{\mathrm{W}(\mu_{\text{true}}, \nu_{\text{distill}}^{(\mathrm{soft})})}_{\text{Dataset Discrepancy}}
+
\underbrace{\mathrm{W}(\nu_{\text{distill}}^{(\mathrm{soft})}, \nu_{\text{new}})}_{\text{Logit Matching Error}}.
\end{equation}
The first term captures the mismatch between the distilled %data distribution 
and %the 
real data distribution. The second term reflects the logit-wise alignment between the student model’s output distribution and the soft labels of the distilled data.
% Directly minimizing the first term is difficult.
% It is difficult to directly minimize the first term. 
%due to the imperfection of representation granularity.
% To further analyze the first term,
Directly minimizing the first term is challenging due to too many variables to optimize. 
To analyze it further, we model the soft label advantage 
%we express the soft label advantage 
via a multiplicative relation:
\begin{equation}
\mathrm{W}(\mu_{\text{true}},\nu_{\text{distill}}^{(\mathrm{soft})})
=
 \mathrm{W}( \mu_{\text{true}},\nu_{\text{distill}}^{(\mathrm{hard})})\cdot\alpha(\nu_{\text{distill}}^{(\mathrm{soft})})  %, \quad \text{where } 0 < \alpha(\nu_{\text{distill}}^{(\mathrm{soft})}) < 1.
\end{equation}
The contraction factor \(\alpha(\nu_{\text{distill}}^{(\mathrm{soft})})\) measures how much soft labels reduce the discrepancy between label and image distributions compared to hard labels alone.
%quantifies the extent to which soft labels reduce the label-image distribution discrepancy compared to hard labels.
% which depends critically on the match between the teacher's label complexity and the size and diversity of the synthetic dataset.
The contraction %by \(\alpha\) 
is achieved by matching the complexity of the teacher-provided soft label distributions %from the teachers with 
to that of the distilled image distribution.
Since both \(\nu_{\text{distill}}^{(\mathrm{hard})}\) and \(\mu_{\text{true}}\) use one-hot (hard) labels, their Wasserstein distance can be computed class-wise, by independently solving optimal transport between images of the same category:
\begin{equation}
    \mathrm{W}( \mu_{\text{true}},\nu_{\text{distill}}^{(\mathrm{hard})}) = \mathbb{E}_{y}\left[\mathrm{W}(\mu_{\text{true}}(\mathbf{x}\mid y),\nu_{\text{distill}}^{(\mathrm{hard})}(\mathbf{x}\mid y)\right]
\end{equation}
where $\mathbb{E}_{y}$ denotes the expectation over label classes,  which measures the average conditional Wasserstein distance across classes.
Putting everything together, we arrive at a structured upper bound:
% \begin{equation}
% \label{upper}
% \mathrm{W}(\mu_{\text{true}},\nu_{\text{new}})
% \leq
% \mathbb{E}_{y}\left[\mathrm{W}(\mu_{\text{true}}(\mathbf{x}\mid y),\nu_{\text{distill}}^{(\mathrm{hard})}(\mathbf{x}\mid y)\right]\cdot\alpha(\nu_{\text{distill}}^{(\mathrm{soft})})+\mathrm{W}( \nu_{\text{distill}}^{(\mathrm{soft})},\nu_{\text{new}}),
% \end{equation}
\begin{equation}
\label{upper}
\mathrm{W}(\mu_{\text{true}},\nu_{\text{new}})
\leq
\color{morandi1}{
\mathbb{E}_{y}\left[\mathrm{W}(\mu_{\text{true}}(\mathbf{x}\mid y),\nu_{\text{distill}}^{(\mathrm{hard})}(\mathbf{x}\mid y)\right]}
\cdot
\color{morandi2}{
\alpha(\nu_{\text{distill}}^{(\mathrm{soft})})}
+
\color{morandi3}{
\mathrm{W}( \nu_{\text{distill}}^{(\mathrm{soft})},\nu_{\text{new}})}
\color{black}
\end{equation}
where each term is controlled by a distinct design choice: 
\textcolor{morandi1}{OT-guided diffusion sampling}, 
\textcolor{morandi2}{label-image-aligned soft label relabeling}, and 
\textcolor{morandi3}{OT-based logit matching between the student model and the distilled dataset}.
% where each term is controlled by a distinct design choice: OT-based guided diffusion sampling, label-image-aligned soft label relabeling, and OT-based logit matching between the student model and the distilled dataset.
This decomposition allows a principled basis of our method, which explicitly targets at minimization of each component. Our pseudocodes are provided in Appendix~\ref{a4}.

% When the synthetic set is small (low IPC), simpler teachers producing smoother soft labels tend to yield smaller \(\alpha\), whereas more complex teachers may be preferable when the synthetic set is large enough to support finer-grained decision boundaries.

\subsection{OT-guided Diffusion Sampling (OTG)}
\label{4.3}
%我们将$\mathcal{T}$中第c类的样本经过encoder得到他们对应的latent$\mathbf{Z}_\mathcal{T}^c$。我们首先考虑上一节表达式右端的第一项，根据guided diffusion的基本原理,为了最小化$\mathrm{W}(\mu_{\text{true}}(\mathbf{x}\mid y),\nu_{\text{distill}}^{(\mathrm{hard})$,我们应该将sample的distilled data与real data的最优传输距离作为guidance function，并在sample每个样本的时候都运用guidance，即当我们在sample第c类的第n个样本时，我们使用如下guidance function：
In the remainder of this section, we optimize the three terms in Equation~\ref{upper} sequentially.
% In the following three subsections, we sequentially optimize the above three terms.
We now concentrate on the first term. %of the upper‑bound. 
For each class $c$, we minimize the class-conditional OT distance \( \mathrm{W}(\mu_{\text{true}}(\mathbf{x}\mid c), \nu_{\text{distill}}^{(\mathrm{hard})}(\mathbf{x}\mid c)) \) through diffusion guidance: we compute the OT distance between the distilled images and the real images in the latent space as the guiding function. %applying this at each sampling step. 
%Concretely, 
At each diffusion step during the generation of the \( n \)-th latent $\mathbf{z}^c_0$, we draw a random batch of class-\( c \) samples from dataset \( \mathcal{T} \) and encode them into %their corresponding
latent representations \( \mathbf{Z}_\mathcal{T}^c \).
% We then compute guidance via:
We then employ the following guidance function:
% Henceforth, we optimize the three terms sequentially. As a preparation step, we encode the samples of class \( c \) from dataset \( \mathcal{T} \) into their corresponding latent representations \( \mathbf{Z}_\mathcal{T}^c \). In this subsection, we begin by focusing on the first term of the upper bound, which was introduced in the previous section. To minimize this term, we leverage the diffusion principle to reduce the optimal transport distance:
% \( \mathrm{W}(\mu_{\text{true}}(\mathbf{x}\mid c), \nu_{\text{distill}}^{(\mathrm{hard})}(\mathbf{x}\mid c)) \)
% To achieve this, the optimal transport distance between the distilled samples and real data serves as the guidance function, which is applied iteratively at each sampling step. Specifically, when sampling the \( n \)-th sample of class \( c \), we employ the following guidance function:
\begin{equation}
    \mathcal{G}_\mathrm{W}(\mathbf{z}^c_t) = \mathrm{W}([\mathbf{M}^c_{n-1},\mathbf{z}^c_t],\mathbf{Z}_\mathcal{T}^c)
\end{equation}
%其中$\mathbf{M}^c$为之前已经sample出的样本。为了快速稳定的计算该最优传输距离，我们使用Sinkhorn方法来得到最优传输矩阵：
where \( \mathbf{M}^c_{n-1} \) denotes previously sampled $n{-}1$ latents for class $c$,
and $[\cdot]$ represents the pythonic concatenation.% operation.
We denote $[\mathbf{M}^c_{n-1},\mathbf{z}^c_t]$ as $\hat{\mathbf{M}}^c_n$. The OT matrix $\mathbf{P}^{\lambda_1}$ can be efficiently approximated:
% as:
\begin{equation}
\mathbf{P}^{\lambda_1}=\underset{\mathbf{P}}{\text{argmin}}\left<\mathbf{P},\mathbf{D}(\hat{\mathbf{M}}^c_n,\mathbf{Z}_\mathcal{T}^c)\right>-\lambda_1 h(\mathbf{P}),\ \text{where}\sum_i\mathbf{P}_{ij}=\frac{1}{|\mathbf{Z}_\mathcal{T}^c|} \; \forall j, \  \sum_j\mathbf{P}_{ij}=\frac{1}{n} \; \forall i.
\end{equation}
%其中\mathbf{D}(\hat{\mathbf{M}}^c,\mathbf{Z}_\mathcal{T}^c) denotes the cost matrix that measures the ``distance" between the real $\mathbf{Z}_\mathcal{T}^c$ and synthetic $\mathbf{M}}^c$ latents. 不失一般性，我们使用$\ell_p$-norm的cost matrix并初始化candidata最优传输矩阵：
where $h(\mathbf{P})$ is the entropy of $\mathbf{P}$,
$\left<\cdot,\cdot\right>$ denotes the Frobenius inner product, $\lambda_1>0$ is the entropy regularization weight, \( \mathbf{D}(\hat{\mathbf{M}}^c_n, \mathbf{Z}_\mathcal{T}^c) \) represents the cost matrix measuring the pairwise distance between the real latent representations \( \mathbf{Z}_\mathcal{T}^c \) and the sampled \( \hat{\mathbf{M}}^c_n \). Without loss of generality, we use the \( \ell_p \)-norm cost matrix and initialize the candidate transport matrix $\mathbf{K}^0$ as:
\begin{equation}
\label{eq:initK}
\mathbf{D}_{ij}(\hat{\mathbf{M}}^c_n,\mathbf{Z}_\mathcal{T}^c)= \;\parallel \hat{\mathbf{M}}^c_n[i]-\mathbf{Z}_\mathcal{T}^c[j]\parallel_p, \quad 
    \mathbf{K}^0=\exp(-\frac{\mathbf{D}}{\lambda_1}).
\end{equation}
Next, Sinkhorn normalization is applied through iterative updates to $\mathbf{K}$:
%Sinkhorn algorithm iteratively normalize a candidate transport matrix $\mathbf{K}^{i}$ to satisfy the marginal constraints:
\begin{equation}
\mathbf{\widehat{K}}^{i} \leftarrow \mathrm{diag} \left(\mathbf{K}^{i-1}\mathbf{1}_{n}\oslash({n}\mathbf{1}_{|\mathbf{Z}_\mathcal{T}^c|})\right)^{-1}\mathbf{K}^{i-1}, \ \
\mathbf{K}^{i} \leftarrow \mathbf{\widehat{K}}^{i}\mathrm{diag} \left(\left(\mathbf{\widehat{K}}^{i}\right)^{\text{T}}\mathbf{1}_{|\mathbf{Z}_\mathcal{T}^c|}\oslash({|\mathbf{Z}_\mathcal{T}^c|}\mathbf{1}_{n}))\right)^{-1},
   \label{12}
\end{equation}
%where $\oslash$ denotes element-wise division. After $T$ iterations, we obtain the optimal transport matrix并且可以计算出最优传输距离：
where \( \oslash \) denotes element-wise division, $(\cdot)^\text{T}$ indicates matrix transpose. After \( T \) iterations, the optimal transport matrix $\mathbf{P}^{\lambda_1}$ is obtained, and we can compute the Sinkhorn distance (an approximation of the OT distance) $\mathrm{W}$ as:
\begin{equation}
\mathbf{P}^{\lambda_1}=\mathbf{K}^{T}, \quad\mathrm{W}([{\mathbf{M}}^c_{n-1},\mathbf{z}^c_t],\mathbf{Z}_\mathcal{T}^c)=\left<\mathbf{P},\mathbf{D}(\hat{\mathbf{M}}^c_n,\mathbf{Z}_\mathcal{T}^c)\right>=\sum_{i,j}{\mathbf{K}^{T}_{ij}\mathbf{D}_{ij}}
\label{13}
\end{equation}
%对于整个guided diffusion sampling过程，we follow previous training approach described in preliminaries to use a combination of trajectory function, diversity function and our optimal transport function: 
For the entire guided diffusion sampling, we follow the previous approach to combine the terms of trajectory and diversity functions with our OT function.
The iteration of \(t=T_D\to1\) yields \(\mathbf{z}_0^c\):
\begin{equation}
   \mathbf{z}^c_{t-1} = s(\mathbf{z}^c_t, t, \epsilon_{\phi}) - \rho_t \nabla_{\mathbf{z}^c_t} \mathcal{G}_I(\mathbf{z}^c_t,t) - \gamma_t \nabla_{\mathbf{z}^c_t} \mathcal{G}_D(\mathbf{z}^c_t)-\beta_1\nabla_{\mathbf{z}^c_t}\mathcal{G}_\mathrm{W}(\mathbf{z}^c_t),
\end{equation}
where $\rho_t$, $\gamma_t$ and $\beta_1$ are weights, $\mathcal{G}_I(\mathbf{z}_t^c,t)$ and $\mathcal{G}_D(\mathbf{z}_t^c)$ are trajectory and diversity functions, respectively.
% introduced earlier. 
By minimizing the OT distance in the image distribution space, we account for the contribution of individual real images and incorporate both global and local structural information, thereby promoting fine-grained geometric alignment between distributions.
Finally, we use decoder $D$ to convert all latent representations into images, forming the distilled image set $\mathcal{S}_\mathbf{x}$.
%通过最小化图像分布空间的最优传输距离，我们充分考虑了图像分布空间中的几何信息，促进了分布的几何对齐。

% \subsection{Soft relabeling with label-data alignment}
% Here we focus on the second term $\alpha(\nu_{\text{distill}}^{(\mathrm{soft})})$, which depends on the alignment between the complexity of soft label distribution and the complexity of distilled image data distribution. As the latter is largely determined by Images-per-Class (IPC), we choose suitable teachers for soft relabeling in different IPC scenarios.
% In low IPC scenarios, the distilled image distribution is relatively simple, we also need relatively simple distribution. So, fewer teachers, which have simple and smooth output distribution, can prevent overfitting and maintain label clarity. In high IPC scenarios, more complex teachers can capture finer distributional geometry and richer semantic structures, which distill more label information on the distilled set. Specifically, the soft label is the average of all the outputs of the teachers in the set $\mathbb{T}$.
% \begin{equation}
% \mathbf{t(x_i)}=\frac{1}{|\mathbb{T}(|\mathcal{S}_\mathbf{x}|)|} \sum_{t \in \mathbb{T}(|\mathcal{S}_\mathbf{x}|)} {F_t(\mathbf{x_i})}=\frac{1}{|\mathbb{T}(\text{IPC})|} \sum_{t \in \mathbb{T}(\text{IPC})} {F_t(\mathbf{x_i})},\quad \text{for each} \ \mathbf{x}_i\in \mathcal{S}_\mathbf{x},
% \end{equation}
% where $F_t$ means the function of the $t$-th teacher. This ensures a more precise match between soft labels and the underlying data distribution, effectively minimizing the total optimal transport distance.

\subsection{Label-Image-Aligned Soft Label Relabeling (LIA)}
We now focus on the contraction factor \(\alpha(\nu_{\text{distill}}^{(\mathrm{soft})})\), which characterizes the alignment between the complexity of the soft label distribution and that of the distilled image distribution (Appendix~\ref{a7} for details). Since the representational capacity of the distilled dataset is primarily governed by the number of images per class (IPC), we adopt an IPC-aware %label assignment 
strategy that minimizes the overall OT distance.
In low-IPC regimes, the distilled image distribution is less expressive and more prone to overfitting. Assigning overly complex soft labels in such cases can introduce distributional mismatch and degrade alignment. To mitigate this, we employ a smaller number of representative teacher models to produce simplified, low-entropy soft labels that offer well-calibrated supervision.
In contrast, high-IPC regimes enable the distilled dataset to support greater semantic diversity. Accordingly, we leverage a larger and more diverse set of teacher models to generate fine-grained soft label distributions, which better capture the intrinsic structure of the true label space.
% We now focus on the contraction facotr \(\alpha(\nu_{\text{distill}}^{(\mathrm{soft})})\), which reflects the alignment between the complexity of the soft label distribution and that of the distilled image distribution. Since the representational capacity of the distilled dataset is primarily governed by the number of images per class (IPC), we adapt the label assignment strategy accordingly to minimize the overall optimal transport distance.
% In low-IPC regimes, the distilled image distribution is less expressive and prone to overfitting.
% Assigning overly complex soft labels in this setting introduces mismatch and degrades the alignment.
% We therefore employ fewer and more representative teacher models to form relatively simply distribution and provide well-calibrated, low-entropy supervision.
% In contrast, in high-IPC settings, the distilled dataset can support richer semantic variation.
% Leveraging stronger and more diverse teacher models enables finer-grained label distributions that better capture the true label geometry.
Formally, for each synthetic image, we assign the soft label as the averaged output from a set of IPC-dependent teachers:
\begin{equation}
\mathbf{t}(\mathbf{x}_i)=\frac{1}{|\mathbb{T}(|\mathcal{S}_\mathbf{x}|)|} \sum_{t \in \mathbb{T}(|\mathcal{S}_\mathbf{x}|)} {F_t(\mathbf{x}_i)}=\frac{1}{|\mathbb{T}(\text{IPC})|} \sum_{t \in \mathbb{T}(\text{IPC})} {F_t(\mathbf{x}_i)},\quad \text{for each} \ \mathbf{x}_i\in \mathcal{S}_\mathbf{x},
\end{equation}
where \(F_t\) denotes the logit output function of the \(t\)-th teacher, $\mathbf{t}(\mathbf{x}_i)$ denotes soft label for image $\mathbf{x}_i$, and \(\mathbb{T}(\text{IPC})\) %is a selected subset of teacher models selected according to the current IPC setting. 
is a subset of teacher models selected to minimize the contraction factor $\alpha$. 
This strategic relabeling ensures that the soft label distribution faithfully matches the capacity of the distilled images, reducing the discrepancy term \(\mathrm{W}(\nu_{\text{distill}}^{(\mathrm{soft})}, \mu_{\text{true}})\) and thereby improving alignment. 
For a fair comparison with prior methods~\cite{sre2l,rded}, we adopt the same region-level soft label storage strategy as in FKD~\cite{fkd}.

\subsection{OT-based Student Model Logit Matching (OTM)}
%在得到了蕴含丰富分布几何信息的distilled set之后，为了能够将这些几何信息传递到新模型上以实现新模型输出空间的分布和原始数据集的分布的几何对齐，我们需要最小化公式~\ref{upper}上界的第三项。我们考虑a batch of $b$ samples，并将这个batch的soft label和student model的输出分别记为$t$和$s$。existing divergence measures can only independently deal with each sample for logit-by-logit matching因此无法考虑样板间关系以从distilled data中充分获得分布的几何信息.为了解决这个问题，我们使用batch-wise的最优传输距离来进一步实现logit matching。具体而言，和4.3节类似，我们使用Sinkhorn方法添加正则项来快速求解最优传输矩阵$\mathbf{P}_{\lambda}$
% After obtaining the distilled set with rich distributional geometric information, we aim to transfer these information to %the new 
% student models in order to align the distribution of %the new
% each
% model’s logit output space with the distribution of the real dataset. 
After obtaining a distilled set that preserves rich geometric structures of the real data, we transfer this information to student models (i.e., new models) by aligning the distribution of their logits with that of the real dataset.
We achieve this alignment by minimizing the last term in the upper bound of Equation~\ref{upper}. We consider a batch of \( b \) samples and denote the soft labels of this batch and the logit output of a student model as
\( \mathbf{t} \) and \( \mathbf{s} \), respectively. %Most divergence measure can only handle logit-by-logit matching independently for each sample, thus failing to consider inter-sample relationships. % and fully capture the geometric distribution information from the distilled data. 
%To address this issue, we employ batch-wise optimal transport distance to further implement logit matching. 
Most traditional divergence measures operate on a per-sample basis and match logits independently, thereby failing to capture inter-sample relationships. To address this limitation, we employ a batch-wise OT distance that aligns logits while capturing global distributional structure.
Specifically, similar to Section~\ref{4.3}, we use the Sinkhorn method to efficiently solve for the OT matrix \(\mathbf{P}^{\lambda_2}\), with entropy regularization $h(\mathbf{P})$ weighted by \(\lambda_2\):
\begin{equation}
    \mathbf{P}^{\lambda_2}=\underset{\mathbf{P}}{\text{argmin}}\left<\mathbf{P},\mathbf{C}(\mathbf{t},\mathbf{s})\right>-\lambda_2 h(\mathbf{P}),\ \text{where}\sum_i\mathbf{P}_{ij}=\frac{1}{b} \, \forall j, \  \sum_j\mathbf{P}_{ij}=\frac{1}{b} \, \forall i.
\end{equation}
% We employ the $\ell_p$-norm to measure the pairwise differences between the $i$-th and $j$-th samples in a batch for the entry $\mathbf{C}_{ij}$ of the ``batchified" cost matrix $\mathbf{C}\in\mathbb{R}^{b\times b}$:
%We employ the \( \ell_p \)-norm to measure the pairwise differences between the \( i \)-th and \( j \)-th samples in a batch for the entry \( \mathbf{C}_{ij} \) of the         ``batchified'' cost matrix \( \mathbf{C} \in \mathbb{R}^{b \times b} \):
Here, \( \mathbf{C} \in \mathbb{R}^{b \times b} \) is the batch-wise cost matrix, where each entry \( \mathbf{C}_{ij} \) measures the distance between the soft label \(\mathbf{t}(\mathbf{x}_i)\) and the synthetic output \(\mathbf{s}(\mathbf{x}_j)\). Specifically, we employ the \( \ell_p \)-norm:
\begin{equation}
\mathbf{C}_{ij}(\mathbf{t},\mathbf{s})= \;\parallel \mathbf{t}(\mathbf{x}_i)-\mathbf{s}(\mathbf{x}_j)\parallel_p, \quad \mathcal{L}_{\text{SD}}=\mathrm{W}(\mathbf{t},\mathbf{s})=\left<\mathbf{P}^{\lambda_2},\mathbf{C}\right>, 
\end{equation}
%参照公式~\ref{init,12,13}的方法可以计算出最有传输矩阵$\mathbf{P}^{\lambda}$和batch-wise Sinkhorn distance loss $\mathcal{L}_{\text{SD}}$。参照G-VBSM和EDC的做法，我们将保持单样本语义一致性的CE loss，MSE loss和我们的SD loss一起使用作为total loss：
Adapting Equations~\ref{eq:initK}, \ref{12}, and \ref{13} to current dimensions, we compute %the OT matrix 
\( \mathbf{P}^{\lambda_2} \) and the corresponding batch-wise Sinkhorn distance loss \( \mathcal{L}_{\text{SD}} \). 
For a fair comparison with previous methods~\cite{gvbsm,edc}, we use the cross-entropy loss $\mathcal{L}_{\text{CE}}$,
the MSE loss $\mathcal{L}_{\text{MSE}}$,
and the Sinkhorn loss $\mathcal{L}_{\text{SD}}$ for distillation:
\begin{equation}
\mathcal{L}=\sum_{i=1}^{b}\kappa_1\mathcal{L}_{\text{CE}}(y_\text{onehot}(\mathbf{x}_i),\mathbf{s}(\mathbf{x}_i))\\
    +\kappa_2\mathcal{L}_{\text{MSE}}(\mathbf{t}(\mathbf{x}_i),\mathbf{s}(\mathbf{x}_i))+\beta_2\mathcal{L}_{\text{SD}},
\end{equation}
where \( \kappa_1 \), \( \kappa_2 \), and \( \beta_2 \) are scalar weights, \( y_{\text{onehot}}(\mathbf{x}_i) \) denotes the hard label for the distilled image \( \mathbf{x}_i \).

\section{Experiments}
\subsection{Experimental Settings}
\paragraph{Dataset.} Given our primary focus on large-scale dataset distillation, we evaluate our method on the full ImageNet-1K dataset~\cite{imagenet}. To ensure comparability across varying category scales, we further conduct experiments on two widely used subsets, ImageNet-100~\cite{idc} and ImageNette~\cite{nette}. To construct a comprehensive benchmark covering both low-resolution and high-resolution settings, we additionally include CIFAR-100~\cite{cifar}. The dataset descriptions are presented in Appendix~\ref{a2}.
%, allowing us to examine the robustness of our method across different dataset granularities.

% \textbf{Network architectures.} To evaluate the generalization capability of our method, we experiment with a diverse set of network architectures, including convolutional neural networks (CNNs), transformer-based models and hybrid models. We consider CNN-based architectures including ResNet~\cite{resnet}, MobileNet~\cite{mobilenet},EfficientNet~\cite{efficientnet}, and ConvNet~\cite{convnet}, transformer-based models such as the Swin Transformer~\cite{SWIN}, and the hybrid architecture ConvNeXt~\cite{convnext}. This selection ensures a comprehensive evaluation across different architectural paradigms and inductive biases.

\paragraph{Network architectures.} To evaluate the generalization capability of our method, we experiment with a diverse set of network architectures, including convolutional neural networks (CNNs), transformer-based models, and hybrid models. Specifically, we consider CNN-based architectures, including ResNet~\cite{resnet}, MobileNet~\cite{mobilenet},EfficientNet~\cite{efficientnet}, and ConvNet~\cite{convnet}; a transformer-based model, the Swin Transformer~\cite{SWIN}; and the hybrid architecture ConvNeXt~\cite{convnext}. This selection provides a comprehensive evaluation across diverse architectural paradigms and inductive biases.

\paragraph{Baselines.} We compare our approach with a broad range of dataset distillation methods. Specifically, we include traditional methods such as DM~\cite{dm}, IDC~\cite{idc}, and DATM~\cite{datm}; model-inversion-based methods including SRe$^2$L~\cite{sre2l}, G-VBSM~\cite{gvbsm}, RDED~\cite{rded}, CDA~\cite{cda}, SC-DD~\cite{scdd}, EDC~\cite{edc}, CV-DD~\cite{cvdd}, and DELT\cite{delt}; as well as generative-model-based methods such as D$^3$M~\cite{d3m}, D$^4$M~\cite{d4m}, TDSDM~\cite{tdsdm}, DiT~\cite{dit}, Minimax~\cite{minimax}, DDPS~\cite{DDPS}, and IGD~\cite{igd}. We report the top-1 test accuracy of models trained on distilled datasets with different IPC (Images Per Class) settings to ensure a fair and consistent comparison. Each network is trained five times from scratch to report error bars.

\begin{table}[tbp]
    \centering
    % \footnotesize
    \caption{Performance comparison on ImageNet-1K~\cite{imagenet} with ResNet-18. The numbers in parentheses for ``Ours'' represent the number of training epochs on the distilled set for new models.}
    \label{tab:imagenet_comparison}
    \resizebox{\textwidth}{!}{
    \begin{tabular}{c | c c c c c c c c c}
        \toprule
        \multicolumn{10}{c}{Comparsion with generative-model-based methods. } \\
        \midrule
        IPC & D$^3$M~\cite{d3m} & D$^4$M~\cite{d4m} & TDSDM~\cite{tdsdm} & DiT~\cite{dit} & Minimax~\cite{minimax} & DDPS~\cite{DDPS} & DiT-IGD~\cite{igd}  & Ours (300) & Ours (1000) \\
        \midrule
        10 & 23.6$\pm$0.1 & 27.9$\pm$0.7 & 44.5$\pm$0.4 & 39.6$\pm$0.4 & 42.1$\pm$0.3 & 42.1$\pm$0.3& 45.5$\pm$0.5  & \textbf{52.9}$\pm$\textbf{0.1} & \textbf{58.6}$\pm$\textbf{0.3}\\
        50 & 32.2$\pm$0.1 & 55.2$\pm$0.3 & 59.4$\pm$0.3 & 52.9$\pm$0.6 & 59.4$\pm$0.2& 59.4$\pm$0.2& 59.8$\pm$0.3  & \textbf{61.9}$\pm$\textbf{0.5} & \textbf{64.2}$\pm$\textbf{0.4}\\
                \midrule
        \multicolumn{10}{c}{Comparsion with model-inversion-based methods} \\
        \midrule
        IPC & SRe$^2$L~\cite{sre2l} & G-VBSM~\cite{gvbsm} & RDED~\cite{rded} & CDA~\cite{cda} & SC-DD~\cite{scdd} & EDC~\cite{edc} & CV-DD~\cite{cvdd} & Ours (300) & Ours (1000)\\
        \midrule
        10 & 21.3$\pm$0.6 & 31.4$\pm$0.5 & 42.0$\pm$0.1 & 33.5$\pm$0.3 & 32.1$\pm$0.2 & 48.6$\pm$0.3 & 46.0$\pm$0.6 &\textbf{52.9}$\pm$\textbf{0.1} & \textbf{58.6}$\pm$\textbf{0.3}\\
        50 & 46.8$\pm$0.2 & 51.8$\pm$0.4 & 56.5$\pm$0.1 & 53.5$\pm$0.3 & 53.1$\pm$0.1 & 58.0$\pm$0.2 & 49.5$\pm$0.4 & \textbf{61.9}$\pm$\textbf{0.5} & \textbf{64.2}$\pm$\textbf{0.4}\\
        \bottomrule
    \end{tabular}}
\end{table}

% \begin{table}[htbp]
%     \centering
%     % \footnotesize
%     \caption{Performance comparison on ImageNet-1K dataset.}
%     \label{tab:imagenet_comparison}
%     \resizebox{\textwidth}{!}{
%     \begin{tabular}{c | c c c c c c c c c}
%         \toprule
%         \multicolumn{10}{c}{ResNet-18} \\
%         \midrule
%         IPC & SRe$^2$L & G-VBSM & RDED & CDA & SC-DD & EDC & CV-DD & Ours (300) & Ours (1000)\\
%         \midrule
%         10 & 21.3$\pm$0.6 & 31.4$\pm$0.5 & 42.0$\pm$0.1 & 33.5$\pm$0.3 & 32.1$\pm$0.2 & 48.6$\pm$0.3 & 46.0$\pm$0.6 &\textbf{52.9}$\pm$\textbf{0.4} & \textbf{58.6}$\pm$\textbf{0.3}\\
%         50 & 46.8$\pm$0.2 & 51.8$\pm$0.4 & 56.5$\pm$0.1 & 53.5$\pm$0.3 & 53.1$\pm$0.1 & 58.0$\pm$0.2 & 49.5$\pm$0.4 & \textbf{61.9}$\pm$\textbf{0.2} & \textbf{64.2}$\pm$\textbf{0.4}\\
%                 \midrule
%         \multicolumn{10}{c}{ResNet-101} \\
%         \midrule
%         IPC & SRe$^2$L & G-VBSM & RDED & CDA & SC-DD & EDC & CV-DD & Ours (300) & Ours (1000)\\

%         \midrule
%         10 & 30.9$\pm$0.1 & 38.2$\pm$0.4 & 42.1$\pm$1.0 & 39.4$\pm$0.3 & 39.6$\pm$0.4 & 51.7$\pm$0.3 & 46.0$\pm$0.6 &\textbf{56.8}$\pm$\textbf{0.4} & \textbf{63.5}$\pm$\textbf{0.2}\\
%         50 & 60.8$\pm$0.5 & 61.0$\pm$0.4 & 61.2$\pm$0.4 & 61.3$\pm$0.3 & 61.0$\pm$0.3 & 64.9$\pm$0.2 & 49.5$\pm$0.4 & \textbf{68.0}$\pm$\textbf{0.1} & \textbf{70.9}$\pm$\textbf{0.3}\\
%         \bottomrule
%     \end{tabular}}
% \end{table}

\begin{table}[tbp]
    \centering
    \caption{Other architecture performance comparison on ImageNet-1K~\cite{imagenet}. }
    \label{tab:imagenet_generalization}
    \resizebox{0.98\textwidth}{!}{
    \begin{tabular}{l|cc|cc|cc|cc}
        \toprule
       \multirow{2}*{Method}  & \multicolumn{2}{c|}{MobileNet-V2} & \multicolumn{2}{c|}{EfficientNet-B0} & \multicolumn{2}{c|}{Swin Transformer} & \multicolumn{2}{c}{ConvNeXt}\\
        & IPC10 & IPC50 & IPC10 & IPC50 & IPC10 & IPC50 & IPC10 & IPC50 \\
        \midrule
        SRe$^2$L~\cite{sre2l} &10.2$\pm$2.6 & 31.8$\pm$0.3&11.4$\pm$2.5 & 34.8$\pm$0.4&4.8$\pm$0.6&42.1$\pm$0.3&4.1$\pm$0.4&48.8$\pm$0.2\\
        RDED~\cite{rded} &  40.4$\pm$0.1 & 53.3$\pm$0.2 & 31.0$\pm$0.1 & 58.5$\pm$0.4 & 42.3$\pm$0.6 & 53.2$\pm$0.8&48.3$\pm$0.5&65.4$\pm$0.4 \\
        EDC~\cite{edc} &45.0$\pm$0.2&57.8$\pm$0.1&51.1$\pm$0.3&60.9$\pm$0.2&46.0$\pm$0.5&57.9$\pm$0.3&54.4$\pm$0.2&66.6$\pm$0.2 \\
        DiT-IGD~\cite{dit} &39.2$\pm$0.2& 57.8$\pm$0.2 & 47.7$\pm$0.1 & 62.0$\pm$0.1 & 44.1$\pm$0.6 & 58.6$\pm$0.5 &51.9$\pm$0.2&66.8$\pm$0.5\\
        Ours (300) & \textbf{51.0}$\pm$\textbf{0.6} & \textbf{61.0}$\pm$\textbf{0.4} & \textbf{56.7}$\pm$\textbf{0.2} & \textbf{64.4}$\pm$\textbf{0.1} & \textbf{50.2}$\pm$\textbf{0.2} & \textbf{68.2}$\pm$\textbf{0.1} & \textbf{61.2}$\pm$\textbf{0.1} & \textbf{70.2}$\pm$\textbf{0.8} \\
        Ours (500) & \textbf{54.6}$\pm$\textbf{0.3} & \textbf{63.0}$\pm$\textbf{0.4} & \textbf{59.6}$\pm$\textbf{0.2} & \textbf{66.0}$\pm$\textbf{0.6} & \textbf{56.2}$\pm$\textbf{1.0} & \textbf{69.4}$\pm$\textbf{0.1} & \textbf{64.5}$\pm$\textbf{0.3} & \textbf{71.1}$\pm$\textbf{1.1} \\
        Ours (1000) & \textbf{57.6}$\pm$\textbf{0.1} & \textbf{63.9}$\pm$\textbf{0.2} & \textbf{62.4}$\pm$\textbf{0.1} & \textbf{66.8}$\pm$\textbf{0.1} & \textbf{63.7}$\pm$\textbf{0.2} & \textbf{70.5}$\pm$\textbf{0.1} & \textbf{67.0}$\pm$\textbf{0.1} & \textbf{71.8}$\pm$\textbf{0.9} \\
        % Ours (300) &51.0$\pm$0.6&61.0$\pm$0.4&56.7$\pm$0.2&64.4$\pm$0.1&50.2$\pm$0.2&68.2$\pm$0.1&61.2$\pm$0.1&70.2$\pm$0.8\\
        % Ours (500) &55.3$\pm$0.3&63.0$\pm$0.4&59.6$\pm$0.2&66.0$\pm$0.6&56.2$\pm$1.0&69.4$\pm$0.1&64.5$\pm$0.3&71.1$\pm$1.1\\
        % Ours (1000) &58.5$\pm$0.1&63.9$\pm$0.2&62.4$\pm$0.1&66.8$\pm$0.1&63.7$\pm$0.2&70.5&67.0$\pm$0.1&71.8$\pm$0.9\\
        % Minimax-IGD & 53.4$\pm$0.9 & 66.8$\pm$0.2 & 39.7$\pm$0.4 & 58.5$\pm$0.3 & \textbf{48.5}$\pm$\textbf{0.1} & \textbf{62.7}$\pm$\textbf{0.2} & \textbf{44.8}$\pm$\textbf{0.8} & 58.2$\pm$0.5 \\
        \bottomrule
    \end{tabular}}
\end{table}

\begin{table}[tb]
    \caption{Performance comparison on ImageNette~\cite{nette}.}
    
    \label{nette}
     \centering
    \resizebox{\textwidth}{!}{
    \begin{tabular}{l | c c c | c c c | c c c}
        \toprule
        Model & \multicolumn{3}{c|}{ConvNet-6} & \multicolumn{3}{c|}{ResNetAP-10} & \multicolumn{3}{c}{ResNet-18} \\
        \midrule
        IPC & 10 & 50 & 100 & 10 & 50 & 100 & 10 & 50 & 100 \\
        \midrule
        \multicolumn{10}{c}{Hard Label} \\
        \midrule
        Random     & 46.0$\pm$0.5  & 71.8$\pm$1.2  & 79.9$\pm$0.8  & 54.2$\pm$1.2  & 77.3$\pm$1.0  & 81.1$\pm$0.6  & 55.8$\pm$1.0  & 75.8$\pm$1.1  & 82.0$\pm$0.4  \\
        DM~\cite{dm}         & 49.8$\pm$1.1  & 70.3$\pm$0.8  & 78.5$\pm$0.8  & 60.2$\pm$0.7  & 76.7$\pm$1.1  & 80.9$\pm$0.7  & 60.9$\pm$0.7  & 75.0$\pm$1.0  & 81.5$\pm$0.4  \\
        IDC-1~\cite{idc}      & 48.2$\pm$1.2  & 72.4$\pm$0.7  & 80.6$\pm$1.1  & 60.4$\pm$0.6  & 77.4$\pm$0.7  & 81.5$\pm$1.2  & 61.0$\pm$0.8  & 77.5$\pm$1.0  & 81.7$\pm$0.8  \\
        DiT~\cite{dit}        & 56.2$\pm$1.3  & 74.1$\pm$0.6  & 78.2$\pm$0.3  & 62.8$\pm$0.8  & 76.9$\pm$0.5  & 80.1$\pm$1.1  & 62.5$\pm$0.9  & 75.2$\pm$0.7  & 77.8$\pm$0.7  \\
        Minimax~\cite{minimax}    & 58.2$\pm$0.9  & 76.9$\pm$0.8  & 81.1$\pm$0.3  & 63.2$\pm$1.0  & 78.2$\pm$0.7  & 81.5$\pm$1.0  & 64.9$\pm$0.6  & 78.1$\pm$0.6  & 81.3$\pm$0.7  \\
        DiT-IGD~\cite{dit}    & 61.9$\pm$1.9  & 80.9$\pm$0.9  & 84.5$\pm$0.7  & 66.5$\pm$1.1  & 81.0$\pm$1.2  & 85.2$\pm$0.8  & 67.7$\pm$0.3  & 80.4$\pm$0.8  & 84.4$\pm$0.8  \\
        Ours & \textbf{67.0}$\pm$\textbf{0.9}  & \textbf{83.1}$\pm$\textbf{1.0}  & \textbf{86.5}$\pm$\textbf{0.5}  & \textbf{68.0}$\pm$\textbf{0.3}  & \textbf{83.8}$\pm$\textbf{0.6}  & \textbf{86.4}$\pm$\textbf{0.6}  & \textbf{69.1}$\pm$\textbf{1.9}  & \textbf{84.6}$\pm$\textbf{0.4}  & \textbf{85.9}$\pm$\textbf{0.2}  \\
        \midrule
        \multicolumn{10}{c}{Soft Label} \\
        \midrule
        SRe$^2$L~\cite{sre2l} &-&-&-&-&-&-&29.4$\pm$3.0&40.9$\pm$0.3&50.2$\pm$0.4 \\
        RDED~\cite{rded} &63.5$\pm$0.6&84.3$\pm$0.3&89.2$\pm$0.7&60.8$\pm$0.5&80.5$\pm$0.3&89.3$\pm$0.6&61.4$\pm$0.4&80.4$\pm$0.4&89.6$\pm$1.0 \\
        D$^4$M~\cite{d4m} &53.5$\pm$0.5&84.4$\pm$0.4&89.6$\pm$0.2&56.2$\pm$0.3&84.7$\pm$0.5&90.2$\pm$0.3&57.4$\pm$0.4&84.8$\pm$0.2&90.4$\pm$0.7 \\
        DDPS$^c$~\cite{DDPS}&-&-&-&-&-&-&62.5$\pm$0.2&83.4$\pm$0.5&90.2$\pm$0.2\\
        DDPS$^s$~\cite{DDPS}&-&-&-&-&-&-&60.4$\pm$0.3&85.8$\pm$0.4&91.6$\pm$0.4\\
        DiT-IGD*~\cite{igd} &69.6$\pm$1.0&86.7$\pm$0.9&89.9$\pm$0.6&73.6$\pm$1.3&86.8$\pm$1.0&90.6$\pm$0.6&74.8$\pm$0.7&86.4$\pm$0.9&90.7$\pm$0.5 \\
        Ours&\textbf{74.5}$\pm$\textbf{0.3} 
&\textbf{89.1}$\pm$\textbf{0.9} 
&\textbf{91.3}$\pm$\textbf{0.2} 
&\textbf{77.8}$\pm$\textbf{0.8} 
&\textbf{89.7}$\pm$\textbf{0.5} 
&\textbf{91.6}$\pm$\textbf{0.3} 
&\textbf{79.0}$\pm$\textbf{0.3} 
&\textbf{89.3}$\pm$\textbf{0.3} 
&\textbf{92.0}$\pm$\textbf{0.6}
 \\
        % &74.5$\pm$0.3&89.1$\pm$0.9&91.3$\pm$0.2&78.6$\pm$0.8&89.7$\pm$0.5&91.6$\pm$0.3&80.5$\pm$0.3&89.3$\pm$0.3&92.0$\pm$0.6\\
        \midrule
        Full  & \multicolumn{3}{c|}{94.3$\pm$0.5}  & \multicolumn{3}{c|}{94.6$\pm$0.5} & \multicolumn{3}{c}{95.3$\pm$0.6} \\
        \bottomrule
    \end{tabular}}
\end{table}

\paragraph{Implementation details.}
\label{setting}
To ensure fair evaluation, we follow the configurations of IGD~\cite{igd} and EDC~\cite{edc}, maintaining consistency in training procedure and hyperparameter settings.
For the OT components, we set $\alpha_1=1$, $\gamma_1 \in \{1000, 3000\}$, $\alpha_2=0.1$, and $\gamma_2=0.1$. 
For simplicity,
we set $p=1$ ($\ell_1$-norm). %All experiments are performed on RTX 4090 GPUs. 
More details are in Appendix~\ref{a5}. 

\subsection{Results and Discussions}
\paragraph{Results on ImageNet-1K.}
We extensively evaluated our generative OT framework on ImageNet-1K~\cite{imagenet}, comparing it against state-of-the-art dataset distillation methods, including both generative model-based and model-inversion-based approaches, across various architectures and IPC settings.
Table~\ref{tab:imagenet_comparison} presents results on ResNet-18~\cite{resnet}. %At IPC=10, our method achieves 52.9\% accuracy at 300 epochs, significantly outperforming prior methods.
Our method significantly outperforms prior methods at 300 epochs.
%Notably, 
When training is extended to 1000 epochs, performance further improves. This shows that our distilled images and soft labels contain sufficient information for continued optimization. 
%This result indicates that the synthetic dataset not only captures essential structural and semantic information but also retains enough diversity and fidelity, allowing the model to further refine its decision boundaries with prolonged training. 
%A similar trend is observed at IPC=50, where accuracy improves from 61.9\% at 300 epochs to 64.2\% at 1000 epochs, though with a smaller margin, suggesting that as IPC increases, most of the distribution is already well covered, reducing the benefit of additional training.
Beyond ResNet, we evaluated generalization on MobileNet-V2~\cite{mobilenet}, EfficientNet-B0~\cite{efficientnet}, Swin Transformer~\cite{SWIN}, and ConvNeXt~\cite{convnext} (Table~\ref{tab:imagenet_generalization}). Our framework consistently surpasses prior approaches across all architectures. The larger performance gain at lower IPC settings highlights its ability to better preserve fine-grained distributional details. When IPC is low, existing dataset distillation methods struggle to cover the full data distribution, leading to significant discrepancies between the learned distribution and the real distribution. In contrast, our approach explicitly aligns the latent space distribution, logit-level semantic consistency, and label-image relationships, ensuring that even with limited synthetic samples, our distilled set comprehensively represents the real data. %In contrast, our approach explicitly aligns latent space distributional structure, logit-level semantic consistency, and label-image relationships, ensuring that even with limited synthetic samples, our distilled set provides a comprehensive representation of the real data.

\begin{table}[t]
\centering
% ImageNet-100 表格
\begin{minipage}[t]{0.59\textwidth}
\centering
\caption{Performance comparison on ImageNet-100~\cite{idc}.}
\label{img100}
\resizebox{\textwidth}{!}{
\begin{tabular}{ll|cccc}
\toprule
Model      & IPC & SRe$^2$L~\cite{sre2l} & RDED~\cite{rded} & DELT~\cite{delt} & Ours \\
\midrule
         \multirow{3}*{ResNet-18}  & 10  & 9.5$\pm$0.4 & 36.0$\pm$0.3 & 28.2$\pm$1.5 & \textbf{47.7}$\pm$\textbf{0.3} \\
                   & 50  & 27.0$\pm$0.4 & 61.6$\pm$0.1 & 67.9$\pm$0.6 & \textbf{72.6}$\pm$\textbf{0.1} \\
                   & 100 & 30.4$\pm$0.3 & 74.5$\pm$0.4 & 75.1$\pm$0.2 & \textbf{79.2}$\pm$\textbf{0.1} \\
        \cmidrule{1-6}
        \multirow{3}*{ResNet-101} & 10  & 6.4$\pm$0.1 & 33.9$\pm$0.1 & 22.4$\pm$3.3 & \textbf{36.3}$\pm$\textbf{0.5} \\
                   & 50  & 25.7$\pm$0.3 & 66.0$\pm$0.6 & 70.8$\pm$2.3 & \textbf{74.3}$\pm$\textbf{0.2} \\
                   & 100 & 27.6$\pm$0.2 & 73.5$\pm$0.8 & 77.6$\pm$1.8 & \textbf{81.6}$\pm$\textbf{0.1} \\
        \cmidrule{1-6}
       \multirow{3}*{MobileNet} & 10 & 4.5$\pm$0.4 & 23.6$\pm$0.7 & 15.8$\pm$0.2 & \textbf{43.2}$\pm$\textbf{0.2} \\
                   & 50  & 18.4$\pm$0.2 & 51.5$\pm$0.8 & 55.0$\pm$1.8 & \textbf{69.5}$\pm$\textbf{0.3} \\
                   & 100 & 22.1$\pm$0.3 & 70.8$\pm$1.1 & 76.7$\pm$0.3 & \textbf{78.0}$\pm$\textbf{0.2}  \\
\bottomrule
\end{tabular}
}
\end{minipage}
\hfill
% CIFAR-100 表格
\begin{minipage}[t]{0.40\textwidth}
\centering
\caption{Performance comparison on CIFAR-100~\cite{cifar} using ConvNet-3~\cite{convnet}.}
\label{tab:cifar}
\resizebox{\textwidth}{!}{
\begin{tabular}{l|ccc}
\toprule
IPC & 10 & 50 & 100  \\
\midrule
DM~\cite{dm}       & 29.7$\pm$0.3 & 43.6$\pm$0.4 & 47.1$\pm$0.4  \\
M3D~\cite{m3d} &42.4$\pm$0.2&50.9$\pm$0.7&52.1$\pm$0.6 \\
DATM~\cite{datm}     & 47.2$\pm$0.4 & 55.0$\pm$0.2 & 57.5$\pm$0.2  \\
\midrule
SRe$^2$L~\cite{sre2l}  & 24.5$\pm$0.4 & 45.2$\pm$0.3 & 46.6$\pm$0.5  \\
RDED~\cite{rded}     & 48.1$\pm$0.3 & 57.0$\pm$0.1 & 58.1$\pm$0.4 \\
\midrule
D$^4$M~\cite{d4m} &45.0$\pm$0.1&48.8$\pm$0.3& 50.3$\pm$0.2 \\
DiT-IDG~\cite{igd}  & 45.8$\pm$0.6 & 53.9$\pm$0.6 & 55.9$\pm$0.4  \\
Ours &\textbf{50.7}$\pm$\textbf{0.2}&\textbf{57.5}$\pm$\textbf{0.3}&\textbf{58.7}$\pm$\textbf{0.2} \\
\bottomrule
\end{tabular}
}
\end{minipage}
\end{table}
\begin{table}[t]
\begin{minipage}[t]{0.655\textwidth}
    \centering
    \small
    \caption{Ablation Study on ImageNette~\cite{nette} under IPC=10. \textit{Note}: Here, ``w/o LIA'' denotes soft relabeling with the teacher ensemble from high-IPC settings, without adapting to the current IPC.}
    \label{ablation}
    \resizebox{\textwidth}{!}{
    \begin{tabular}{c|cc|cccc}
    \toprule
        \multirow{2}*{Model} & \multicolumn{2}{c|}{Hard Label} & \multicolumn{4}{c}{Soft Label} \\
        & w/o OTG & w OTG &w/o OTG&w/o LIA&w/o OTM & Full\\
        \midrule
        ConvNet-6 &61.9 &\bf67.0&72.5&74.3&73.2&\bf74.5 \\ 
        ResNetAP-10 &66.5 &\bf68.0&74.2&76.4&75.9&\bf77.8 \\ 
        ResNet-18 &67.7 &\bf69.1&77.2&77.8&77.5&\bf79.0 \\ 
        \bottomrule
    \end{tabular}}
\end{minipage}
\hfill
\begin{minipage}[t]{0.325\textwidth}
    \centering
    \small
    \caption{Mean runtime per class (sampling) or per epoch (matching) on ImageNet-1K~\cite{imagenet}.}
    \label{time1}
    \resizebox{\textwidth}{!}{
    \begin{tabular}{c|c|cc}
    \toprule
    Stage& Method & IPC=10 & IPC=50 \\
    \midrule
   \multirow{2}*{Samp.}& w/o OTG &97.1s&537.4s \\
  &  w OTG &97.7s&540.3s \\
    \midrule 
    \multirow{2}*{Match.} & w/o OTM &23.2s &126.1s \\
  &  w OTM &23.3s &126.6s \\
    \bottomrule
    \end{tabular}}
\end{minipage}
\end{table}

\begin{table}[!ht]
\begin{minipage}[t]{0.565\textwidth}
    \caption{Distilled set generation time (IPC=10, ImageNet-1K, 8×4090). PreS: Presample, PostS: Postsample.}
    \label{Runtime2}
    \centering
    \small
    \resizebox{\textwidth}{!}{
    \begin{tabular}{c|c}
    \toprule
       EDC~\cite{edc}  & 3h PreS +3h PostS + 5h Recover + 0.4h Relabel\\
        Ours & 3.4h Diffusion Sample +  0.3h Relabel \\
         \bottomrule
    \end{tabular}}
\end{minipage}
\hfill
\begin{minipage}[t]{0.43\textwidth}
   \caption{Effect of $\alpha$ on ImageNette~\cite{nette}.}
   \vspace{0.2em}
    \label{alpha}
    \centering
    \small
    \resizebox{\textwidth}{!}{
    \begin{tabular}{c|c|c|c}
    \toprule
        Teachers & ResNet-18 & w/o LIA & w LIA\\
        $\alpha$ &0.906 &0.903&\bf0.643  \\
        Avg. Acc. &76.0 &76.2 &\bf77.1 \\
         \bottomrule
    \end{tabular} }
\end{minipage}
\end{table}

\paragraph{Results on ImageNet subsets.} %To further compare with previous works and evaluate our method on a smaller number of categories, we conduct experiments using two subsets of ImageNet with different classes. %consisting of 10 and 100 classes. 
% To further compare with prior works and evaluate our method under reduced category settings, we conduct experiments on two ImageNet subsets with different class selections and numbers of classes.
To further compare with prior works and to evaluate our method under reduced-category settings, we conduct experiments on two ImageNet subsets, varying both class selection and class count.
As shown in Tables~\ref{nette} and~\ref{img100}, our method consistently outperforms all baselines. Notably, we observe significant performance improvements under both hard label and soft label settings. 
This demonstrates that OT-guided sampling effectively captures fine-grained sample information, contributing to the learning of the new model. During the subsequent OT distance minimization phases, this extracted information is systematically transferred to the new model, resulting in enhanced performance.
% This demonstrates that the individual sample information captured through OT-guided sampling effectively contributes to the new model's learning. Furthermore, 
% the subsequent phases of OT distance minimization 
% further 
% transfers this fine-grained information to the new model, leading to enhanced model performance. 
Robustness tests are conducted in Appendix~\ref{robustness}.

\begin{figure}[tb]
    \centering
    \includegraphics[width=0.98\linewidth]{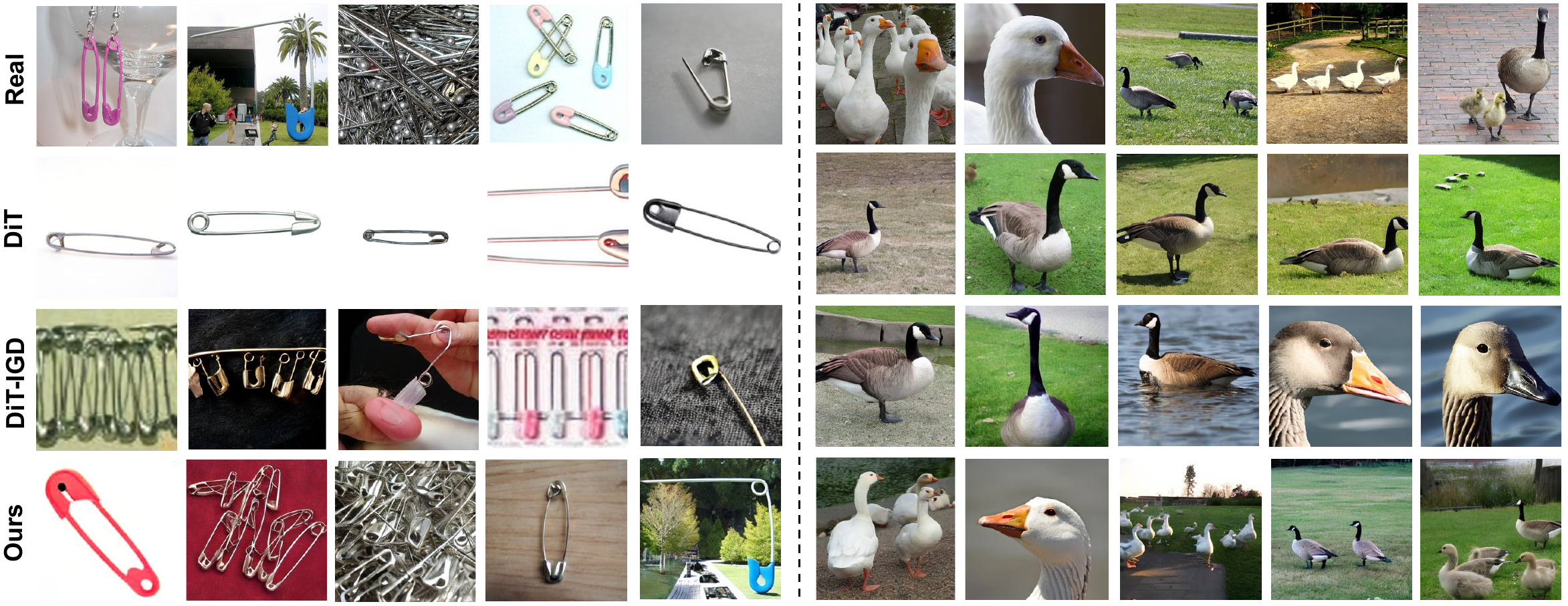}
    \caption{Comparison of generated images from different methods on ImageNet-100 (IPC = 10). 
}
    \label{image}
\end{figure}

\paragraph{Results on CIFAR-100.}
% Using ConvNet-3~\cite{convnet}, we evaluate our method on CIFAR-100~\cite{cifar} in Table~\ref{tab:cifar} to assess its generalizability to low-resolution datasets. 
% We use ConvNet-3~\cite{convnet} to evaluate our method on CIFAR-100~\cite{cifar}, as shown in Table~\ref{tab:cifar}, to assess its generalizability to low-resolution datasets.
We evaluate our method on CIFAR-100~\cite{cifar} to assess its generalizability on low-resolution datasets, as summarized in Table~\ref{tab:cifar}.
To ensure a broad comparison, we include traditional low-resolution-oriented methods, along with model-inversion-based and generative-model-based methods, both specially designed for large-scale datasets. %we consider traditional low-resolution-oriented 
%methods, model-inversion-based approaches, and generative-model-based methods. %As shown in Table~\ref{tab:cifar}, our method consistently outperforms all baselines across different IPC settings.
%, demonstrating its effectiveness in distilling compact yet highly informative synthetic datasets. 
Unlike most existing methods that specialize in either low-resolution or high-resolution datasets, our approach achieves state-of-the-art performance on ImageNet~\cite{imagenet} while maintaining superior results on CIFAR. This further highlights the robustness of our %optimal transport
OT-driven strategy in preserving distributional characteristics across scales. 
%These results establish our method as a unified and scalable solution for dataset distillation across diverse resolutions.

\paragraph{Impact of different components.}
Our OT-guided diffusion sampling effectively transfers the geometric structure of the image space distribution to the distilled images. This alignment is further enhanced by the Label-Image-Aligned Soft Relabeling, which narrows the distributional gap between the distilled and real data. During student model training, the OT-based student logit matching module faithfully propagates this information to the new model. This further reinforces alignment between the original distribution $\mu_{\text{true}}$ and the learned distribution $\nu_{\text{new}}$. As shown in Table~\ref{ablation}, each component involved in minimizing the OT distance plays a critical role, underscoring the necessity of aligning distributions throughout the entire %distillation 
pipeline. More validations are provided in Appendix~\ref{a8}.

\paragraph{Runtime analysis.}
As shown in Table~\ref{time1}, the additional time overhead introduced by our OT constraint is consistently less than 1\%. 
Table~\ref{Runtime2} provides a breakdown of the time required for each step in generating the distilled set for both our method and the state-of-the-art model-inversion-based method, EDC~\cite{edc}. Our approach is notably faster than EDC, which further demonstrates its efficiency.
% Table~\ref{Runtime2} shows time required for each step in generating the distilled set for our method and the state-of-the-art model-inversion-based method, EDC~\cite{edc}. Our approach requires substantially less time than EDC, further highlighting the efficiency of our method.

\paragraph{Discussion of contraction factor $\alpha$.} Table~\ref{alpha} reports the values of $\alpha$ measured under different soft label generation strategies. Our LIA strategy significantly reduces the OT distance between the distilled data and the real data, allowing the distilled data to capture more information of the real distribution. This leads to a substantial performance improvement. More discussion in Appendix~\ref{a7}. %This leads to a substantial performance improvement

\begin{table}[tb]
%\begin{minipage}{0.5\linewidth}
  \caption{Impact of different OT hyperparameters.}
  %\vspace{0.35em}
  \label{hyper}
  \centering
  \small
  % \resizebox{\textwidth}{!}{
    \begin{tabular}{cccc|ccc}
      \toprule
      $\beta_1$ & $\lambda_1$ & $\beta_2$ & $\lambda_2$ & ConvNet & ResNetAP & ResNet \\
      \midrule
      \multicolumn{7}{c}{Hard Label} \\
      \midrule
      1 & 1000 & - & - & 67.0$\pm$0.9 & 68.0$\pm$0.3 & 69.1$\pm$1.9 \\
      1 & 10000 & - & - & 66.3$\pm$0.7 & 68.5$\pm$0.3 & 68.9$\pm$0.8 \\
     10 & 1000 & - & - &65.8$\pm$0.5 &67.5$\pm$0.5 &68.7$\pm$1.1 \\
      \midrule
      \multicolumn{7}{c}{Soft Label} \\
      \midrule
      1 & 1000 & 0.1 & 0.1 & 74.5$\pm$0.3 & 77.8$\pm$0.8 & 79.0$\pm$0.3 \\
      1 & 1000 &0.1 &1 &74.6$\pm$0.8 &76.3$\pm$0.8 &78.2$\pm$1.0 \\
    1 & 1000 &1 &0.1 &74.3$\pm$0.5&78.1$\pm$0.3 &77.4$\pm$0.2 \\
      \bottomrule
    \end{tabular}%}
    \end{table}
%\end{minipage}
% \hspace{0.015\linewidth}
    %\begin{minipage}{0.48\linewidth}  % 右侧：Figure
    %\vskip 0.1in
    \begin{figure}[tb]
        \centering
        \includegraphics[width=0.3\linewidth]{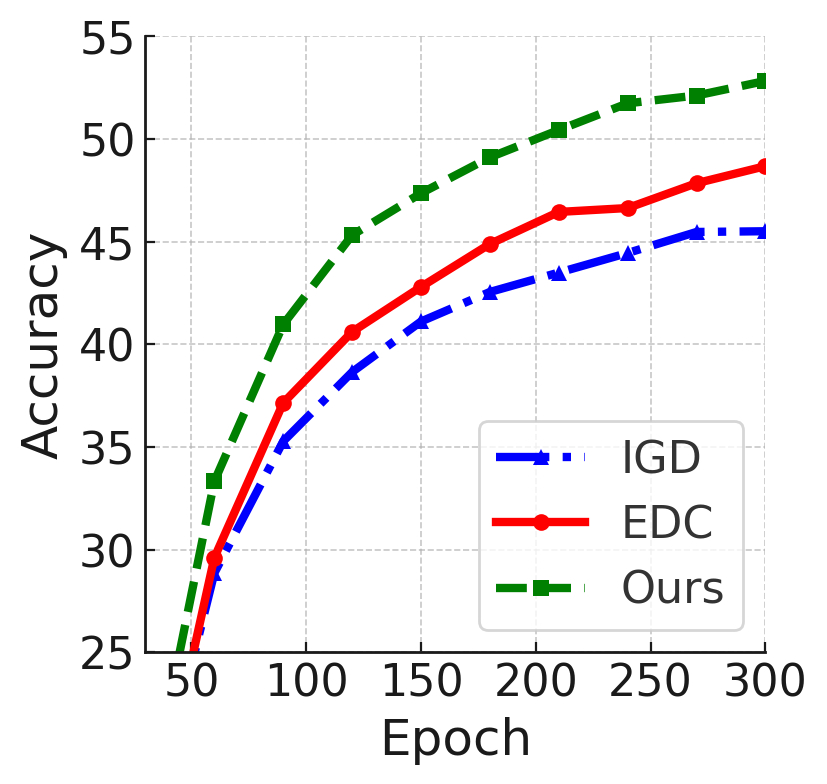}
        \hspace{1em}
        \includegraphics[width=0.3\linewidth]{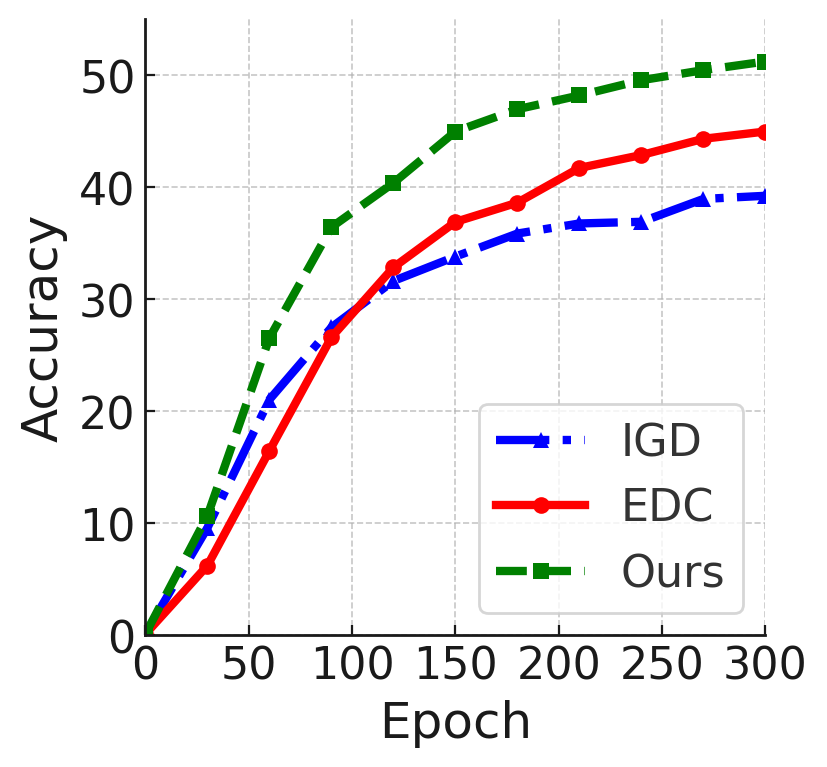}
        % \vspace{-1.3em}
        \caption{Training accuracy curves of ResNet-18 (left) and MobileNet (right). IGD: DiT-IGD.} 
        % \captionof{figure}{Training accuracy curves of ResNet-18 (left) and MobileNet (right). IGD: DiT-IGD.}  % 使用 \captionof{figure}
        %\vskip -0.1in
        \label{compare}
    \end{figure}
    %\end{minipage}
%\end{table}

\textbf{Sensitivity analysis.} 
As shown in Table~\ref{hyper}, our method delivers consistently high accuracy over a broad range of OT-related hyperparameter settings, demonstrating low sensitivity.
This robustness eliminates exhaustive tuning and enables straightforward deployment across diverse scenarios. 
It also makes our approach readily scalable.
%, making our approach readily scalable. 
Additional results and analyses are available in Appendix~\ref{a6}.
 %This robustness removes the need for exhaustive tuning across different scenarios, 
%As shown in Table~\ref{hyper}, our method exhibits robust performance across a wide range of OT hyperparameter values, indicating low sensitivity. This stability %allows a single parameter configuration to be used across different settings
%without the need for extensive tuning, 
% indicates that we do not need to carefully optimize these hyperparameters on different settings, 
% making our approach readily scalable. Additional figures and analyses are provided in Appendix~\ref{a6}.

\paragraph{Visualizations.} Figure~\ref{image} illustrates a qualitative comparison among DiT~\cite{dit}, DiT-IGD~\cite{igd}, and our method. DiT often produces visually similar outputs that lack semantic diversity. %DiT-IGD disregards the underlying data distribution in its attempt to introduce diversity, 
DiT-IGD introduces diversity without aligning with the underlying real data distribution,
leading to non-representative or incorrect generations. Furthermore, its influence estimation is based solely on intraclass averaged statistics, which results in perceptible blurring. In contrast, our approach explicitly models both instance-specific characteristics and fine-grained distributional structures, thereby enabling faithful approximation of the real data manifold.
%Figure~\ref{image} compares generations from DiT~\cite{dit}, DiT-IGD~\cite{igd}, and ours. DiT produces repetitive content with low diversity, while DiT-IGD increases variation at the cost of realism, due to its neglect of the true data distribution and reliance on averaged sample influence—resulting in semantic errors and blur. Our method preserves both instance-level fidelity and distributional detail, achieving closer alignment with real data.
We also present the test accuracy at each logit-matching step in Figure~\ref{compare}. Our method achieves faster convergence and consistently higher accuracy, especially in early epochs, demonstrating superior sample informativeness and stronger distribution alignment when compared to EDC~\cite{edc} and DiT-IGD~\cite{igd}. Please refer to Appendix~\ref{a11} for more visualizations.

% \subsection{Visualizations}
% \setlength{\jot}{15pt}  % Increase 
% \begin{gather*}
% \mathrm{W}(\mu_{\text{true}}, \nu_{\text{new}}) \\
% \leq \mathrm{W}(\mu_{\text{true}}, \nu_{\text{distill}}^{(\mathrm{hard})}) \\
% \quad \cdot \alpha(\nu_{\text{distill}}^{(\mathrm{soft})}) \\
% + \ \mathrm{W}(\nu_{\text{distill}}^{(\mathrm{soft})}, \nu_{\text{new}}).
% \end{gather*}
\section{Conclusion}
We propose a principled framework for generative large-scale dataset distillation by formulating it as an OT distance minimization problem. Our approach explicitly decomposes the total OT distance into three interpretable components and systematically minimizes each to ensure comprehensive 
distributional alignment. %This enables distilled data to induce behavior in new models that closely matches full-data training, regardless of architecture. 
This allows new models trained on distilled data to behave similarly to models trained on the full dataset, regardless of architecture.
Extensive experiments across diverse datasets and model architectures validate the effectiveness and generalizability %, and efficiency 
 of our method.

\paragraph{Broader Impact.} 
% Our distilled datasets can help generate smaller, more efficient models, reducing computational and storage costs, and have the potential to support efficient learning in federated and continual learning scenarios, promoting data privacy and model adaptation across distributed systems.
Our distilled datasets lower carbon footprints associated with new model training, fostering sustainable AI development. They also enable efficient learning in federated and continual learning scenarios, enhancing data privacy and model adaptation across distributed systems.

\paragraph{Acknowledgement} This work was supported by the GPU cluster built by MCC Lab of Information Science and Technology Institution, USTC, and the Supercomputing Center of the USTC.

\paragraph{Competing Interests} The authors declare no competing interests.

\bibliography{reference_full}

@article{imagenet,
  title  = {Imagenet large scale visual recognition challenge},
  author = {Russakovsky, Olga and Deng, Jia and Su, Hao and Krause, Jonathan and Satheesh, Sanjeev and Ma, Sean and Huang, Zhiheng and Karpathy, Andrej and Khosla, Aditya and Bernstein, Michael and others},
  journal= {International Journal of Computer Vision (IJCV)},
  year   = {2015}
}

@inproceedings{datm,
  title     = {Towards Lossless Dataset Distillation via Difficulty-Aligned Trajectory Matching},
  author    = {Guo, Ziyao and Wang, Kai and Cazenavette, George and Li, Hui and Zhang, Kaipeng and You, Yang},
  booktitle = {International Conference on Learning Representations (ICLR)},
  year      = {2024}
}

@article{ljf,
  title={Neighbor-aware Geodesic Transportation for Neighborhood Refinery},
  author={Luo, Jifei and Cui, Xiao and Yao, Hantao and Xu, Changsheng},
  journal={OpenReview Preprint},
  year={2025},
}

@inproceedings{cui2024exploring,
  title     = {Exploring {GPT-4} Vision for Text-to-Image Synthesis Evaluation},
  author    = {Cui, Xiao and Sun, Qi and Zhou, Wengang and Li, Houqiang},
  booktitle = {Tiny Papers Track at the International Conference on Learning Representations (ICLR)},
  year      = {2024},
  url       = {https://openreview.net/forum?id=xmQoodG82a}
}

@article{cui2021heredity,
  title={Heredity-aware child face image generation with latent space disentanglement},
  author={Cui, Xiao and Zhou, Wengang and Hu, Yang and Wang, Weilun and Li, Houqiang},
  journal={arXiv preprint arXiv:2108.11080},
  year={2021}
}

@article{ddim,
  title  = {Denoising diffusion implicit models},
  author = {Song, Jiaming and Meng, Chenlin and Ermon, Stefano},
  journal= {arXiv preprint arXiv:2010.02502},
  year   = {2020}
}

@article{DDPS,
  title  = {Efficient dataset distillation via diffusion-driven patch selection for improved generalization},
  author = {Zhong, Xinhao and Sun, Shuoyang and Gu, Xulin and Xu, Zhaoyang and Wang, Yaowei and Zhang, Min and Chen, Bin},
  journal= {arXiv preprint arXiv:2412.09959},
  year   = {2024}
}

@inproceedings{sre2l,
  title     = {Squeeze, recover and relabel: Dataset condensation at imagenet scale from a new perspective},
  author    = {Yin, Zeyuan and Xing, Eric and Shen, Zhiqiang},
  booktitle = {Conference on Neural Information Processing Systems (NeurIPS)},
  year      = {2023}
}

@inproceedings{minimax,
  title     = {Efficient dataset distillation via minimax diffusion},
  author    = {Gu, Jianyang and Vahidian, Saeed and Kungurtsev, Vyacheslav and Wang, Haonan and Jiang, Wei and You, Yang and Chen, Yiran},
  booktitle = {IEEE Conference on Computer Vision and Pattern Recognition (CVPR)},
  year      = {2024}
}

@article{cifar,
  title    = {Learning multiple layers of features from tiny images},
  author   = {Krizhevsky, Alex and Hinton, Geoffrey and others},
  journal  = {Technical Report, University of Toronto},
  year     = {2009},
  publisher= {Toronto, ON, Canada}
}

@inproceedings{delt,
  title     = {DELT: A Simple Diversity-driven EarlyLate Training for Dataset Distillation},
  author    = {Shen, Zhiqiang and Sherif, Ammar and Yin, Zeyuan and Shao, Shitong},
  booktitle = {IEEE Conference on Computer Vision and Pattern Recognition (CVPR)},
  year      = {2025}
}

@inproceedings{mttnew,
  title     = {Towards Stable and Storage-efficient Dataset Distillation: Matching Convexified Trajectory},
  author    = {Zhong, Wenliang and Tang, Haoyu and Zheng, Qinghai and Xu, Mingzhu and Hu, Yupeng and Nie, Liqiang},
  booktitle = {IEEE Conference on Computer Vision and Pattern Recognition (CVPR)},
  year      = {2025}
}

@article{ncf,
  title  = {Dataset Distillation with Neural Characteristic Function: A Minmax Perspective},
  author = {Wang, Shaobo and Yang, Yicun and Liu, Zhiyuan and Sun, Chenghao and Hu, Xuming and He, Conghui and Zhang, Linfeng},
  journal= {arXiv preprint arXiv:2502.20653},
  year   = {2025}
}

@inproceedings{d4m,
  title     = {D\^{ }4: Dataset Distillation via Disentangled Diffusion Model},
  author    = {Su, Duo and Hou, Junjie and Gao, Weizhi and Tian, Yingjie and Tang, Bowen},
  booktitle = {IEEE Conference on Computer Vision and Pattern Recognition (CVPR)},
  year      = {2024}
}

@inproceedings{rded,
  title     = {On the diversity and realism of distilled dataset: An efficient dataset distillation paradigm},
  author    = {Sun, Peng and Shi, Bei and Yu, Daiwei and Lin, Tao},
  booktitle = {IEEE Conference on Computer Vision and Pattern Recognition (CVPR)},
  year      = {2024}
}

@article{cda,
  title  = {Dataset distillation in large data era},
  author = {Yin, Zeyuan and Shen, Zhiqiang},
  journal= {Transactions on Machine Learning Research (TMLR)},
  year   = {2024}
}

@inproceedings{edc,
  title     = {Elucidating the Design Space of Dataset Condensation},
  author    = {Shao, Shitong and Zhou, Zikai and Chen, Huanran and Shen, Zhiqiang},
  booktitle = {Conference on Neural Information Processing Systems (NeurIPS)},
  year      = {2024}
}

@inproceedings{gvbsm,
  title     = {Generalized large-scale data condensation via various backbone and statistical matching},
  author    = {Shao, Shitong and Yin, Zeyuan and Zhou, Muxin and Zhang, Xindong and Shen, Zhiqiang},
  booktitle = {IEEE Conference on Computer Vision and Pattern Recognition (CVPR)},
  year      = {2024}
}

@article{wmdd,
  title={Dataset distillation via the wasserstein metric},
  author={Liu, Haoyang and Li, Yuchen and Xing, Tianyu and Dalal, Vivek and Li, Lirong and He, Jun and Wang, Hao},
  journal={arXiv preprint arXiv:2311.18531},
  year={2023}
}

@article{scdd,
  title  = {Self-supervised Dataset Distillation: A Good Compression Is All You Need},
  author = {Zhou, Muxin and Yin, Zeyuan and Shao, Shitong and Shen, Zhiqiang},
  journal= {arXiv preprint arXiv:2404.07976},
  year   = {2024}
}

@article{cvdd,
  title  = {Dataset Distillation via Committee Voting},
  author = {Cui, Jiacheng and Li, Zhaoyi and Ma, Xiaochen and Bi, Xinyue and Luo, Yaxin and Shen, Zhiqiang},
  journal= {arXiv preprint arXiv:2501.07575},
  year   = {2025}
}

@inproceedings{igd,
  title     = {Influence-Guided Diffusion for Dataset Distillation},
  author    = {Chen, Mingyang and Du, Jiawei and Huang, Bo and Wang, Yi and Zhang, Xiaobo and Wang, Wei},
  booktitle = {International Conference on Learning Representations (ICLR)},
  year      = {2025}
}

@article{d3m,
  title  = {One Category One Prompt: Dataset Distillation using Diffusion Models},
  author = {Abbasi, Ali and Shahbazi, Ashkan and Pirsiavash, Hamed and Kolouri, Soheil},
  journal= {arXiv preprint arXiv:2403.07142},
  year   = {2024}
}

@inproceedings{m3d,
  title     = {M3d: Dataset condensation by minimizing maximum mean discrepancy},
  author    = {Zhang, Hansong and Li, Shikun and Wang, Pengju and Zeng, Dan and Ge, Shiming},
  booktitle = {AAAI Conference on Artificial Intelligence (AAAI)},
  volume    = {38},
  number    = {8},
  pages     = {9314--9322},
  year      = {2024}
}

@inproceedings{dm,
  title     = {Dataset condensation with distribution matching},
  author    = {Zhao, Bo and Bilen, Hakan},
  booktitle = {IEEE/CVF Winter Conference on Applications of Computer Vision (WACV)},
  pages     = {6514--6523},
  year      = {2023}
}

@article{idc,
  title  = {Dataset Condensation via Efficient Synthetic-Data Parameterization},
  author = {Kim, Jang-Hyun and Kim, Jinuk and Oh, Seong Joon and Yun, Sangdoo and Song, Hwanjun and Jeong, Joonhyun and Ha, Jung-Woo and Song, Hyun Oh},
  journal= {arXiv preprint arXiv:2205.14959},
  year   = {2022}
}

@inproceedings{mtt,
  title     = {Dataset distillation by matching training trajectories},
  author    = {Cazenavette, George and Wang, Tongzhou and Torralba, Antonio and Efros, Alexei A and Zhu, Jun-Yan},
  booktitle = {IEEE Conference on Computer Vision and Pattern Recognition (CVPR)},
  pages     = {4750--4759},
  year      = {2022}
}

@inproceedings{dc,
  title     = {Dataset Condensation with Gradient Matching},
  author    = {Zhao, Bo and Mopuri, Konda Reddy and Bilen, Hakan},
  booktitle = {International Conference on Learning Representations (ICLR)},
  year      = {2020}
}

@article{dance,
  title  = {DANCE: Dual-View Distribution Alignment for Dataset Condensation},
  author = {Zhang, Hansong and Li, Shikun and Lin, Fanzhao and Wang, Weiping and Qian, Zhenxing and Ge, Shiming},
  journal= {arXiv preprint arXiv:2406.01063},
  year   = {2024}
}

@article{survey,
  title  = {The Evolution of Dataset Distillation: Toward Scalable and Generalizable Solutions},
  author = {Liu, Ping and Du, Jiawei},
  journal= {arXiv preprint arXiv:2502.05673},
  year   = {2025}
}

@inproceedings{idm,
  title     = {Improved distribution matching for dataset condensation},
  author    = {Zhao, Ganlong and Li, Guanbin and Qin, Yipeng and Yu, Yizhou},
  booktitle = {IEEE Conference on Computer Vision and Pattern Recognition (CVPR)},
  pages     = {7856--7865},
  year      = {2023}
}

@article{tdsdm,
  title  = {Real-fake: Effective training data synthesis through distribution matching},
  author = {Yuan, Jianhao and Zhang, Jie and Sun, Shuyang and Torr, Philip and Zhao, Bo},
  journal= {arXiv preprint arXiv:2310.10402},
  year   = {2023}
}

@article{mnist,
  title  = {Gradient-based learning applied to document recognition},
  author = {LeCun, Yann and Bottou, L{\'e}on and Bengio, Yoshua and Haffner, Patrick},
  journal= {Proceedings of the IEEE (Proc. IEEE)},
  volume = {86},
  number = {11},
  pages  = {2278--2324},
  year   = {1998}
}

@inproceedings{dwa,
  title     = {Diversity-Driven Synthesis: Enhancing Dataset Distillation through Directed Weight Adjustment},
  author    = {Du, Jiawei and Zhang, Xin and Hu, Juncheng and Huang, Wenxin and Zhou, Joey Tianyi},
  booktitle = {Conference on Neural Information Processing Systems (NeurIPS)},
  year      = {2024}
}

@article{tsne,
  title  = {Visualizing data using {t-SNE}.},
  author = {Van der Maaten, Laurens and Hinton, Geoffrey},
  journal= {Journal of Machine Learning Research (JMLR)},
  volume = {9},
  number = {11},
  pages  = {2579--2605},
  year   = {2008}
}

@inproceedings{dit,
  title     = {Scalable diffusion models with transformers},
  author    = {Peebles, William and Xie, Saining},
  booktitle = {IEEE Conference on Computer Vision and Pattern Recognition (CVPR)},
  pages     = {4195--4205},
  year      = {2023}
}

@misc{agnostic,
  title         = {Towards Model-Agnostic Dataset Condensation by Heterogeneous Models},
  author        = {Moon, Jun-Yeong and Kim, Jung Uk and Park, Gyeong-Moon},
  howpublished  = {arXiv preprint arXiv:2409.14538},
  year          = {2024},
  url           = {https://arxiv.org/abs/2409.14538}
}

@inproceedings{glad,
  title     = {Generalizing Dataset Distillation via Deep Generative Prior},
  author    = {Cazenavette, George and Wang, Tongzhou and Torralba, Antonio and Efros, Alexei A and Zhu, Jun-Yan},
  booktitle = {IEEE Conference on Computer Vision and Pattern Recognition (CVPR)},
  year      = {2023}
}

@article{convnet,
  title  = {An introduction to convolutional neural networks},
  author = {O'Shea, Keiron and Nash, Ryan},
  journal= {arXiv preprint arXiv:1511.08458},
  year   = {2015}
}

@article{nette,
  title  = {A smaller subset of 10 easily classified classes from imagenet, and a little more french},
  author = {Howard, Jeremy},
  journal= {Project URL: https://github.com/fastai/imagenette},
  volume = {4},
  year   = {2019}
}

@article{wd,
  title  = {The wasserstein distances},
  author = {Villani, C{\'e}dric and Villani, C{\'e}dric},
  journal= {Monograph: Optimal Transport: Old and New},
  pages  = {93--111},
  year   = {2009},
  publisher={Springer}
}

@article{tnnls3,
  title  = {An optimal transport analysis on generalization in deep learning},
  author = {Zhang, Jingwei and Liu, Tongliang and Tao, Dacheng},
  journal= {IEEE Transactions on Neural Networks and Learning Systems (TNNLS)},
  volume = {34},
  number = {6},
  pages  = {2842--2853},
  year   = {2021}
}

@inproceedings{wgan1,
  title     = {Wasserstein generative adversarial networks},
  author    = {Arjovsky, Martin and Chintala, Soumith and Bottou, L{\'e}on},
  booktitle = {International Conference on Machine Learning (ICML)},
  pages     = {214--223},
  year      = {2017}
}

@article{wgan2,
  title  = {Improved training of wasserstein gans},
  author = {Gulrajani, Ishaan and Ahmed, Faruk and Arjovsky, Martin and Dumoulin, Vincent and Courville, Aaron C.},
  journal= {Conference on Neural Information Processing Systems (NeurIPS)},
  volume = {30},
  year   = {2017}
}

@article{wgan3,
  title  = {Renormalization group flow, optimal transport, and diffusion-based generative model},
  author = {Sheshmani, Artan and You, Yi-Zhuang and Buyukates, Baturalp and Ziashahabi, Amir and Avestimehr, Salman},
  journal= {Physical Review E (Phys. Rev. E)},
  volume = {111},
  number = {1},
  pages  = {015304},
  year   = {2025}
}

@inproceedings{resnet,
  title     = {Deep residual learning for image recognition},
  author    = {He, Kaiming and Zhang, Xiangyu and Ren, Shaoqing},
  booktitle = {IEEE Conference on Computer Vision and Pattern Recognition (CVPR)},
  pages     = {770--778},
  year      = {2016},
  publisher = {IEEE},
  month     = {Jun.},
  address   = {Las Vegas, NV, USA}
}

@inproceedings{dream,
  title     = {Dream: Efficient dataset distillation by representative matching},
  author    = {Liu, Yanqing and Gu, Jianyang and Wang, Kai and Zhu, Zheng and Jiang, Wei and You, Yang},
  booktitle = {IEEE International Conference on Computer Vision (ICCV)},
  pages     = {17314--17324},
  year      = {2023}
}

@inproceedings{mobilenet,
  title     = {MobileNetV2: Inverted Residuals and Linear Bottlenecks},
  author    = {Sandler, Mark and Howard, Andrew G. and Zhu, Menglong and Zhmoginov, Andrey and Chen, Liang-Chieh},
  booktitle = {IEEE Conference on Computer Vision and Pattern Recognition (CVPR)},
  pages     = {4510--4520},
  year      = {2018},
  publisher = {IEEE},
  month     = {Jun.},
  address   = {Salt Lake City, UT, USA}
}

@article{dd_begin,
  title  = {Dataset distillation},
  author = {Wang, Tongzhou and Zhu, Jun-Yan and Torralba, Antonio and Efros, Alexei A},
  journal= {arXiv preprint arXiv:1811.10959},
  year   = {2018}
}

@article{dd_comprehensive_review,
  title  = {Dataset Distillation: A Comprehensive Review},
  author = {Yu, Ruonan and Liu, Songhua and Wang, Xinchao},
  journal= {arXiv preprint arXiv:2301.07014},
  year   = {2023}
}

@inproceedings{transfer,
  title     = {Dreaming to distill: Data-free knowledge transfer via deepinversion},
  author    = {Yin, Hongxu and Molchanov, Pavlo and Alvarez, Jose M. and Li, Zhizhong and Mallya, Arun and Hoiem, Derek and Jha, Niraj K. and Kautz, Jan},
  booktitle = {IEEE Conference on Computer Vision and Pattern Recognition (CVPR)},
  pages     = {8715--8724},
  year      = {2020}
}

@article{layoutenc,
  title={LayoutEnc: Leveraging Enhanced Layout Representations for Transformer-based Complex Scene Synthesis},
  author={Cui, Xiao and Sun, Qi and Wang, Min and Li, Li and Zhou, Wengang and Li, Houqiang},
  journal={ACM Transactions on Multimedia Computing, Communications and Applications},
  year={2025},
  publisher={ACM New York, NY}
}

@article{continual,
  title  = {An efficient dataset condensation plugin and its application to continual learning},
  author = {Yang, Enneng and Shen, Li and Wang, Zhenyi and Liu, Tongliang and Guo, Guibing},
  journal= {Conference on Neural Information Processing Systems (NeurIPS)},
  volume = {36},
  year   = {2023}
}

@article{ddrobust,
  title  = {Dd-robustbench: An adversarial robustness benchmark for dataset distillation},
  author = {Wu, Yifan and Du, Jiawei and Liu, Ping and Lin, Yuewei and Xu, Wei and Cheng, Wenqing},
  journal= {IEEE Transactions on Image Processing (TIP)},
  year   = {2025}
}

@article{kim2020torchattacks,
  title  = {Torchattacks: A pytorch repository for adversarial attacks},
  author = {Kim, Hoki},
  journal= {arXiv preprint arXiv:2010.01950},
  year   = {2020}
}

@inproceedings{continual2,
  title     = {Summarizing stream data for memory-constrained online continual learning},
  author    = {Gu, Jianyang and Wang, Kai and Jiang, Wei and You, Yang},
  booktitle = {AAAI Conference on Artificial Intelligence (AAAI)},
  volume    = {38},
  number    = {11},
  pages     = {12217--12225},
  year      = {2024}
}

@inproceedings{federate,
  title     = {Unlocking the potential of federated learning: The symphony of dataset distillation via deep generative latents},
  author    = {Jia, Yuqi and Vahidian, Saeed and Sun, Jingwei and Zhang, Jianyi and Kungurtsev, Vyacheslav and Gong, Neil Zhenqiang and Chen, Yiran},
  booktitle = {European Conference on Computer Vision (ECCV)},
  pages     = {18--33},
  year      = {2024},
  organization={Springer}
}

@inproceedings{federate2,
  title     = {Adaptive Backdoor Attacks Against Dataset Distillation for Federated Learning},
  author    = {Chai, Ze and Gao, Zhipeng and Lin, Yijing and Zhao, Chen and Yu, Xinlei and Xie, Zhiqiang},
  booktitle = {IEEE International Conference on Communications (ICC)},
  pages     = {4614--4619},
  year      = {2024}
}

@article{insight,
  title  = {Dataset Distillers Are Good Label Denoisers In the Wild},
  author = {Cheng, Lechao and Chen, Kaifeng and Li, Jiyang and Tang, Shengeng and Zhang, Shufei and Wang, Meng},
  journal= {arXiv preprint arXiv:2411.11924},
  year   = {2024}
}

@inproceedings{insight2,
  title     = {Rethinking data distillation: Do not overlook calibration},
  author    = {Zhu, Dongyao and Lei, Bowen and Zhang, Jie and Fang, Yanbo and Xie, Yiqun and Zhang, Ruqi and Xu, Dongkuan},
  booktitle = {IEEE Conference on Computer Vision and Pattern Recognition (CVPR)},
  pages     = {4935--4945},
  year      = {2023}
}

@inproceedings{efficientnet,
  title     = {Efficientnet: Rethinking model scaling for convolutional neural networks},
  author    = {Tan, Mingxing and Le, Quoc},
  booktitle = {International Conference on Machine Learning (ICML)},
  pages     = {6105--6114},
  year      = {2019},
  organization={PMLR}
}

@inproceedings{SWIN,
  title     = {Swin transformer: Hierarchical vision transformer using shifted windows},
  author    = {Liu, Ze and Lin, Yutong and Cao, Yue and Hu, Han and Wei, Yixuan and Zhang, Zheng and Lin, Stephen and Guo, Baining},
  booktitle = {IEEE International Conference on Computer Vision (ICCV)},
  pages     = {10012--10022},
  year      = {2021}
}

@inproceedings{convnext,
  title     = {A convnet for the 2020s},
  author    = {Liu, Zhuang and Mao, Hanzi and Wu, Chao-Yuan and Feichtenhofer, Christoph and Darrell, Trevor and Xie, Saining},
  booktitle = {IEEE Conference on Computer Vision and Pattern Recognition (CVPR)},
  pages     = {11976--11986},
  year      = {2022}
}

@inproceedings{un1,
  title     = {Integrating efficient optimal transport and functional maps for unsupervised shape correspondence learning},
  author    = {Le, Tung and Nguyen, Khai and Sun, Shanlin and Ho, Nhat and Xie, Xiaohui},
  booktitle = {IEEE Conference on Computer Vision and Pattern Recognition (CVPR)},
  pages     = {23188--23198},
  year      = {2024}
}

@article{un2,
  title  = {Enhancing Unsupervised Graph Few-shot Learning via Set Functions and Optimal Transport},
  author = {Liu, Yonghao and Giunchiglia, Fausto and Li, Ximing and Huang, Lan and Feng, Xiaoyue and Guan, Renchu},
  journal= {arXiv preprint arXiv:2501.05635},
  year   = {2025}
}

@article{un4,
  title  = {Unsupervised Anomaly Detection through Mass Repulsing Optimal Transport},
  author = {Montesuma, Eduardo Fernandes and Habazi, Adel El and Mboula, Fred Ngole},
  journal= {arXiv preprint arXiv:2502.12793},
  year   = {2025}
}

@article{causal1,
  title  = {On the Role of Entropy-Based Loss for Learning Causal Structure With Continuous Optimization},
  author = {Chen, Weilin and Qiao, Jie and Cai, Ruichu and Hao, Zhifeng},
  journal= {IEEE Transactions on Neural Networks and Learning Systems (TNNLS)},
  year   = {2023}
}

@article{causal,
  title  = {Conceptual Learning and Causal Reasoning for Semantic Communication},
  author = {Wheeler, Dylan and Natarajan, Balasubramaniam},
  journal= {IEEE Transactions on Cognitive Communications and Networking (TCCN)},
  year   = {2025}
}

@inproceedings{zeng2024hierarchical,
  title     = {Hierarchical multi-marginal optimal transport for network alignment},
  author    = {Zeng, Zhichen and Du, Boxin and Zhang, Si and Xia, Yinglong and Liu, Zhining and Tong, Hanghang},
  booktitle = {AAAI Conference on Artificial Intelligence (AAAI)},
  volume    = {38},
  number    = {15},
  pages     = {16660--16668},
  year      = {2024}
}

@article{re1,
  title  = {Wasserstein Adaptive Value Estimation for Actor-Critic Reinforcement Learning},
  author = {Baheri, Ali and Sharooei, Zahra and Salgarkar, Chirayu},
  journal= {arXiv preprint arXiv:2501.10605},
  year   = {2025}
}

@article{jin2025semi,
  title  = {Semi-discrete optimal transport for long-tailed classification},
  author = {Jin, Lian-Bao and Lei, Na and Luo, Zhong-Xuan and Wu, Jin and Ai, Chao and Gu, Xianfeng},
  journal= {Journal of Computer Science and Technology (JCST)},
  volume = {40},
  number = {1},
  pages  = {252--266},
  year   = {2025}
}

@article{nguyen2025lightspeed,
  title  = {Lightspeed Geometric Dataset Distance via Sliced Optimal Transport},
  author = {Nguyen, Khai and Nguyen, Hai and Pham, Tuan and Ho, Nhat},
  journal= {arXiv preprint arXiv:2501.18901},
  year   = {2025}
}

@article{da2,
  title  = {Domain Adaptation and Entanglement: an Optimal Transport Perspective},
  author = {Koç, Okan and Soen, Alexander and Chiang, Chao-Kai and Sugiyama, Masashi},
  journal= {arXiv preprint arXiv:2503.08155},
  year   = {2025}
}

@article{re2,
  title  = {On the benefit of optimal transport for curriculum reinforcement learning},
  author = {Klink, Pascal and D'Eramo, Carlo and Peters, Jan and Pajarinen, Joni},
  journal= {IEEE Transactions on Pattern Analysis and Machine Intelligence (TPAMI)},
  year   = {2024}
}

@article{re3,
  title  = {Sample efficient deep reinforcement learning with online state abstraction and causal transformer model prediction},
  author = {Lan, Yixing and Xu, Xin and Fang, Qiang and Hao, Jianye},
  journal= {IEEE Transactions on Neural Networks and Learning Systems (TNNLS)},
  year   = {2023}
}

@article{re4,
  title  = {CVaR-Constrained Policy Optimization for Safe Reinforcement Learning},
  author = {Zhang, Qiyuan and Leng, Shu and Ma, Xiaoteng and Liu, Qihan and Wang, Xueqian and Liang, Bin and Liu, Yu and Yang, Jun},
  journal= {IEEE Transactions on Neural Networks and Learning Systems (TNNLS)},
  year   = {2024}
}

@article{sinkhornbase,
  title  = {Sinkhorn distances: Lightspeed computation of optimal transport},
  author = {Cuturi, Marco},
  journal= {Conference on Neural Information Processing Systems (NeurIPS)},
  volume = {26},
  year   = {2013}
}

@article{sink,
  title  = {Multi-Level Optimal Transport for Universal Cross-Tokenizer Knowledge Distillation on Language Models},
  author = {Cui, Xiao and Zhu, Mo and Qin, Yulei and Xie, Liang and Zhou, Wengang and Li, Houqiang},
  journal= {AAAI Conference on Artificial Intelligence (AAAI)},
  year   = {2025}
}

@inproceedings{yin2020dreaming,
  title     = {Dreaming to distill: Data-free knowledge transfer via deepinversion},
  author    = {Yin, Hongxu and Molchanov, Pavlo and Alvarez, Jose M. and Li, Zhizhong and Mallya, Arun and Hoiem, Derek and Jha, Niraj K. and Kautz, Jan},
  booktitle = {IEEE Conference on Computer Vision and Pattern Recognition (CVPR)},
  pages     = {8715--8724},
  year      = {2020}
}

@inproceedings{optical,  
title={OPTICAL: Leveraging Optimal Transport for Contribution Allocation in Dataset Distillation},
  author={Cui, Xiao and Qin, Yulei and Zhou, Wengang and Li, Hongsheng and Li, Houqiang},
  booktitle = {IEEE Conference on Computer Vision and Pattern Recognition (CVPR)},
  year={2025}
}

@article{fkd,
  title   = {A Fast Knowledge Distillation Framework for Visual Recognition},
  author  = {Shen, Zhiqiang and Xing, Eric},
  journal = {arXiv preprint arXiv:2112.01528},
  year    = {2021},
  url     = {https://arxiv.org/abs/2112.01528}
}

@inproceedings{smith2021always,
  title     = {Always be dreaming: A new approach for data-free class-incremental learning},
  author    = {Smith, James and Hsu, Yen-Chang and Balloch, Jonathan and Shen, Yilin and Jin, Hongxia and Kira, Zsolt},
  booktitle = {IEEE International Conference on Computer Vision (ICCV)},
  pages     = {9374--9384},
  year      = {2021}
}

@inproceedings{pmlr-v97-zhang19j,
  title     = {Theoretically Principled Trade-off between Robustness and Accuracy},
  author    = {Zhang, Hongyang and Yu, Yaodong and Jiao, Jiantao and Xing, Eric and El Ghaoui, Laurent and Jordan, Michael},
  booktitle = {International Conference on Learning Representations (ICLR)},
  pages     = {7472--7482},
  year      = {2019}
}

@article{tsuzuku2018lipschitz,
  title  = {Lipschitz-margin training: Scalable certification of perturbation invariance for deep neural networks},
  author = {Tsuzuku, Yusuke and Sato, Issei and Sugiyama, Masashi},
  journal= {Conference on Neural Information Processing Systems (NeurIPS)},
  volume = {31},
  year   = {2018}
}

@inproceedings{zhang2019theoretically,
  title     = {Theoretically principled trade-off between robustness and accuracy},
  author    = {Zhang, Hongyang and Yu, Yaodong and Jiao, Jiantao and Xing, Eric and El Ghaoui, Laurent and Jordan, Michael},
  booktitle = {International Conference on Machine Learning (ICML)},
  pages     = {7472--7482},
  year      = {2019},
  organization={PMLR}
}

@inproceedings{moosavi2019robustness,
  title     = {Robustness May Be at Odds with Accuracy},
  author    = {Moosavi-Dezfooli, Seyed-Mohsen and Fawzi, Alhussein and Frossard, Pascal},
  booktitle = {International Conference on Learning Representations (ICLR)},
  year      = {2019}
}

@article{sink2,
  title  = {SinKD: Sinkhorn Distance Minimization for Knowledge Distillation},
  author = {Cui, Xiao and Qin, Yulei and Gao, Yuting and Zhang, Enwei and Xu, Zihan and Wu, Tong and Li, Ke and Sun, Xing and Zhou, Wengang and Li, Houqiang},
  journal= {IEEE Transactions on Neural Networks and Learning Systems (TNNLS)},
  year   = {2025}
}

@article{cui2025rethinking,
  title={Rethinking Long-tailed Dataset Distillation: A Uni-Level Framework with Unbiased Recovery and Relabeling},
  author={Cui, Xiao and Qin, Yulei and Li, Xinyue and Zhou, Wengang and Li, Hongsheng and Li, Houqiang},
  journal={arXiv preprint arXiv:2511.18858},
  year={2025}
}

@inproceedings{otda,
  title     = {Cross-Domain Offline Policy Adaptation with Optimal Transport and Dataset Constraint},
  author    = {Lyu, Jiafei and Yan, Mengbei and Qiao, Zhongjian and Liu, Runze and Ma, Xiaoteng and Ye, Deheng and Yang, Jing-Wen and Lu, Zongqing and Li, Xiu},
  booktitle = {International Conference on Learning Representations (ICLR)},
  year      = {2025}
}

@inproceedings{dhariwal2021diffusion,
  title     = {Diffusion Models Beat GANs on Image Synthesis},
  author    = {Dhariwal, Prafulla and Nichol, Alexander Quinn},
  booktitle = {Conference on Neural Information Processing Systems (NeurIPS)},
  volume    = {34},
  pages     = {8780--8794},
  year      = {2021}
}

@article{ho2022classifier,
  title  = {Classifier-Free Diffusion Guidance},
  author = {Ho, Jonathan and Salimans, Tim},
  journal= {arXiv preprint arXiv:2207.12598},
  year   = {2022}
}

@article{wang2022zero,
  title  = {Zero-Shot Image Restoration Using Denoising Diffusion Restoration Models},
  author = {Wang, Tuo and Song, Jiaming and Elad, Michael and Ermon, Stefano},
  journal= {arXiv preprint arXiv:2201.11793},
  year   = {2022}
}

@inproceedings{kawar2022denoising,
  title     = {Denoising Diffusion Restoration Models},
  author    = {Kawar, Bahjat and Song, Jiaming and Elad, Michael and Ermon, Stefano},
  booktitle = {Conference on Neural Information Processing Systems (NeurIPS)},
  volume    = {35},
  pages     = {23502--23516},
  year      = {2022}
}

@inproceedings{chung2022improving,
  title     = {Improving Diffusion Models for Inverse Problems using Manifold Constraints},
  author    = {Chung, Hyungjin and Sim, Byeongsu and Ryu, Dohoon and Ye, Jong Chul},
  booktitle = {Conference on Neural Information Processing Systems (NeurIPS)},
  volume    = {35},
  pages     = {27439--27452},
  year      = {2022}
}

@inproceedings{graikos2022diffusion,
  title     = {Diffusion Models as Plug-and-Play Priors},
  author    = {Graikos, Alexandros and Malkin, Nikolay and Jojic, Nebojsa and Samaras, Dimitris},
  booktitle = {Conference on Neural Information Processing Systems (NeurIPS)},
  volume    = {35},
  pages     = {14715--14728},
  year      = {2022}
}

@inproceedings{nair2023steered,
  title     = {Steered Diffusion: A Generalized Framework for Plug-and-Play Conditional Image Generation},
  author    = {Gopalakrishnan Nair, Nithin and Patel, Vishal M.},
  booktitle = {IEEE Conference on Computer Vision and Pattern Recognition (CVPR)},
  pages     = {12345--12354},
  year      = {2023}
}

@inproceedings{yu2023freedom,
  title     = {FreeDoM: Training-Free Energy-Guided Conditional Diffusion Model},
  author    = {Yu, Jiwen and Wang, Wei and Zhang, Xiaobo and Huang, Bo and Wang, Yi},
  booktitle = {IEEE Conference on Computer Vision and Pattern Recognition (CVPR)},
  pages     = {4567--4576},
  year      = {2023}
}

@inproceedings{bansal2023universal,
  title     = {Universal Guidance for Diffusion Models},
  author    = {Bansal, Arpit and Chu, Hong-Min and Schwarzschild, Avi and Sengupta, Soumyadip and Goldblum, Micah and Geiping, Jonas and Goldstein, Tom},
  booktitle = {IEEE Conference on Computer Vision and Pattern Recognition (CVPR)},
  pages     = {123--132},
  year      = {2023}
}

@article{chung2022diffusion,
  title  = {Diffusion Posterior Sampling for General Noisy Inverse Problems},
  author = {Chung, Hyungjin and Kim, Jeongsol and McCann, Michael T. and Klasky, Marc L. and Ye, Jong Chul},
  journal= {arXiv preprint arXiv:2209.14687},
  year   = {2022},
  url    = {https://arxiv.org/abs/2209.14687}
}

\newpage
\appendix
% \section{Appendix}
% In this appendix, we first provide additional information of related works, focusing on optimal transport in \textbf{A.1}. \textbf{A.2} offers the detailed descriptions of the datasets used in our experiments.
% \textbf{A.3} presents a comprehensive table to clarify the symbols utilized throughout the paper.
% In \textbf{A.4}, we include the pseudo-code for our method to facilitate reproducibility.
% \section*{Appendix}
% \addcontentsline{toc}{section}{Appendix}
\noindent\textbf{\Large Appendix}
\section{Overview}

This appendix provides comprehensive supplementary materials to further elaborate on our method’s theoretical foundations, experimental setup, and empirical findings. It includes the following sections:

\begin{itemize}
    \item \textbf{Section~\ref{a1}: More Related Work.} Detailed discussions on prior studies, with an emphasis on optimal transport and guided diffusion sampling

    \item \textbf{Section~\ref{a2}: Dataset Descriptions.} Comprehensive descriptions of all datasets used in our experiments, including ImageNet-1K~\cite{imagenet}, ImageNette~\cite{nette}, ImageNet-100~\cite{idc}, and CIFAR-100~\cite{cifar}.

    \item \textbf{Section~\ref{a3}: Symbol Table.} A complete summary of key mathematical notations, hyperparameters, and definitions referenced throughout the paper.

    \item \textbf{Section~\ref{a4}: Pseudocode.} Step-by-step pseudocode for the proposed pipeline, detailed procedures for OT-alignment in different stages, and the calculation process for the contraction factor $\alpha$.

    \item \textbf{Section~\ref{a5}: Implementation Details.} Full specifications of hyperparameter settings, training schedules, and augmentation strategies across all datasets used in our experiments.

    \item \textbf{Section~\ref{a610}: Further Experimental Analyses.} Additional experiments, including sensitivity studies~(\ref{a6}), in-depth analysis of the contraction factor~(\ref{a7}), comparisons with alternative distance metrics~(\ref{a8}), runtime analysis~(\ref{a9}), data coverage evaluation~(\ref{a10}), expanded comparisons with other baselines~(\ref{dwacompare} and \ref{wmddcompare}), robustness evaluation under adversarial attacks~(\ref{robustness}) and evaluation under extremely low-IPC settings~\ref{evallow}.\\
    \footnotesize{\textit{This section constitutes the core of the appendix, offering deep empirical analyses of the contraction factor $\alpha$ and the robustness properties of the distilled models. Other experiments further strengthen and extend the key findings presented in the main text.}}

    \item \textbf{Section~\ref{a11}: More Visualization Results.} Additional qualitative visualizations, including t-SNE plots and synthesized images, to assess semantic coverage and distributional diversity.

    \item \textbf{Section~\ref{a12}: Limitations.} Critical discussion of the imitations of our framework.

    \item \textbf{Section~\ref{a13}: Broader Impact.} Reflections on the broader societal, ethical, and practical implications of our dataset distillation method.
\end{itemize}

% These supplementary materials provide a complete and transparent view into our method’s motivation, design choices, theoretical justification, and empirical behavior, ensuring full reproducibility and a deeper understanding of the contributions made.
Together, these supplementary materials provide a complete and transparent view of our method, support full reproducibility, and offer additional insights that complement and strengthen the main paper.

\section{More Related Work}
%考虑放进附录
\label{a1}
\paragraph{Optimal transport}
OT theory provides a principled mathematical framework for comparing probability distributions by computing the minimal cost required to transform one distribution into another.
Compared to KL divergence and Jensen-Shannon (JS) divergence, OT provides a more geometrically faithful measure of distributional differences, particularly when dealing with distributions with non-overlapping supports~\cite{wd,tnnls3}. 
The Wasserstein distance, also known as OT distance, effectively quantifies distributional discrepancies and has been widely applied in image generation~\cite{wgan1,wgan2,wgan3}, causal discovery~\cite{causal,causal1}, unsupervised learning~\cite{un1,un2,un4,ljf}, and reinforcement learning~\cite{re1,re2,re3,re4}. However, its exact computation is intractable for high-dimensional data due to prohibitive complexity. To overcome this, the Sinkhorn distance introduces entropy regularization, making OT computation more efficient and numerically stable~\cite{sinkhornbase}. This regularized variant extends OT applications to domain adaptation~\cite{otda,zeng2024hierarchical,da2}, classification~\cite{nguyen2025lightspeed,jin2025semi}, and knowledge distillation~\cite{sink,sink2}. In this work, we propose a generative model-based OT framework designed to achieve precise distributional alignment throughout the dataset distillation process. Our approach optimizes the distilled dataset to minimize the OT distance between any student model's output distribution and the real data distribution, ensuring improved generalization.
\paragraph{Guided diffusion sampling} Guided diffusion sampling enhances the generative capabilities of pre-trained diffusion models by incorporating external guidance during the reverse process to steer generation toward desired semantics~\cite{chung2022diffusion}. Early methods, such as classifier guidance~\cite{dhariwal2021diffusion}, inject gradients from pre-trained classifiers into the sampling process to condition generation. However, this approach necessitates domain-specific classifiers trained on noisy intermediate latents, which are often impractical. Classifier-free guidance~\cite{ho2022classifier} addresses this limitation by training the model with both conditional and unconditional objectives, enabling control without external models. Building upon this, Wang et al.~\cite{wang2022zero} introduced linear operator-based guidance, constraining the diffusion process to the null space of known measurement operators; nonetheless, this strategy faces challenges in generalizing to nonlinear mappings. Subsequent works~\cite{kawar2022denoising, chung2022improving, graikos2022diffusion} extend guidance to inverse problems through iterative optimization and plug-and-play conditioning. Concurrently, methods like those proposed by Gopalakrishnan Nair et al.~\cite{nair2023steered}, Yu et al.~\cite{yu2023freedom}, and Bansal et al.~\cite{bansal2023universal} introduce generic guidance functions by injecting gradients from task-specific losses computed on denoised intermediate states, thereby broadening applicability without necessitating model retraining. Inspired by these methods, Influence-Guided Diffusion (IGD)~\cite{igd} leverages guided diffusion for dataset distillation by modifying the reverse sampling process to generate training-optimal data. However, its reliance on matching global distributional trajectories and introducing diversity through random perturbations often leads to suboptimal alignment, neglecting discriminative yet informative local characteristics in favor of global averaging. To overcome this limitation, we propose an optimal transport-based guidance strategy that explicitly aligns the geometric structure of real and synthetic distributions, achieving fine-grained consistency in guided diffusion sampling.

\nocite{cui2024exploring}

\section{Dataset Description}
\label{a2}

\paragraph{ImageNet-1K}
ImageNet-1K~\cite{imagenet}, also known as the ILSVRC 2012 dataset, is a large-scale image classification benchmark comprising 1,000 object categories. It contains approximately 1.28 million training images, 50,000 validation images, and 100,000 test images. The dataset is organized according to the WordNet hierarchy, with each synset corresponding to a distinct semantic concept. ImageNet-1K has been instrumental in advancing deep learning research and remains a standard benchmark for evaluating image classification models in large-scale settings.

\paragraph{ImageNette}
ImageNette~\cite{nette} is a curated subset of ImageNet, consisting of 10 relatively easy categories, including ``tench'', ``English springer'', ``cassette player'', ``chain saw'', ``church'', ``French horn'', ``garbage truck'', ``gas pump'', ``golf ball'', and ``parachute''. It was introduced to facilitate rapid experimentation and prototyping of image classification models, particularly under limited computational budgets. All images are resized to a resolution of $224\times224$ pixels, providing a lightweight yet meaningful benchmark for distillation and robustness studies.

\paragraph{ImageNet-100}
ImageNet-100~\cite{idc} is another subset derived from ImageNet-1K, comprising 100 randomly selected classes. Each class typically contains around 1,000 training images and 300 test images, maintaining a relatively balanced distribution. ImageNet-100 provides a manageable yet challenging benchmark for evaluating classification performance, especially in scenarios where computational efficiency and rapid iteration are prioritized.

\paragraph{CIFAR-100}
CIFAR-100~\cite{cifar} is a widely used benchmark dataset for image classification, extending the number of classes from 10 (CIFAR-10) to 100. Each class contains 600 images, with all images having a resolution of $32\times32$ pixels. Despite its compact size, CIFAR-100 presents a significant classification challenge due to its high intra-class variability and fine-grained label structure, making it a valuable resource for developing and assessing lightweight classification models.

\section{Symbol Description}
\label{a3}
To enhance clarity, a detailed description of mathematic symbols in the present study is provided in Table~\ref{symbol}.

\begin{table}[htbp]
    \centering
    \caption{Descriptions of all symbols, functions, and hyperparameters introduced in the main paper.}
     \label{symbol}
    \begin{tabular}{c|c}
    \toprule
    Symbol & Definition \\
    \midrule
    $E$  & Encoder to transform image into the latent space \\
    $D$ & Decoder to reconstruct latent code back to the image space \\
    $\mathcal{S}$ &Distilled (synthetic) dataset \\
    $\mathcal{T}$ &Real (full) dataset \\ 
    $\mathbf{x}$ & Image \\
    %$\mathbf{z}$ & Latent code \\
    %$\mathbf{z}^c$ & Latent code of class c \\
    $\mathbf{z}_0$ & Latent code of clean sample \\
    $\mathbf{z}_t$ & Latent code of noisy sample at time step \( t \) \\
    $\mathbf{z}^c_t$ & Latent code of  class-$c$ noisy sample at time step \( t \) \\
    $\alpha_t$ & Noise schedule controlling the perturbation at time step \( t \) \\
    $\bm{\epsilon}$ & Gaussian noise \\
    $\epsilon_{\phi}$ & Denoising function parameterized by \( \phi \) \\
    $s$ & Reverse diffusion update function\\
    $\mathcal{G}$ & Guidance function in guided diffusion \\
    $\mathcal{G}_I$ & Influence function for general alignment \\
    $\mathcal{G}_D$ & Diversity function enforcing diversity in distilled data \\
    $\mathcal{G}_\mathrm{W}$ & Guidance function based on optimal transport \\
    $\mathbf{M}^c_n$ & Previously sampled $n$ latents for class \( c \) \\
    $\hat{\mathbf{M}}_n^c$ & Concatenation of $\mathbf{M}_{n-1}^c$ and the latent $\mathbf{z}^c_t$ under sampling  \\
    $\mathbf{Z}_\mathcal{T}^c$ & Latent representations of class \( c \) from the real dataset \\
    $p$ & Norm order \\
    $\mathbf{P}^{\lambda_1}$ & Optimal transport matrix for guided diffusion sampling with regularization \\
    $\mathbf{D}$ & Latent space cost matrix for guided diffusion sampling \\
    $\mathbf{K}^t$ & Transport matrix at the $t$-th step of Sinkhorn normalization \\
    $T$ & Sinkhorn iterations\\
    $T_D$ & Diffusion denoising iterations \\
$\mathbf{P}^{\lambda_2}$ & Optimal transport matrix for logit matching with regularization \\
    $\mathbf{C}$ & Batch-wise cost matrix for logit matching \\
    $F_t$ & The logit output function of the \(t\)-th teacher \\
    $\mathbf{t}$ & Soft label for a batch\\
    $\mathbf{s}$ & Student model output for a batch\\
    $\mathbf{t}_i$ & Soft label for the $i$-th image in a batch \\
    $\mathbf{y}_{\text{onehot}}(\mathbf{x}_i)$ & One-hot hard label for the $i$-th image in a batch \\
    $b$ & Batch size \\
    $h(\mathbf{P})$ & The entropy of $\mathbf{P}$ \\
    $\nu_{\text{distill}}$ & Distilled data distribution \\
    $\nu_{\text{distill}}^{\text{soft}}$ & Distilled data distribution with soft label\\
    $\nu_{\text{distill}}^{\text{hard}}$ & Distilled data distribution with hard label\\
    $\mu_{\text{true}}$ & Real dataset distribution \\
    $\nu_{\text{new}}$ & Output of the student model after training on the distilled set \\
    $\mathrm{W}(\mu_{\text{true}}, \nu_{\text{new}})$ & Wasserstein distance between real dataset and student model output \\
    $\rho_t$ & Weight for influence function in the reverse sampling process \\
    $\gamma_t$ & Weight for diversity function in the reverse sampling process \\
    $\beta_1$ & Weight for the optimal transport guidance in the reverse sampling process \\
    $\lambda_1$ & Entropy regularization weight for optimal transport matrix \\
    $\lambda_2$ & Entropy regularization weight for logit matching \\
    $\alpha(\nu_{\text{distill}}^{(\mathrm{soft})})$ & Contraction factor quantifying the benefit of soft labels \\
    IPC & Images per class in the dataset \\
    $\mathbb{T}$ & Set of teacher models \\
    $\mathcal{S}_\mathbf{x}$ & Distilled image set \\
    $\kappa_1$ & Weight for cross-entropy loss in logit matching \\
    $\kappa_2$ & Weight for mean squared error loss in logit matching \\
    $\beta_2$ & Weight for Sinkhorn distance loss in logit matching \\
    $\mathcal{L}_{\text{CE}}$ & Cross-entropy loss \\
    $\mathcal{L}_{\text{MSE}}$ & Mean squared error loss \\
    $\mathcal{L}_{\text{SD}}$ & Sinkhorn distance loss \\
    \bottomrule
    \end{tabular}

\end{table}

\newpage
\section{PseudoCode}
\label{a4}
We present the pseudocode for our pipeline in Algorithm~\ref{alg:frame}. The detailed calculation of the optimal transport (OT) distance for OT-guided Diffusion Sampling is provided in Algorithm~\ref{alg:ot_guidance}, while the OT-based Student Model Logit matching is outlined in Algorithm~\ref{alg:ot-ma}. For efficient computation, we approximate the contraction factors using features in the latent space, enabling dimensionality reduction while preserving critical information.

\begin{algorithm}
\caption{OT-based Generative Dataset Distillation Framework}
\label{alg:frame}
\begin{algorithmic}[1]
\REQUIRE Real dataset $\mathcal{T} = \{(\mathbf{x}_i, y_i)\}$, teacher models $\mathbb{T}$, target IPC, diffusion model $G$, encoder $E$, decoder $D$, student model $S$
\ENSURE Distilled dataset $\mathcal{S}_\mathbf{x}$ and trained student model $S$

\FOR{each class $c = 1$ to $C$}
    \STATE Encode real samples: $\mathbf{Z}_\mathcal{T}^c \gets E(\{\mathbf{x}_i : y_i = c\})$
    \FOR{sample index $n = 1$ to IPC}
        \STATE Sample latent $\mathbf{z}^c_{T_D}$ using diffusion model $G$
         \FOR{$t={T_D}$   to 1}
        \STATE Compute OT-guidance $\mathcal{G}_\mathrm{W}(\mathbf{z}^c_t)$ w.r.t. $\mathbf{Z}_\mathcal{T}^c$ and previously sampled latents $\mathbf{M}_{n-1}^c$
        \STATE Update latent using guidance:
        $
        \mathbf{z}_{t-1}^c \gets s(\mathbf{z}_t^c, t, \epsilon_{\phi}) - \rho_t \nabla \mathcal{G}_I - \gamma_t \nabla \mathcal{G}_D - \beta_1 \nabla \mathcal{G}_\mathrm{W}
        $
         \ENDFOR
        \STATE Append $\mathbf{z}_{t-1}^c$ to $\mathbf{M}^c$
    \ENDFOR
\ENDFOR

\STATE Decode all latents: $\mathcal{S}_\mathbf{x} \gets D(\mathbf{M}^c_{\text{IPC}})$ for all $c$
\STATE Select teacher set $\mathbb{T}(\text{IPC})$ according to IPC level
\FOR{each image $\mathbf{x}_i \in \mathcal{S}_\mathbf{x}$}
    
    \STATE Generate soft label: $\mathbf{t}(\mathbf{x}_i) \gets \frac{1}{|\mathbb{T}|} \sum_{t \in \mathbb{T}} F_t(\mathbf{x}_i)$
\ENDFOR

\FOR{each training batch $\mathcal{B} \subset \mathcal{S}_\mathbf{x}$}
    \STATE Get soft labels $\mathbf{t}$ and student outputs $\mathbf{s} \gets S(\mathcal{B})$
    \STATE Compute batch-wise OT loss $\mathcal{L}_{\text{SD}} \gets \mathrm{W}(\mathbf{t}, \mathbf{s})$
    \STATE Compute per-sample CE and MSE loss:
    $
    \mathcal{L}_{\text{CE}} = \sum \mathcal{L}_{\text{CE}}(y_{\text{onehot}}, \mathbf{s}), \quad 
    \mathcal{L}_{\text{MSE}} = \sum \mathcal{L}_{\text{MSE}}(\mathbf{t}, \mathbf{s})
    $
    \STATE Total loss: 
    $
    \mathcal{L} = \kappa_1 \mathcal{L}_{\text{CE}} + \kappa_2 \mathcal{L}_{\text{MSE}} + \beta_2 \mathcal{L}_{\text{SD}}
    $
    \STATE Update student model $S$ using gradient descent
\ENDFOR
\RETURN $\mathcal{S}_\mathbf{x}$, $S$
\end{algorithmic}
\end{algorithm}
%\vspace{-1em}
\begin{algorithm}
\caption{Computation of OT-based Guidance for Image Latent Sampling}
\label{alg:ot_guidance}
\begin{algorithmic}[1]
\REQUIRE Previously sampled latents $\mathbf{M}_{n-1}^c$, current latent $\mathbf{z}_t^c$, a random batch of real class latents $\mathbf{Z}_\mathcal{T}^c$, regularization weight $\lambda_1$, iteration number $T$
\ENSURE Optimal transport distance as guidance value $\mathcal{G}_\mathrm{W}(\mathbf{z}_t^c)$

\STATE Concatenate latent: $\hat{\mathbf{M}}_n^c \gets [\mathbf{M}_{n-1}^c, \mathbf{z}_t^c]$
\STATE Compute cost matrix: $\mathbf{D}_{ij} \gets \|\hat{\mathbf{M}}_n^c(i) - \mathbf{Z}_\mathcal{T}^c(j)\|_p$
\STATE Initialize kernel matrix: $\mathbf{K} \gets \exp(-\mathbf{D} / \lambda_1)$

\STATE Set iteration counter $t \gets 0$
  \WHILE{$t < T$}
    \STATE Row normalization: $\mathbf{K} \gets \mathrm{diag}\left( \mathbf{K} \mathbf{1}_{n} \oslash (n \cdot \mathbf{1}_{|\mathbf{Z}_\mathcal{T}^c|}) \right)^{-1} \cdot \mathbf{K}$
    \STATE Column normalization: $\mathbf{K} \gets \mathbf{K} \cdot \mathrm{diag}\left( \mathbf{K}^\top \mathbf{1}_{|\mathbf{Z}_\mathcal{T}^c|} \oslash ({|\mathbf{Z}_\mathcal{T}^c|} \cdot \mathbf{1}_{n}) \right)^{-1}$
  \STATE Increment iteration counter $t \gets t + 1$
  \ENDWHILE

\STATE Final transport matrix: $\mathbf{P}^{\lambda_1} \gets \mathbf{K}$
\STATE Compute guidance: $\mathcal{G}_\mathrm{W}(\mathbf{z}_t^c) \gets \langle \mathbf{P}^{\lambda_1}, \mathbf{D} \rangle = \sum_{i,j} \mathbf{P}^{\lambda_1}_{ij} \mathbf{D}_{ij}$
\RETURN $\mathcal{G}_\mathrm{W}(\mathbf{z}_t^c)$
\end{algorithmic}
\end{algorithm}

% Class-wise optimal transport distance calculation for Label-Image Aligned Soft Label Relabeling is given in Algorithm~\ref{alg:classwise-ot}, which is used to compute the contraction factor $\alpha(\nu_{\text{distill}}^{(\mathrm{soft})})$. Full code implementations are available in the supplementary materials and the anonymous GitHub repository.

\begin{algorithm}
\color{black}{
  \caption{Computation of Batch-wise OT for Student Logit Matching}
  \label{alg:ot-ma}
  \begin{algorithmic}[1]
  \REQUIRE 
     Teacher output $\mathbf{t}$, Student output $\mathbf{s}$, \\
     Hyper-parameter $\lambda_2$, Maximum number of iterations $T$
  \ENSURE
     Sinkhorn loss $\mathcal{L}_{\text{SD}}$
     \STATE Apply softmax: $\mathbf{t} \gets \operatorname{Softmax}(\mathbf{t}),\quad \mathbf{s} \gets \operatorname{Softmax}(\mathbf{s})$
  \STATE  Compute distance matrix $
\mathbf{C}_{ij}(\mathbf{t},\mathbf{s})= \;\parallel \mathbf{t}(\mathbf{x}_i)-\mathbf{s}(\mathbf{x}_j)\parallel_p$
  \STATE  Compute kernel matrix $\mathbf{K} \gets \exp\left(-\frac{\mathbf{C}}{\lambda_2}\right)$
  \STATE Set iteration counter $t \gets 0$
  \WHILE{$t < T$}
    \STATE Row normalization: $\mathbf{K} \gets \mathbf{K} \oslash \left(\mathbf{K} \mathbf{1}_b \mathbf{1}_b^{\operatorname{T}}\right)$
    \STATE Column normalization: $\mathbf{K} \gets \mathbf{K} \oslash \left(\mathbf{1}_b\mathbf{1}_b^{\operatorname{T}} \mathbf{K}  \right)$
    \STATE Increment iteration counter $t \gets t + 1$
  \ENDWHILE
  \STATE Sinkhorn loss $\mathcal{L}_{\text{SD}} \gets \left<\mathbf{K},\mathbf{C}\right>=\sum_{i,j}{\mathbf{K}_{ij} \mathbf{C}_{ij}}$
  \RETURN $\mathcal{L}_{\text{SD}}$
  \end{algorithmic}
}
\end{algorithm}

\begin{algorithm}
\caption{Class-wise OT Distance in Label–Image Space for Contraction Factor 
$\alpha$ Calculation}
\label{alg:classwise-ot}
\begin{algorithmic}[1]
\REQUIRE Real latent sets $\mathbf{Z}_\mathcal{T} \in \mathbb{R}^{N_1 \times d}$,  distilled latent sets $\mathbf{Z}_\mathcal{S} \in \mathbb{R}^{N_2 \times d}$; \\
         One-hot labels $\mathbf{H}_\mathcal{T} \in \{0,1\}^{N_1 \times C}$, soft labels $\mathbf{S}_\mathcal{S} \in [0,1]^{N_2 \times C}$; \\
         Regularization parameter $\varepsilon > 0$, iterations $T$
\ENSURE Average classwise OT distance $\mathcal{L}_{\text{avg}}$

\STATE Compute pairwise cost matrix $\mathbf{C}_{ij} \gets \|\mathbf{Z}_\mathcal{T}(i) - \mathbf{Z}_\mathcal{S}(j)\|_p$
\STATE Initialize list of valid class distances $\mathrm{W}$

\FOR{each class $c = 1$ to $C$}
    \STATE $\tilde{\mathbf{a}} \gets \mathbf{H}_\mathcal{T}[:, c]$, \quad $\tilde{\mathbf{b}} \gets \mathbf{S}_\mathcal{S}[:, c]$
    \IF{$\sum_i \tilde{\mathbf{a}}_i = 0$ \OR $\sum_j \tilde{\mathbf{b}}_j = 0$}
        \STATE \textbf{continue}
    \ENDIF
    \STATE $\mathbf{a} \gets \tilde{\mathbf{a}} / \sum_i \tilde{\mathbf{a}}_i$ \STATE $\mathbf{b} \gets \tilde{\mathbf{b}} / \sum_j \tilde{\mathbf{b}}_j$
    \STATE $\mathbf{K} \gets \exp(-\mathbf{C} / \varepsilon)$
    \STATE Initialize $\mathbf{u} \gets \mathbf{1} / N_1$ 
    \STATE Initialize $\mathbf{v} \gets \mathbf{1} / N_2$
    \FOR{$t = 1$ to $T$}
        \STATE $\mathbf{u} \gets \mathbf{a} / (\mathbf{K} \cdot \mathbf{v} + \delta)$
        \STATE $\mathbf{v} \gets \mathbf{b} / (\mathbf{K}^\top \cdot \mathbf{u} + \delta)$
    \ENDFOR
    \STATE $\gamma \gets \mathrm{diag}(\mathbf{u}) \cdot \mathbf{K} \cdot \mathrm{diag}(\mathbf{v})$
    \STATE $\mathcal{L}_c \gets \sum_{i,j} \gamma_{ij} \mathbf{C}_{ij}$
    \STATE Append $\mathcal{L}_c$ to list $\mathcal{L}$
\ENDFOR

\STATE $\mathrm{W} \gets \frac{1}{|\mathcal{L}|} \sum_{c} \mathcal{L}_c$

\RETURN $\mathrm{W}$
\end{algorithmic}
\end{algorithm}

\section{Implementation Details}
\label{a5}
To ensure a fair and rigorous evaluation, we adopt the training protocols and experimental configurations established by IGD~\cite{igd} and EDC~\cite{edc}, maintaining full consistency in model architecture, optimization settings, and evaluation pipelines. Following Minimax~\cite{minimax} and IGD, we utilize a latent DiT model from Pytorch’s official repository and an open-source VAE model from
Stable Diffusion. DDIM~\cite{ddim} with 50 denoised steps is used as the vanilla sampling
method for generation. 
Also, all hyperparameters related to trajectory and diversity guidance are directly inherited from IGD, while the settings for student model logit matching follow those of EDC, with the exception of parameters introduced by our optimal transport (OT) framework. Notably, most of the OT-specific hyperparameters are set to fixed values across all datasets, and we observe that they require minimal tuning to achieve strong performance. This demonstrates the robustness of our method and its low sensitivity to OT parameter variations. Comprehensive hyperparameter configurations for all benchmark datasets including ImageNet-1K, ImageNette, ImageNet-100, and CIFAR-100 are detailed in Tables~\ref{tab:setting_imagenet_1k}, \ref{tab:setting_imagenet_nette}, \ref{tab:setting_imagenet_100}, and \ref{tab:setting_cifar}, respectively.

\begin{table}[htbp]
\centering
\caption{Hyperparameter setting on ImageNet-1K~\cite{imagenet}.}
\label{tab:setting_imagenet_1k}
\begin{tabular}{lcc}
\toprule
\textbf{Config} & \textbf{Value} & \textbf{Explanation} \\
\midrule
\multicolumn{3}{c}{Guided Diffusion Sampling}
\\
\midrule
$k$ & 5 & 
\makecell{$\rho_t = k \cdot \sqrt{1 - \alpha_t} \cdot$
$\frac{\left\| \epsilon_{\phi}(\mathbf{z}_t, t, c) \right\|}{\left\| \nabla_{\mathbf{z}_t} \mathcal{G}_I(\hat{\mathbf{z}}_0 \mid t) \right\|}$} \\
$\gamma_t$ & 120 & Weight for Diversity Guidance\\
$\beta_1$ & 1 & Weight for OT Sampling Guidance\\
$\lambda_1$ & 1000 & Entropy Regularization Weight \\ 
$T$ & 20 & Sinkhorn Iterations, Same for Logit Matching \\
\midrule
\multicolumn{3}{c}{ Soft Label Relabeling}
\\
\midrule
Epochs & 300, 500, 1000 & 300 for comparison with most baselines \\
Batch Size & 50 & Use 100 when IPC = 50 \\
$\mathbb{T}(\text{IPC}=10)$ &ResNet-18, ShuffleNet &NA  \\
$\mathbb{T}(\text{IPC}=50)$&\makecell{ResNet-18, MobileNet, \\EfficientNet, ShuffleNet} & NA\\
\midrule
\multicolumn{3}{c}{Student Model Logit Matching}
\\
\midrule
Optimizer & AdamW & NA \\
Learning Rate & 0.001 & Only use 1e-4 for Swin-Transformer  \\
EMA Rate & 0.99 & Control EMA-based Evaluation \\
$\kappa_1, \kappa_2$ & 1, 0.025& Inherit from EDC \\
$\beta_2$ & 0.1 & Weight for $\mathcal{L}_{\text{SD}}$\\
$\lambda_2$ & 0.1 & Entropy Regularization Weight \\
Scheduler & Smoothing LR Schedule & $\zeta=2$ \\
Augmentation & \makecell{RandomResizedCrop\\RandomHorizontalFlip} & NA \\
\bottomrule
\end{tabular}

\end{table}

\begin{table}[htbp]
\centering
\caption{Hyperparameter setting on ImageNette~\cite{nette}.}
\label{tab:setting_imagenet_nette}
\begin{tabular}{lcc}
\toprule
\textbf{Config} & \textbf{Value} & \textbf{Explanation} \\
\midrule
\multicolumn{3}{c}{Guided Diffusion Sampling}
\\
\midrule
$k$ & 5 & 
\makecell{$\rho_t = k \cdot \sqrt{1 - \alpha_t} \cdot$
$\frac{\left\| \epsilon_{\phi}(\mathbf{z}_t, t, c) \right\|}{\left\| \nabla_{\mathbf{z}_t} \mathcal{G}_I(\hat{\mathbf{z}}_0 \mid t) \right\|}$} \\
$\gamma_t$ & \makecell{50 when IPC=10 \\ 120 when IPC=50 or 100} & Weight for Diversity Guidance\\
$\beta_1$ & 1 & Weight for OT Sampling Guidance\\
$\lambda_1$ & \makecell{1000 when IPC=10 \\ 3000 when IPC=50 or 100} & Entropy Regularization Weight \\ 
$T$ & 20 & Sinkhorn Iterations, Same for Logit Matching \\
\midrule
\multicolumn{3}{c}{ Soft Label Relabeling}
\\
\midrule
Epochs & 1000 & Same for reproducing IGD \\
Batch Size & \makecell{50 when IPC=10 \\ 100 when IPC=50 or 100} & NA \\
 $\mathbb{T}(\text{IPC}=10)$ &ResNet-18, MobileNet &NA  \\
$\mathbb{T}(\text{IPC}=100)$&\makecell{ResNet-18, MobileNet, \\EfficientNet, ShuffleNet} & Same for IPC=50\\
\midrule
\multicolumn{3}{c}{Student Model Logit Matching}
\\
\midrule
Optimizer & AdamW & NA \\
Learning Rate & 0.001 & NA  \\
EMA Rate & 0.99 & Control EMA-based Evaluation \\
$\kappa_1, \kappa_2$ & 1, 0.025& Inherit from EDC \\
$\beta_2$ & 0.1 & Weight for $\mathcal{L}_{\text{SD}}$\\
$\lambda_2$ & 0.1 & Entropy Regularization Weight \\
Scheduler & Smoothing LR Schedule & $\zeta=2$ \\
Augmentation & \makecell{RandAugment\\RandomResizedCrop\\RandomHorizontalFlip} & NA \\
\bottomrule
\end{tabular}
\end{table}

\begin{table}[htbp]
\centering
\caption{Hyperparameter setting on ImageNet-100~\cite{idc}.}
\label{tab:setting_imagenet_100}
\begin{tabular}{lcc}
\toprule
\textbf{Config} & \textbf{Value} & \textbf{Explanation} \\
\midrule
\multicolumn{3}{c}{Guided Diffusion Sampling}
\\
\midrule
$k$ & 5 & 
\makecell{$\rho_t = k \cdot \sqrt{1 - \alpha_t} \cdot$
$\frac{\left\| \epsilon_{\phi}(\mathbf{z}_t, t, c) \right\|}{\left\| \nabla_{\mathbf{z}_t} \mathcal{G}_I(\hat{\mathbf{z}}_0 \mid t) \right\|}$} \\
$\gamma_t$ & 120 & Weight for Diversity Guidance\\
$\beta_1$ & 1 & Weight for OT Sampling Guidance\\
$\lambda_1$ & 1000 & Entropy Regularization Weight \\ 
$T$ & 20 & Sinkhorn Iterations, Same for Logit Matching \\
\midrule
\multicolumn{3}{c}{ Soft Label Relabeling}
\\
\midrule
Epochs & 300 & NA \\
Batch Size & 100 & NA \\
 $\mathbb{T}(\text{IPC}=10)$ &ResNet-18, ShuffleNet &NA  \\
$\mathbb{T}(\text{IPC}=100)$&\makecell{ResNet-18, MobileNet, \\EfficientNet, ShuffleNet} & Same for IPC=50\\
\midrule
\multicolumn{3}{c}{Student Model Logit Matching}
\\
\midrule
Optimizer & AdamW & NA \\
Learning Rate & 0.001 & NA  \\
EMA Rate & 0.99 & Control EMA-based Evaluation \\
$\kappa_1, \kappa_2$ & 1, 0.025& Inherit from EDC \\
$\beta_2$ & 0.1 & Weight for $\mathcal{L}_{\text{SD}}$\\
$\lambda_2$ & 0.1 & Entropy Regularization Weight \\
Scheduler & Smoothing LR Schedule & $\zeta=2$ \\
Augmentation & \makecell{RandomResizedCrop\\RandomHorizontalFlip} & NA \\
\bottomrule
\end{tabular}

\end{table}

\begin{table}[htbp]
\centering
\caption{Hyperparameter setting on CIFAR-100~\cite{cifar}.}
\label{tab:setting_cifar}
\begin{tabular}{lcc}
\toprule
\textbf{Config} & \textbf{Value} & \textbf{Explanation} \\
\midrule
\multicolumn{3}{c}{Guided Diffusion Sampling}
\\
\midrule
$k$ & 5 & 
\makecell{$\rho_t = k \cdot \sqrt{1 - \alpha_t} \cdot$
$\frac{\left\| \epsilon_{\phi}(\mathbf{z}_t, t, c) \right\|}{\left\| \nabla_{\mathbf{z}_t} \mathcal{G}_I(\hat{\mathbf{z}}_0 \mid t) \right\|}$} \\
$\gamma_t$ & 120 & Weight for Diversity Guidance\\
$\beta_1$ & 1 & Weight for OT Sampling Guidance\\
$\lambda_1$ & 1000 & Entropy Regularization Weight \\ 
$T$ & 20 & Sinkhorn Iterations, Same for Logit Matching \\
\midrule
\multicolumn{3}{c}{ Soft Label Relabeling}
\\
\midrule
Epochs & 1000 & NA \\
Batch Size & 50 & NA \\
$\mathbb{T}(\text{IPC}=10)$ &ResNet-18, ShuffleNet &NA  \\
$\mathbb{T}(\text{IPC}=100)$&\makecell{ResNet18, ConvNet, \\MobileNet, WRN, ShuffleNet} & Same for IPC=50\\
\midrule
\multicolumn{3}{c}{Student Model Logit Matching}
\\
\midrule
Optimizer & AdamW & NA \\
Learning Rate & 0.001 & NA  \\
EMA Rate & 0.99 & Control EMA-based Evaluation \\
$\kappa_1, \kappa_2$ & 1, 0.025& Inherit from EDC \\
$\beta_2$ & 0.1 & Weight for $\mathcal{L}_{\text{SD}}$\\
$\lambda_2$ & 0.1 & Entropy Regularization Weight \\
Scheduler & Smoothing LR Schedule & $\zeta=2$ \\
Augmentation & \makecell{RandAugment\\RandomResizedCrop\\RandomHorizontalFlip} & NA \\
\bottomrule
\end{tabular}

\end{table}

\section{Further Experimental Analyses}
\label{a610}
\subsection{More Sensitivity Analysis}
\label{a6}
\begin{figure}[h!]
    \centering
    \begin{minipage}{0.48\linewidth}
        \centering
        \includegraphics[width=\linewidth]{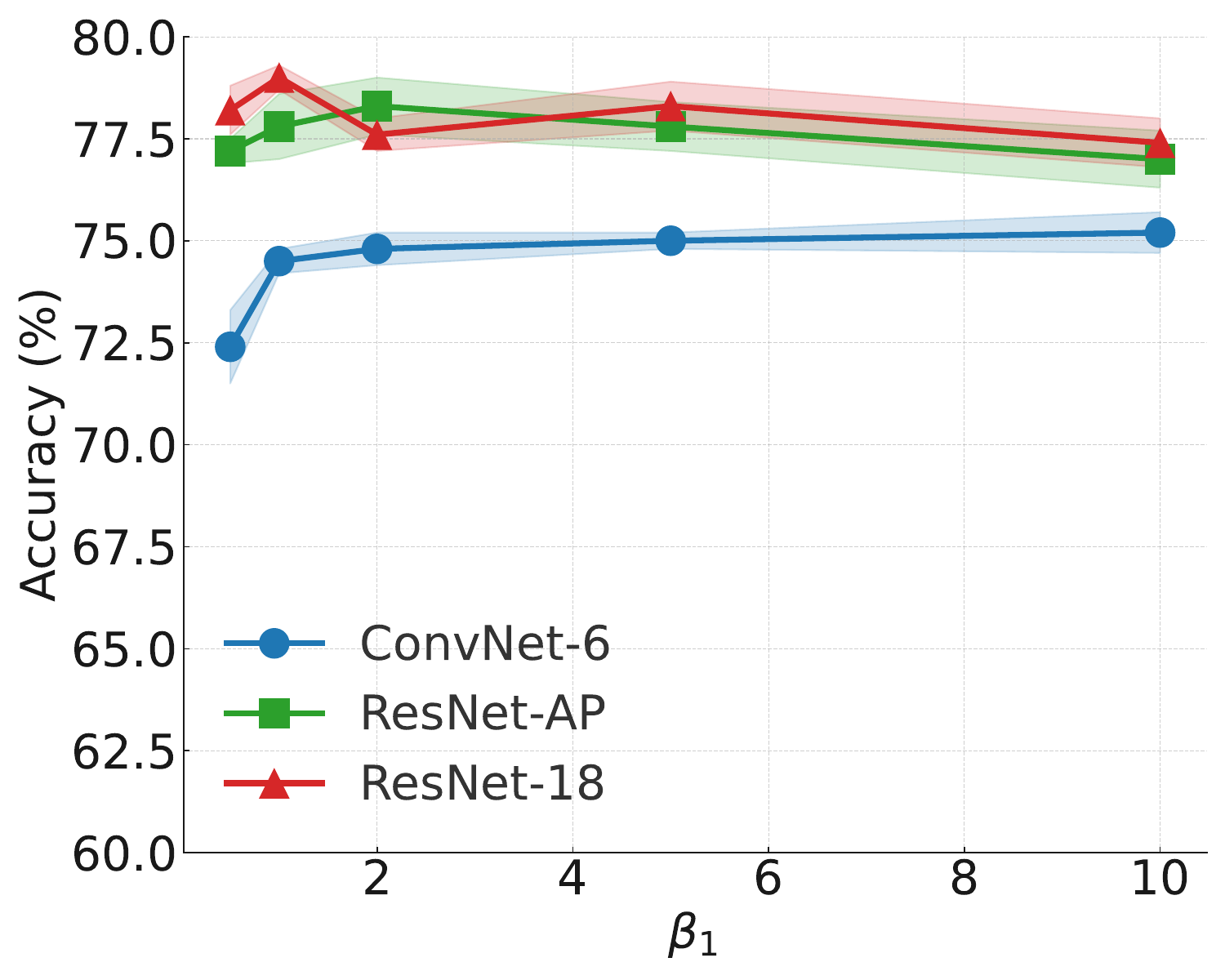}
        \caption{Effect of $\beta_1$ (OT sampling weight) on ImageNette~\cite{nette} (IPC=10).}
        \label{fig:beta1}
    \end{minipage}%
    \hfill
    \begin{minipage}{0.48\linewidth}
        \centering
        \includegraphics[width=\linewidth]{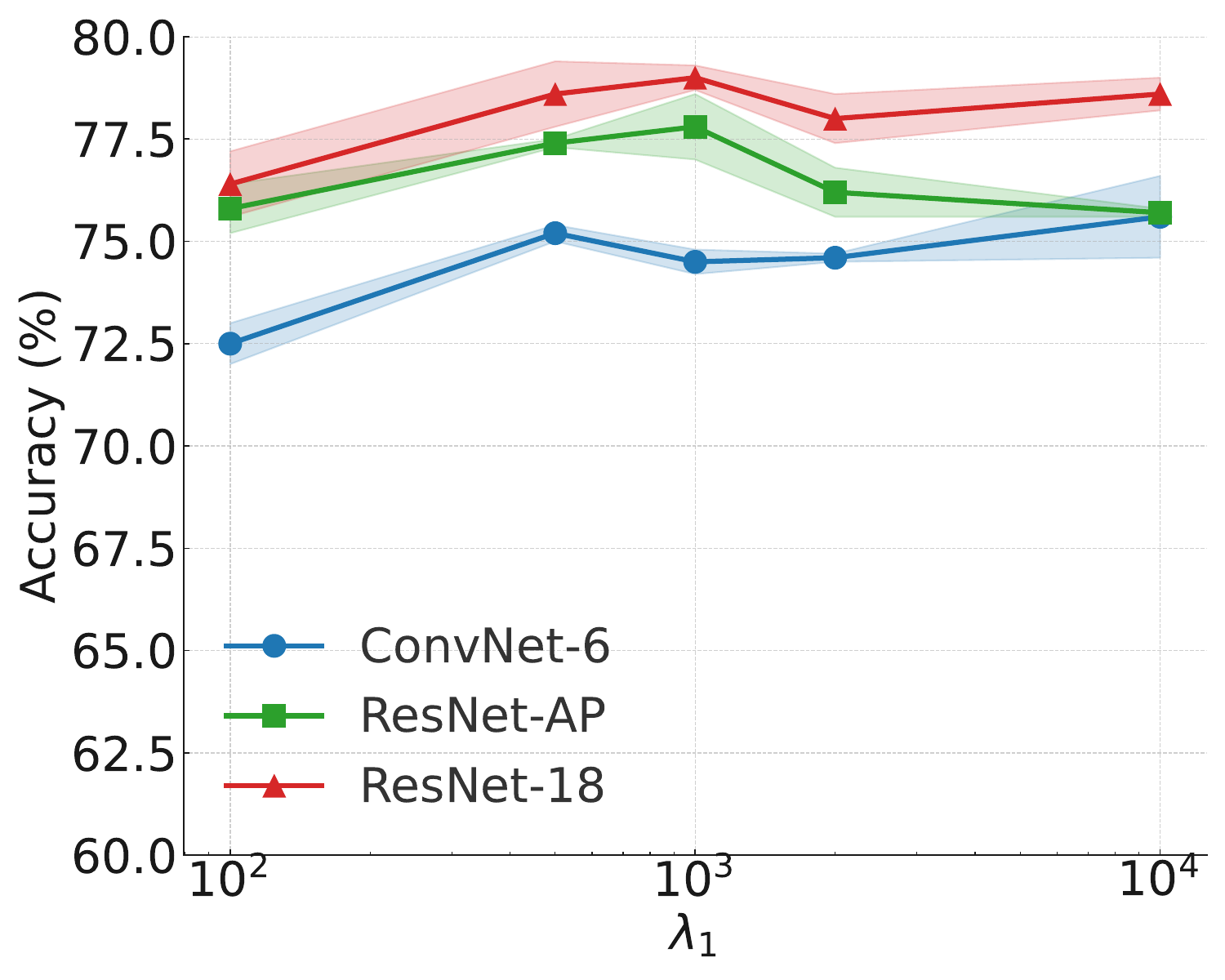}
        \caption{Effect of $\lambda_1$ (entropy regularization weight for sampling) on ImageNette~\cite{nette} (IPC=10).}
        \label{fig:lambda1}
    \end{minipage}
\end{figure}
\begin{figure}[h!]
    \centering
    \begin{minipage}{0.48\linewidth}
        \centering
        \includegraphics[width=\linewidth]{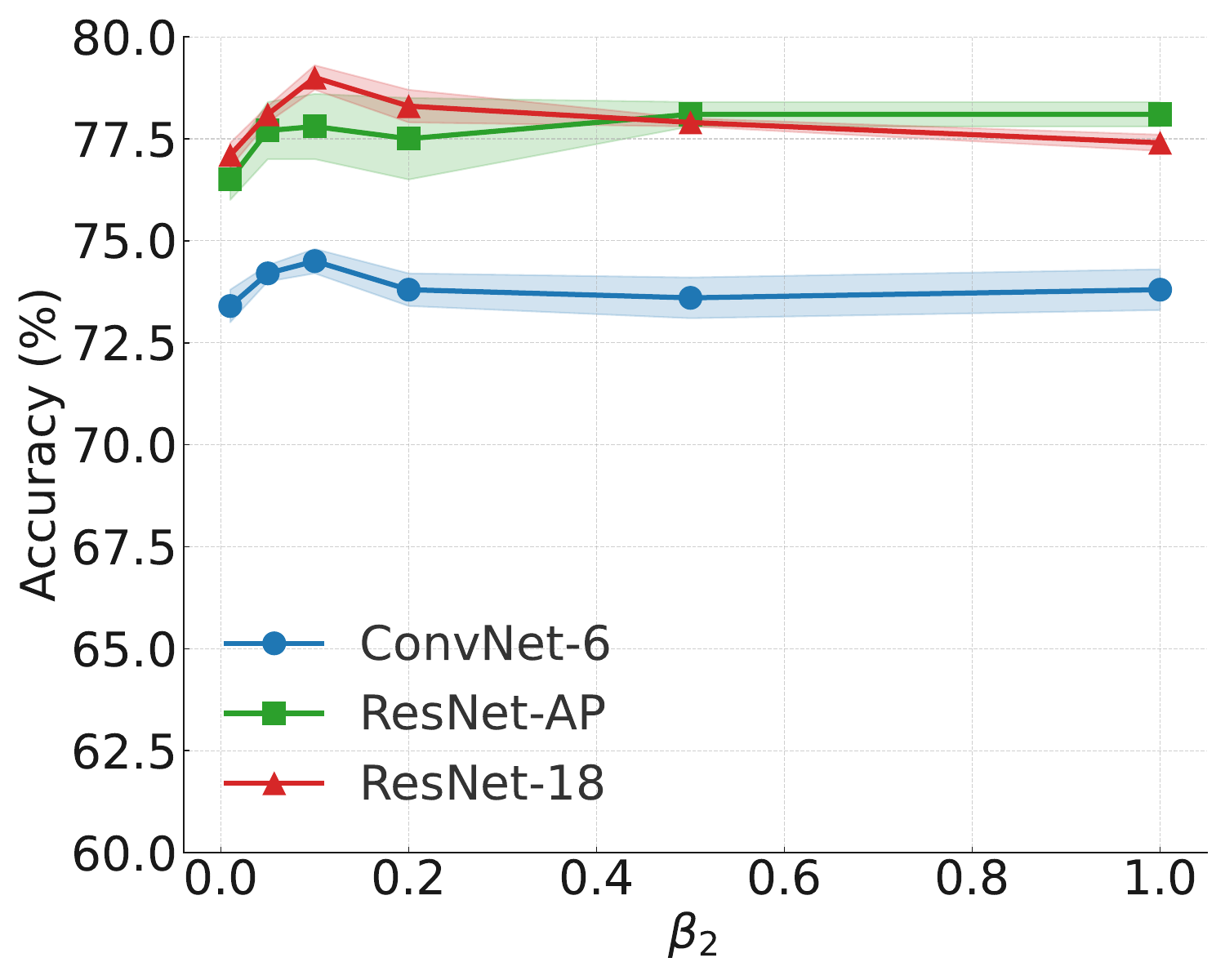}
        \caption{Effect of $\beta_2$ (OT matching weight) on ImageNette~\cite{nette} (IPC=10).}
        \label{fig:beta2}
    \end{minipage}%
    \hfill
    \begin{minipage}{0.48\linewidth}
        \centering
        \includegraphics[width=\linewidth]{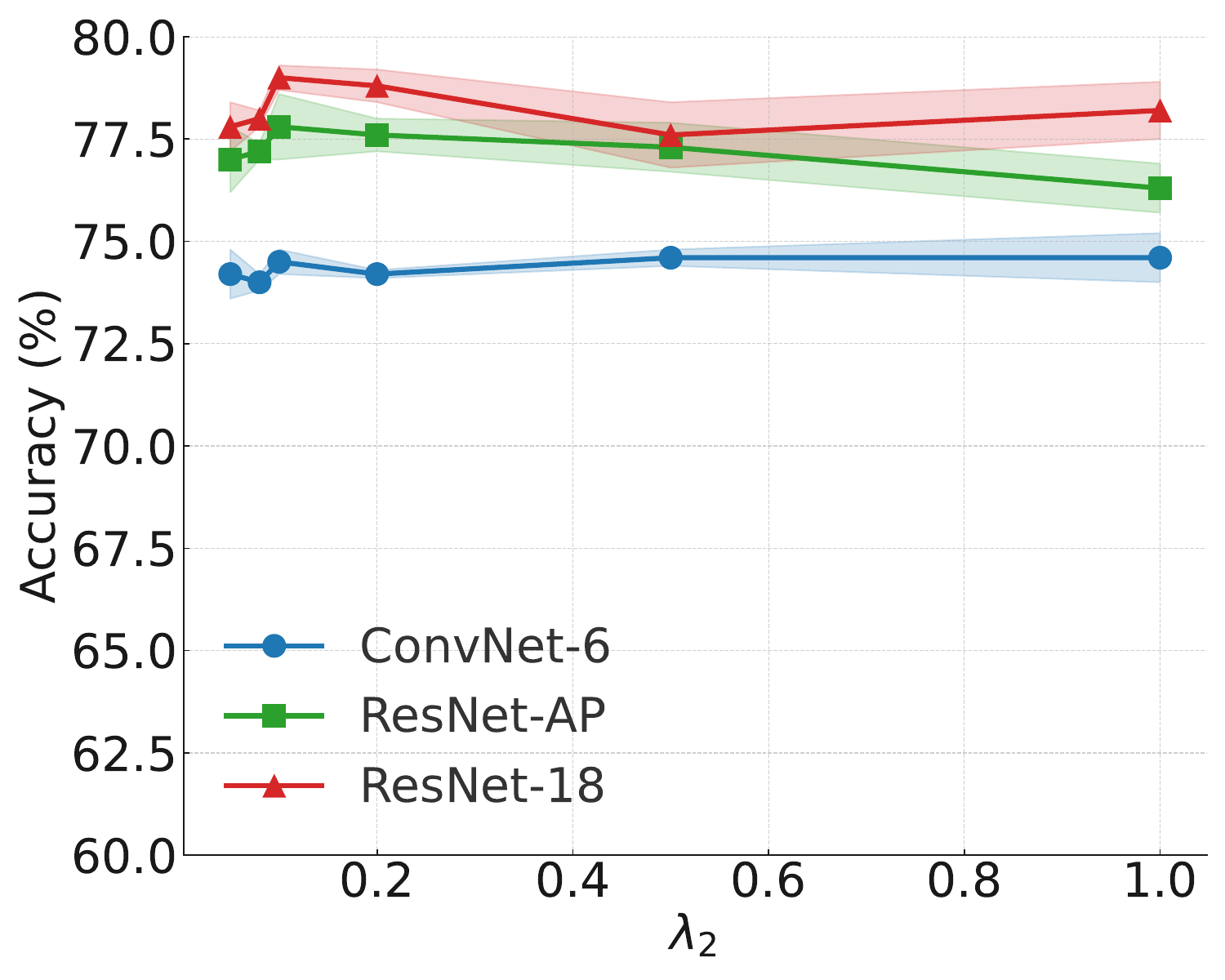}
        \caption{Effect of $\lambda_2$ (entropy regularization weight for logit matching) on ImageNette~\cite{nette} (IPC=10).}
        \label{fig:lambda2}
    \end{minipage}
\end{figure}

\begin{table}[tb]
\centering
\caption{Effect of $T$ on ImageNette~\cite{nette} (IPC=10).}
\label{tab:imagenette_T}
\small
\begin{tabular}{lccccc}
\toprule
$T$ & 5 & 10 & 20 & 50 & 100 \\
\midrule
ResNet-18 & 77.9 & 78.6 & 79.0 & 78.8 & 78.9 \\
ConvNet-6 & 73.8 & 74.3 & 74.5 & 74.7 & 74.6 \\
\bottomrule
\end{tabular}
\end{table}
In the main text, we presented a preliminary sensitivity analysis of key hyperparameters. To further evaluate their impact on performance, we conduct extensive ablation studies, with results summarized in Figures~\ref{fig:beta1},~\ref{fig:lambda1}, ~\ref{fig:beta2},~\ref{fig:lambda2} and Table~\ref{tab:imagenette_T}.
Overall, we observe that our method exhibits strong robustness to most hyperparameter settings: performance remains stable across a broad range of values.
We select the hyperparameters by considering trade-offs among different architectures.
Specifically, increasing the value of $\beta_1$ slightly improves the performance of ConvNet, but degrades that of ResNet-18. We therefore set $\beta_1=1$ to balance this trade-off. Similarly, increasing $\beta_2$ enhances performance on ResNet-AP, but negatively affects both ResNet-18 and ConvNet. Thus, we choose $\beta_2=0.1$ to achieve the best average performance.
In contrast, reducing either $\beta_1$ or $\beta_2$ consistently harms all model variants, which highlights the importance and effectiveness of our OT-based regularization terms.
Since the latent and feature spaces differ in scale, we apply separate scaling factors $\lambda_1$ and $\lambda_2$ to normalize their contributions. As shown in the figures, setting $\lambda_1=1000$ and $\lambda_2=0.1$ yields a favorable trade-off across architectures.
Taken together, our approach demonstrates two complementary aspects of robustness:  
(1) most OT-related hyperparameters exhibit consistent behavior across different scenarios, requiring little to no manual adjustment, and in practice we only select \(\lambda_1\) from the set \(\{1000,3000\}\) while keeping all other OT-related hyperparameters fixed (see Section \ref{a5} for default values); and  
(2) performance remains stable even when these parameters vary within reasonable ranges, eliminating the need of careful tuning.
% Taken together, our approach show two complementary aspects of robustness: 
% (1) most OT-related hyperparameters exhibit consistent behavior across different scenarios, requiring little to no manual adjustment, in fact, we only search $\lambda_1$ between 1000 and 3000 while keep all other OT-related hyperparameters fixed in all the tasks and IPCs (see Section~\ref{a5} for default values); 
% and (2) performance remains relatively stable even when these parameters vary within reasonable ranges.
% These properties make our method broadly adaptable to diverse settings, without the need for extensive hyperparameter tuning.
% Importantly, most OT-related hyperparameters do not require manual tuning across datasets or models (see Section~\ref{implementation-details} for default values). This robustness makes our method widely applicable to new settings, without extensive hyperparameter search or adaptation.

\subsection{Further Analysis of the Contraction Factor $\alpha$}
\label{a7}

We provide a formal characterization and empirical analysis of the contraction factor $\alpha$, which quantifies the degree to which soft labels reduce the discrepancy between the label and image distributions compared to hard labels. This factor plays a critical role in interpreting the effectiveness of soft supervision in dataset distillation. For efficient computation, we approximate the contraction factors using features in the latent space, enabling dimensionality reduction while preserving critical information.

\paragraph{Definition and Computation}
To compute $\alpha$, we compare the class-conditional optimal transport (OT) distances from the real dataset to two variants of the distilled dataset: one annotated with soft labels $\nu_{\text{distill}}^{(\mathrm{soft})}$ and one with hard labels $\nu_{\text{distill}}^{(\mathrm{hard})}$. The contraction factor is then defined as the relative improvement in transport distance under soft supervision: $
\alpha = \mathrm{W}(\mu_{\text{true}}, \nu_{\text{distill}}^{(\mathrm{soft})})/\mathrm{W}(\mu_{\text{true}}, \nu_{\text{distill}}^{(\mathrm{hard})}).
$

Let $\mathbf{Z}_\mathcal{T} \in \mathbb{R}^{N_1 \times d}$ and $\mathbf{Z}_\mathcal{S} \in \mathbb{R}^{N_2 \times d}$ denote latent embeddings extracted from real and distilled images, respectively. We construct the pairwise cost matrix using an $\ell_p$ norm:
\begin{equation}
    \mathbf{C}_{ij} = \|\mathbf{Z}_\mathcal{T}(i) - \mathbf{Z}_\mathcal{S}(j)\|_p, \quad \mathbf{C} \in \mathbb{R}^{N_1 \times N_2}.
\end{equation}
For each class $c \in \{1, \dots, C\}$, we extract the marginal label distributions over samples: $\tilde{\mathbf{a}}^{(c)} = \mathbf{H}_\mathcal{T}[:, c]$ for real hard labels, and $\tilde{\mathbf{b}}^{(c)} = \mathbf{S}_\mathcal{S}[:, c]$ for distilled soft labels, where $\mathbf{H}_\mathcal{T} \in \{0,1\}^{N_1 \times C}$ and $\mathbf{S}_\mathcal{S} \in [0,1]^{N_2 \times C}$. These are normalized into valid probability vectors:
\begin{equation}
    \mathbf{a}^{(c)} = \frac{\tilde{\mathbf{a}}^{(c)}}{\sum_i \tilde{a}_i^{(c)}}, \quad
    \mathbf{b}^{(c)} = \frac{\tilde{\mathbf{b}}^{(c)}}{\sum_j \tilde{b}_j^{(c)}}.
\end{equation}

We then perform entropic regularized OT using the Sinkhorn algorithm. The Gibbs kernel is defined as:
\begin{equation}
    \mathbf{K} = \exp\left(-\frac{\mathbf{C}}{\varepsilon}\right),
\end{equation}
where $\varepsilon$ controls regularization strength. The scaling vectors 
$\mathbf{u}$
and $\mathbf{v}$ are initialized uniformly as:
\begin{equation}
\mathbf{u}^{0} \gets \mathbf{1} / N_1, \quad \mathbf{v}^{0} \gets \mathbf{1} / N_2,
\end{equation}
and iteratively updated as:
\begin{align}
    \mathbf{u}^{t+1} &= \frac{\mathbf{a}^{(c)}}{\mathbf{K} \mathbf{v}^{t} + \delta}, \\
    \mathbf{v}^{t+1} &= \frac{\mathbf{b}^{(c)}}{\mathbf{K}^\top \mathbf{u}^{t+1} + \delta},
\end{align}
where $\delta$ ensures numerical stability. After $T$ iterations, the transport plan is:
\begin{equation}
    \gamma^{(c)} = \mathrm{diag}(\mathbf{u}) \cdot \mathbf{K} \cdot \mathrm{diag}(\mathbf{v}),
\end{equation}
and the classwise OT cost becomes:
\begin{equation}
    \mathcal{L}_c = \langle \gamma^{(c)}, \mathbf{C} \rangle = \sum_{i,j} \gamma^{(c)}_{ij} C_{ij}.
\end{equation}
Averaging over the valid class set $\mathcal{C}$ (i.e., classes with non-zero support in both distributions) yields:
\begin{equation}
    \mathrm{W}(\mu_{\text{true}}, \nu_{\text{distill}}^{(\mathrm{soft})}) = \frac{1}{|\mathcal{C}|} \sum_{c \in \mathcal{C}} \mathcal{L}_c.
\end{equation}
To compute the counterpart $\mathrm{W}(\mu_{\text{true}}, \nu_{\text{distill}}^{(\mathrm{hard})})$, we replace $\mathbf{S}_\mathcal{S}$ with its hard label projection $\mathbf{S}_\mathcal{S}^h \in \{0,1\}^{N_2 \times C}$ and repeat the same computation.

\paragraph{Empirical Insights}
\begin{table}[]
    \centering
    \small
    \caption{Effect of $\alpha$ on ImageNet-1K~\cite{imagenet} (IPC=10). \textbf{Config A:} ResNet-18. \textbf{Config B:} ResNet-18, MobileNet, EfficientNet, ShuffleNet. \textbf{Config C:} ResNet-18, MobileNet, AlexNet, ShuffleNet. \textbf{Config D:} ResNet-18, ShuffleNet.}
    \label{alphaa1}
    \begin{tabular}{c|cccc}
        \toprule
      Teachers  &Config A &Config B&Config C&Config D \\
        \midrule
       $\alpha$  &1.00  &0.99& 0.95&0.93\\
        Acc (ResNet-18)      &50.3 &52.3 &52.7&52.9 \\
        Acc (Swin) &47.2&47.8&49.2&50.2 \\
         \bottomrule
    \end{tabular}
\end{table}
\begin{table}[]
    \centering
    \small
    \caption{Effect of $\alpha$ on ImageNet-1K~\cite{imagenet} (IPC=50). \textbf{Config A:} ResNet-18. \textbf{Config B:} ResNet-18, MobileNet, EfficientNet, ShuffleNet. \textbf{Config C:} ResNet-18, MobileNet, AlexNet, ShuffleNet. \textbf{Config D:} ResNet-18, ShuffleNet.}
    \label{alphaa2}
    \begin{tabular}{c|cccc}
        \toprule
      Teachers  &Config A &Config B&Config C&Config D \\
        \midrule
       $\alpha$  &0.97  &0.16 &0.97 &1.00\\
        Acc (ResNet-18)  &62.3 &61.9 &60.8 &60.5 \\
        Acc (Swin) &65.5&68.2&65.5&65.3 \\
         \bottomrule
    \end{tabular}
\end{table}
We conduct several additional experiments, with results shown in Tables~\ref{alphaa1} and~\ref{alphaa2}.
Our empirical analysis provides several important observations regarding the role of the contraction factor $\alpha$ in guiding effective distillation. We first find that $\alpha$ is highly sensitive to the diversity and calibration quality of the teacher ensemble, and this sensitivity is modulated by the IPC (images per class) setting. In low-IPC regimes (e.g., IPC=10), using overly complex or inconsistent teacher predictions increases the optimal transport distance $\mathrm{W}(\mu_{\text{true}}, \nu_{\text{distill}}^{(\mathrm{soft})})$, leading to smaller $\alpha$ values and ultimately harming the generalization ability of the distilled dataset. Conversely, when IPC is sufficiently high (e.g., IPC=50), stronger and more expressive teacher distributions better capture the semantic structure of the real data, resulting in larger $\alpha$ values and improved alignment between real and synthetic distributions.

Second, we observe that deliberately reducing $\alpha$, thereby %contracting the teacher's distribution and 
explicitly minimizing the overall optimal transport distance, leads to significant improvements in downstream model performance. This effect is particularly evident under both settings, where configurations with smaller $\alpha$ (e.g., Config D) achieve better Top-1 accuracy across both the ResNet-18 and Swin Transformer. These results empirically confirm that shrinking the distributional gap through controlling $\alpha$ facilitates more efficient and effective knowledge transfer.
Third, in high-IPC regimes, when a single teacher is used (e.g., Config A), student models that share the same architecture as the teacher can fully exploit the teacher’s architectural biases, achieving strong performance. However, such tight alignment may limit generalization to unseen architectures. By appropriately contracting $\alpha$, we encourage the distilled dataset to encode more transferable, architecture-independent features, thereby improving the student's adaptability to diverse downstream architectures.
Overall, these findings validate $\alpha$ as a principled and tunable indicator of distillation quality, and highlight the importance of strategic contraction strategies tailored to both teacher complexity and downstream generalization targets.

\subsection{Comparison with Other Distance Measure}
\label{a8}
\paragraph{For guided diffusion sampling}
Unlike conventional metrics such as cosine similarity, KL divergence, or mean squared error, which rely on explicit instance-level alignment, optimal transport (OT) enables distribution-level alignment without enforcing one-to-one correspondences. This property makes OT particularly well-suited for dataset distillation scenarios, where the number of synthetic samples is significantly smaller than that of the original dataset, and direct pairing is often infeasible or suboptimal.
While Maximum Mean Discrepancy (MMD)-based measures have also been adopted for distribution alignment without requiring exact correspondences, they primarily focus on matching global distributional statistics and fail to capture fine-grained pairwise relations between individual instances in the real and synthetic distributions. In contrast, our OT-based formulation explicitly models such pairwise interactions and thus facilitates more accurate and semantically consistent guidance during the diffusion sampling process.
As shown in Table~\ref{mmd}, our method consistently outperforms MMD and MMD with reproducing kernel Hilbert spaces (RKHS) baselines on ImageNet-1K under both low-IPC and high-IPC settings. These results underscore the importance of modeling instance-level correspondences for effective guidance and highlight the superiority of OT in capturing the geometry of complex data distributions.

\begin{table}[]
    \centering
\caption{Comparison of our optimal transport-based distance with other measures for guided diffusion sampling on ImageNet-1K~\cite{imagenet}.}
    \label{mmd}
    \small
    \begin{tabular}{c|ccc}
    \toprule
    IPC &MMD &MMD (RKHS)&Ours \\
    \midrule
     10   &49.4&50.3 &  52.9\\
     50   &60.4&60.6 &  61.9\\
     \bottomrule
    \end{tabular}
    
\end{table}

\paragraph{For student model logit matching}
\begin{table}[]
    \centering
    \caption{Comparison between the sample-wise and batch-wise optimal transport distance for student model logit matching on ImageNet-1K~\cite{imagenet}. OOM: CUDA out of memory.}
    \label{batch}
    \small
    \begin{tabular}{c|cc|cc}
    \toprule
\multirow{2}{*}{Level} 
& \multicolumn{2}{c|}{ResNet-18} 
& \multicolumn{2}{c}{MobileNet-V2} \\
& IPC=10 & IPC=50 & IPC=10 & IPC=50 \\
    \midrule
     Sample-wise   & 50.6 &60.8&OOM&OOM\\
      Batch-wise   &52.9&61.9&51.0&61.0  \\
    \bottomrule
    \end{tabular}
\end{table}

\begin{table}[tbp]
\centering
\caption{Performance comparison of logit matching methods on ImageNette (IPC=10).}
\label{tab:imagenette_logit}
\small
\begin{tabular}{lccc}
\toprule
Network   & ConvNet-6 & ResNetAP-10 & ResNet-18 \\
\midrule
MMD       & 72.6      & 75.3        & 76.4 \\
KL        & 73.0      & 75.4        & 77.6 \\
OTM& \textbf{74.5} & \textbf{77.8} & \textbf{79.0} \\
\bottomrule
\end{tabular}
\end{table}
Table~\ref{batch} illustrates the consistent superiority of batch-wise OT distance over sample-wise OT distance. This result highlights that batch-wise Sinkhorn distance is more effective in transferring the distributional geometry captured by the distilled set from label-image space to the newly trained student models. 
The sample-wise logit matching approach treats each instance independently, failing to account for the global structure and correlations within a batch. In contrast, our batch-wise formulation preserves inter-sample relationships, enabling more faithful distributional alignment and resulting in more robust knowledge transfer. Moreover, when dealing with datasets containing a large number of classes (e.g., 1,000 classes in ImageNet-1K), the batch-wise approach substantially reduces memory consumption and avoids the CUDA out-of-memory issues frequently encountered by sample-wise matching, further enhancing its scalability.% to large-scale settings. 
Also, although KL and MMD serve as simpler divergences, they are inherently limited. KL divergence is applied per sample and ignores inter-sample relationships, while MMD matches only global statistics. In contrast, our OTM applies batch-wise OT alignment between student logits and soft labels, capturing the joint distributional structure of samples. This enables OT to faithfully preserve inter-sample geometry and match structural uncertainty in the soft labels, which KL and MMD overlook. As shown in Table~\ref{tab:imagenette_logit}, this leads to a clear performance gain:

\subsection{More Discussions on Runtime}
\label{a9}
While EDC~\cite{edc} reduces the runtime during the recovery phase compared to previous work, it does not fully optimize for multi-GPU parallelism across its various processes. Specifically, the pre-sampling and post-sampling phases during initialization do not benefit from multi-GPU parallelism, as parallelizing these steps does not result in substantial time reduction. Moreover, the recovery phase is inherently constrained by data loading and model-inversion methods, and beyond four GPUs, further increases in parallelism yield minimal improvements in runtime. In contrast, our approach is designed to optimize each class separately, with the sampling process dependent only on images sampled from the same class and the corresponding real images. As a result, our method scales more efficiently with the number of GPUs, with runtime decreasing nearly inversely proportional to the number of GPUs. Furthermore, when the need for high IPC arises, our method can be adapted to split high-IPC tasks into several lower-IPC ones for parallel processing, maintaining strong parallel efficiency and further enhancing its applicability in real-world scenarios.

\subsection{Data Coverage Analysis}
\label{a10}
To assess the representational fidelity of the distilled dataset, we adopt a coverage-based evaluation metric. Specifically, for each data point in the original dataset, we determine whether it has at least one nearest neighbor in the distance dataset within a predefined distance threshold. This metric reflects how well the surrogate data captures the underlying structure of the original distribution.
As shown in Table~\ref{coverage}, our method consistently achieves higher coverage compared to baseline methods across multiple thresholds. The improvements are observed over both the original DiT~\cite{dit} model and DiT-IGD~\cite{igd}, indicating that our approach provides better distributional alignment. Notably, the performance gap widens as the threshold increases, further validating the robustness of our distilled data in covering diverse modes of the original dataset.

\begin{table}[]
\caption{Distribution coverage comparison among different methods.}
    \label{coverage}
    \small
    \centering
    \begin{tabular}{c|ccc}
     \toprule
     Threshold & DiT~\cite{dit} &DiT-IGD~\cite{igd}&Ours \\
     \midrule
       10  &40.2&40.8&41.6 \\
       12  &54.6&56.3&57.5 \\
    \bottomrule
    \end{tabular}
\end{table}

\subsection{Comparison with DWA}
\label{dwacompare}
DWA~\cite{dwa} enhances diversity by adjusting the statistics of the squeezed network based on each generated sample. However, it still relies solely on global statistics, specifically the mean and variance associated with batch normalization (BN), and thus fails to capture the rich instance-level information and geometric distributional structures inherent in the real dataset. Visualizations from the DWA paper further illustrate that, while the directed weight adjustment improves the diversity of the distilled dataset, the distribution remains concentrated, failing to adequately cover the majority of the real data distribution. In Table~\ref{dwa}, we compare our method with DWA across multiple student models, and the results clearly demonstrate a significant performance advantage of our approach. This further emphasizes the importance of leveraging fine-grained instance-level information for achieving improved model performance and more faithful distributional alignment.

\begin{table}[tbp]
    \centering
    \small
    \caption{Comparison with DWA~\cite{dwa} on ImageNet-1K~\cite{imagenet}. }
    \label{dwa}
    \begin{tabular}{l|cc|cc|cc}
        \toprule
       \multirow{2}*{Method}  &\multicolumn{2}{c|}{ResNet-18} & \multicolumn{2}{c|}{MobileNet-V2} & \multicolumn{2}{c}{EfficientNet-B0} \\
        & IPC10 & IPC50 & IPC10 & IPC50 & IPC10 & IPC50 \\
        \midrule
        DWA~\cite{dwa} &37.9$\pm$0.2 & 55.2$\pm$0.2&29.1$\pm$0.3 & 51.6$\pm$0.5&37.4$\pm$0.5&56.3$\pm$0.4\\
        Ours & \textbf{52.9}$\pm$\textbf{0.1} & \textbf{61.9}$\pm$\textbf{0.5} &\textbf{51.0}$\pm$\textbf{0.6} & \textbf{61.0}$\pm$\textbf{0.4} & \textbf{56.7}$\pm$\textbf{0.2} & \textbf{64.4}$\pm$\textbf{0.1}  \\
        \bottomrule
    \end{tabular}
\end{table}

\begin{table}[tb]
\centering
\caption{Comparison between WMDD~\cite{wmdd} and our method on ImageNette~\cite{nette} and ImageNet-1K~\cite{imagenet} under different IPC settings.}
\begin{tabular}{lcccc}
\hline
Dataset & \multicolumn{2}{c|}{ImageNette} &  \multicolumn{2}{c}{ImageNet-1K}  \\
IPC     & 10         & 50         & 10          & 50          \\
\hline
WMDD    & 64.8$\pm$0.4 & 83.5$\pm$0.3 & 38.2$\pm$0.2 & 57.6$\pm$0.5 \\
Ours    & \textbf{79.0$\pm$0.3} & \textbf{89.3$\pm$0.3} & \textbf{52.9$\pm$0.1} & \textbf{61.9$\pm$0.5} \\
\hline
\end{tabular}

\label{tab:wmdd}
\end{table}

\subsection{Comparison with WMDD}
\label{wmddcompare}
Although both our method and WMDD~\cite{wmdd} utilize optimal transport (OT), they differ significantly in both methodology and motivation, leading to distinct formulations and implementations.

WMDD~\cite{wmdd} is a distribution-matching-based distillation method that applies OT in a single, offline step to compute a Wasserstein barycenter over the real data’s feature distribution. This barycenter is then used as a fixed target throughout training, where synthetic images are optimized to match it using a standard L2 loss in the feature space. In contrast, we introduce a fundamentally different generative paradigm where OT is not a static, one-off computation, but a dynamic guidance mechanism integrated throughout the entire data synthesis and training pipeline. Specifically, OT guides the sampling of synthetic images by aligning latent representations, regulates the soft label relabeling process by matching label complexity to the image distribution, and structures the training loss of the student model by aligning its logits to the relabeled targets.

The motivation behind WMDD is to replace the use of simple data summaries, such as the feature means often targeted by MMD-based methods, with a more geometrically meaningful summary, namely the Wasserstein barycenter, derived from the Wasserstein metric. In contrast, our method is driven by the need to address inherent limitations in generative distillation pipelines, which often fail to preserve the fine-grained geometry of the real data distribution—particularly intra-class variations and local modes. These aspects are explicitly addressed in our framework through a multi-stage, OT-guided design.

We compare the top-1 accuracy of our method with WMDD under different images-per-class (IPC) settings on both the ImageNette and ImageNet-1K datasets in Table~\ref{tab:wmdd}. 

\subsection{Robustness Evaluation}
\label{robustness}
To assess the robustness of student models trained with distilled datasets, we follow the evaluation protocol established in DD-RobustBench~\cite{ddrobust}, utilizing adversarial attacks implemented in the TorchAttacks library~\cite{kim2020torchattacks}. As shown in Tables~\ref{tab:attack_methods1} and~\ref{tab:attack_methods2}, we evaluate models trained on ImageNette~\cite{nette} under IPC=10 and IPC=50 settings, measuring both standard test accuracy and adversarial robustness against a variety of attack methods.

Our method consistently achieves higher clean accuracy and substantially improves robustness compared to MTT~\cite{mtt} across different perturbation budgets ($|\varepsilon|=4/255$ and $|\varepsilon|=8/255$). These improvements can be attributed to the distributional properties enforced by our optimal transport (OT)-based distillation framework. By minimizing the OT distance between the synthetic and real data distributions, our method preserves not only class-level statistics but also fine-grained, instance-level geometric structures. This leads to the learning of semantically faithful and smoother decision boundaries, which are inherently more resilient to adversarial perturbations. Moreover, our OT-guided diffusion sampling produces visually more coherent and perceptually realistic images compared to other types of approaches. The generated synthetic samples better preserve the semantic integrity and natural variability of the original data, providing stronger perceptual signals during model training. As a result, the student model benefits from a more robust feature space that aligns well with human perception, further enhancing adversarial robustness beyond purely decision-boundary-level effects.

In contrast, methods that primarily match global statistics or rely on heuristic trajectory guidance, such as MTT, often produce synthetic datasets lacking such structural fidelity, resulting in brittle decision boundaries that are more vulnerable to attacks.

From a theoretical perspective, prior works~\cite{pmlr-v97-zhang19j, moosavi2019robustness} have established a strong connection between adversarial robustness and the sharpness of decision boundaries: sharper, more irregular boundaries tend to amplify adversarial vulnerability, whereas flatter, smoother boundaries promote robustness. By aligning not only global distributions but also the local transportation cost between real and synthetic samples, OT encourages the distilled student model to form flatter and more coherent decision surfaces aligned with the real data geometry.

Moreover, from a loss landscape perspective, minimizing the OT distance guides optimization towards flatter minima, where small input perturbations induce minimal output changes. This connection is well supported by prior studies~\cite{tsuzuku2018lipschitz, zhang2019theoretically}, which show that flatter loss surfaces correlate strongly with improved adversarial robustness. Together, these empirical and theoretical insights demonstrate that preserving distributional geometry via optimal transport provides a principled and effective pathway for enhancing the adversarial robustness of models trained on distilled datasets.

\begin{table}[h]
\centering
\small
\caption{Performance comparison on DD-RobustBench~\cite{ddrobust} evaluated on ImageNette~\cite{nette}, under a perturbation budget of $|\varepsilon|=4/255$. Results for MTT~\cite{mtt} are directly copied from the DD-RobustBench benchmark.}
\label{tab:attack_methods1}
\begin{tabular}{c|cc|cc}
\toprule
\multirow{2}*{Attack Methods}&\multicolumn{2}{c|}{IPC=10}&\multicolumn{2}{c}{IPC=50} \\
 & MTT~\cite{mtt} &  Ours& MTT~\cite{mtt} &  Ours \\ \midrule
Clean Accuracy &66.4&\bf69.1&67.7&\bf84.6  \\
FGSM &10.8&\bf20.8&8.4&\bf24.0  \\
PGD &4.6&\bf9.2&2.6&\bf9.8 \\
CW &4.6&\bf12.0&1.4&\bf14.8  \\
VMI &5.4&\bf9.0&2.0&\bf11.2  \\
Jitter &12.2&\bf20.4&13.0&\bf23.8  \\
\bottomrule
\end{tabular}
\end{table}

\begin{table}[h]
\centering
\small
\caption{Performance comparison on DD-RobustBench~\cite{ddrobust} evaluated on ImageNette~\cite{nette}, under a perturbation budget of $|\varepsilon|=8/255$. Results for MTT~\cite{mtt} are directly copied from the DD-RobustBench benchmark.}
\label{tab:attack_methods2}
\begin{tabular}{c|cc|cc}
\toprule
\multirow{2}*{Attack Methods}&\multicolumn{2}{c|}{IPC=10}&\multicolumn{2}{c}{IPC=50} \\
 & MTT~\cite{mtt} &  Ours& MTT~\cite{mtt} &  Ours \\ \midrule
Clean Accuracy &66.4&\bf69.1&67.6&\bf84.6  \\
FGSM &0.8&\bf11.0&1.8&\bf14.8  \\
PGD &0.2&\bf2.8&1.2&\bf15.0 \\
CW &0.2&\bf9.6&0.2&\bf6.8  \\
VMI &0.2&\bf0.8&0.2&\bf2.0  \\
Jitter &11.4&\bf12.4&9.8&\bf14.6  \\
\bottomrule
\end{tabular}
\end{table}

\subsection{Evaluation on Low-IPC Settings}
\label{evallow}
We have conducted additional experiments on ImageNet-1K~\cite{imagenet} for the challenging settings of IPC=1, IPC=2, and IPC=5. The results of these new experiments are presented in the Table~\ref{tab:imagenet1k_ipc}.

Importantly, in the IPC=1 setting, since only one synthetic image is generated per class, the OTG process cannot leverage previously distilled samples for alignment. Instead, for each class, we compute the OT distance between its single synthetic candidate and the corresponding real images in the latent space to guide generation.

\begin{table}[ht]
\centering
\caption{Performance comparison of different methods on ImageNet-1K under small IPCs (1, 2, 5). Best results are in bold.}
\label{tab:imagenet1k_ipc}
\small
\begin{tabular}{lcccccccc}
\toprule
\multirow{2}{*}{IPC} & \multicolumn{8}{c}{Method} \\
\cmidrule(lr){2-9}
 & DM & FrePo & TESLA & SRe2L & RDED & EDC & DiT-IGD & Ours \\
\midrule
1 & 1.5 & 7.5 & 7.7 & 0.4 & 6.6 & 12.8 & 10.7 & \textbf{15.9} \\
2 & 1.7 & 9.7 & 10.5 & --  & 16.5 & 22.8 & 20.6 & \textbf{25.9} \\
5 & --  & --  & --   & --  & 23.8 & 39.5 & 38.6 & \textbf{45.7} \\
\bottomrule
\end{tabular}
\end{table}

\section{More Visualization Results}
\label{a11}
\subsection{T-SNE Results}
\begin{figure}[h]
    \centering
    \begin{minipage}{0.49\linewidth}
        \centering
        \includegraphics[width=\linewidth]{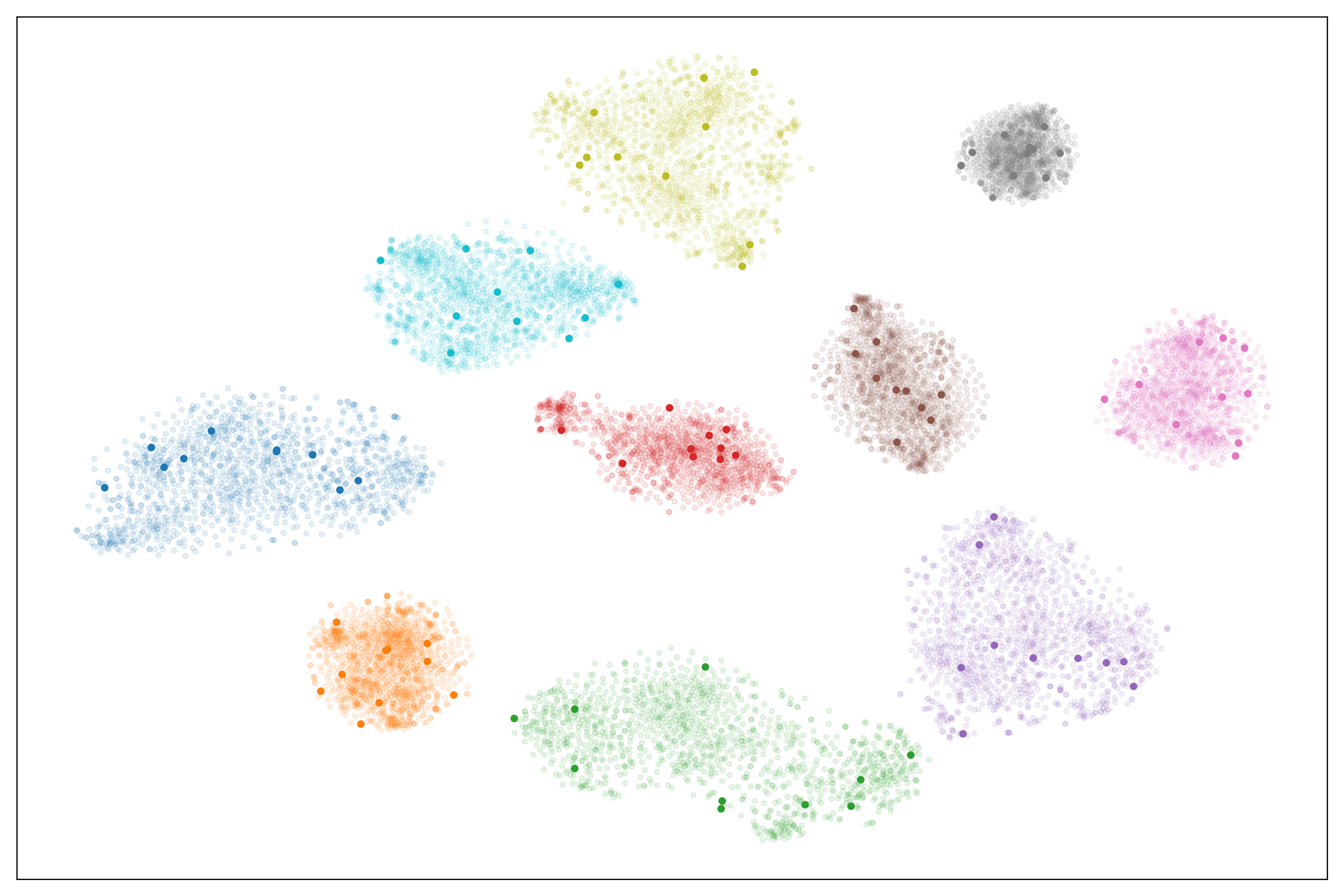}
        \subcaption{IGD}
    \end{minipage}%
    \hfill
    \begin{minipage}{0.49\linewidth}
        \centering
\includegraphics[width=\linewidth]{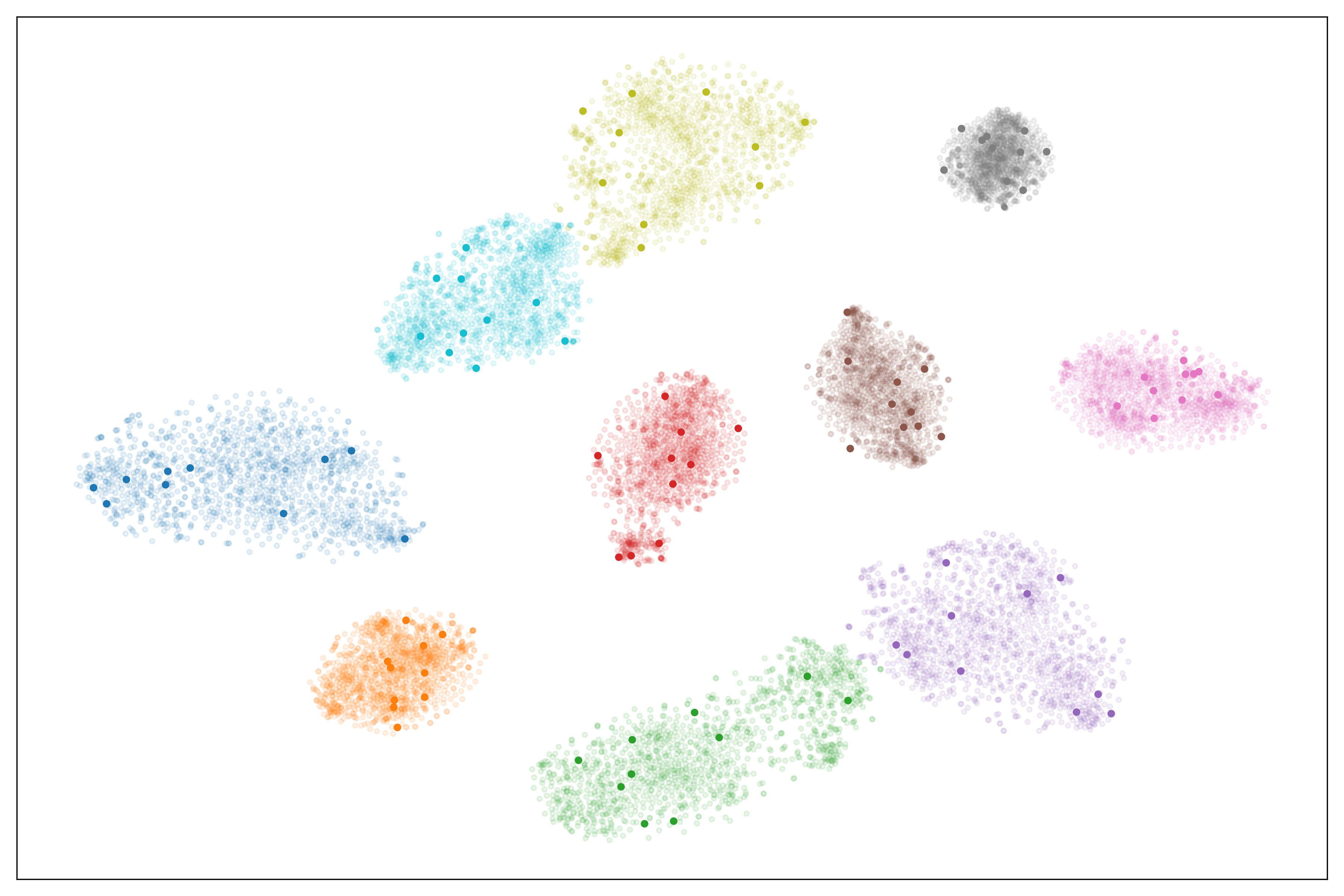}
        \subcaption{Ours}
    \end{minipage}
     \caption{Visualization study for sample distributions of distilled datasets (IPC=10) generated by IGD~\cite{igd} and Ours versus the original ImageNette~\cite{nette} dataset. The dark points represent the distilled set, while the light points represent the real (original) set. The diversity in IGD~\cite{igd}  is driven solely by random diversity guidance, lacking awareness of the real data distribution. As a result, it fails to cover critical regions such as such as the central region of the green class (Cassette player), the lower part of the blue class (Tench), and the middle-upper section of the purple class (Church). In contrast, our method incorporates both intra-class structures and fine-grained substructures of the real data, which allows it to effectively cover most subclass regions.}
    \label{fig:tsne-comparison}
\end{figure}

To assess the effectiveness of our OT-guided diffusion sampling, we present the t-SNE \cite{tsne} results in Figure~\ref{fig:tsne-comparison}. The diversity in IGD is driven solely by cosine-similarity based diversity guidance, without leveraging the distributional structure of the real dataset. This limitation leads to insufficient coverage of critical regions in the true data distribution, such as the central region of the green (Cassette player), the lower part of the blue (Tench), and the middle-upper section of the purple (Church) areas. Consequently, several important subclasses are absent from the distilled dataset, resulting in the new model failing to learn relevant intra-class variations and important subclass-specific information. In contrast, our approach iteratively computes the optimal transport distance between the real dataset and the distilled set, explicitly incorporating both intra-class structures and finer substructures of the real data. This enables our distilled dataset to capture a broader range of essential submodalities and regions, facilitating a more comprehensive transfer of information to the new model, and minimizing information loss. By employing the optimal transport distance as an additional supervision signal during the new model’s training, we ensure the effective transfer of this enriched information, leading to significant improvements in model performance.
 
\subsection{Distilled Images}
\begin{figure}
    \centering
    \includegraphics[width=\linewidth]{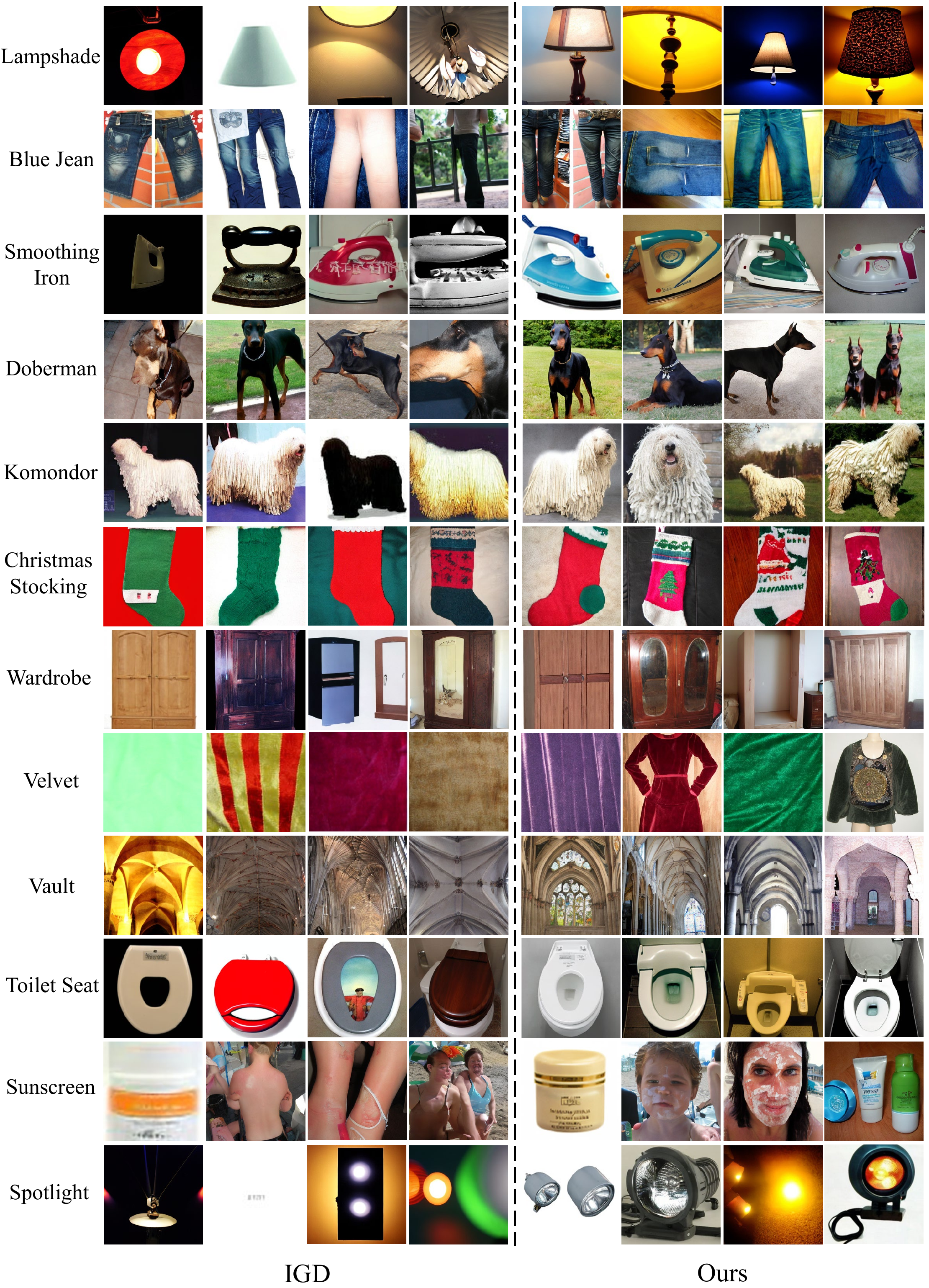}
    \caption{Additional visualization comparing the distilled datasets generated by IGD~\cite{igd} and our approach on ImageNet-1K~\cite{imagenet} (IPC=10).}
    \label{visualimage1}
\end{figure}

\begin{figure}
    \centering
    \includegraphics[width=1\linewidth]{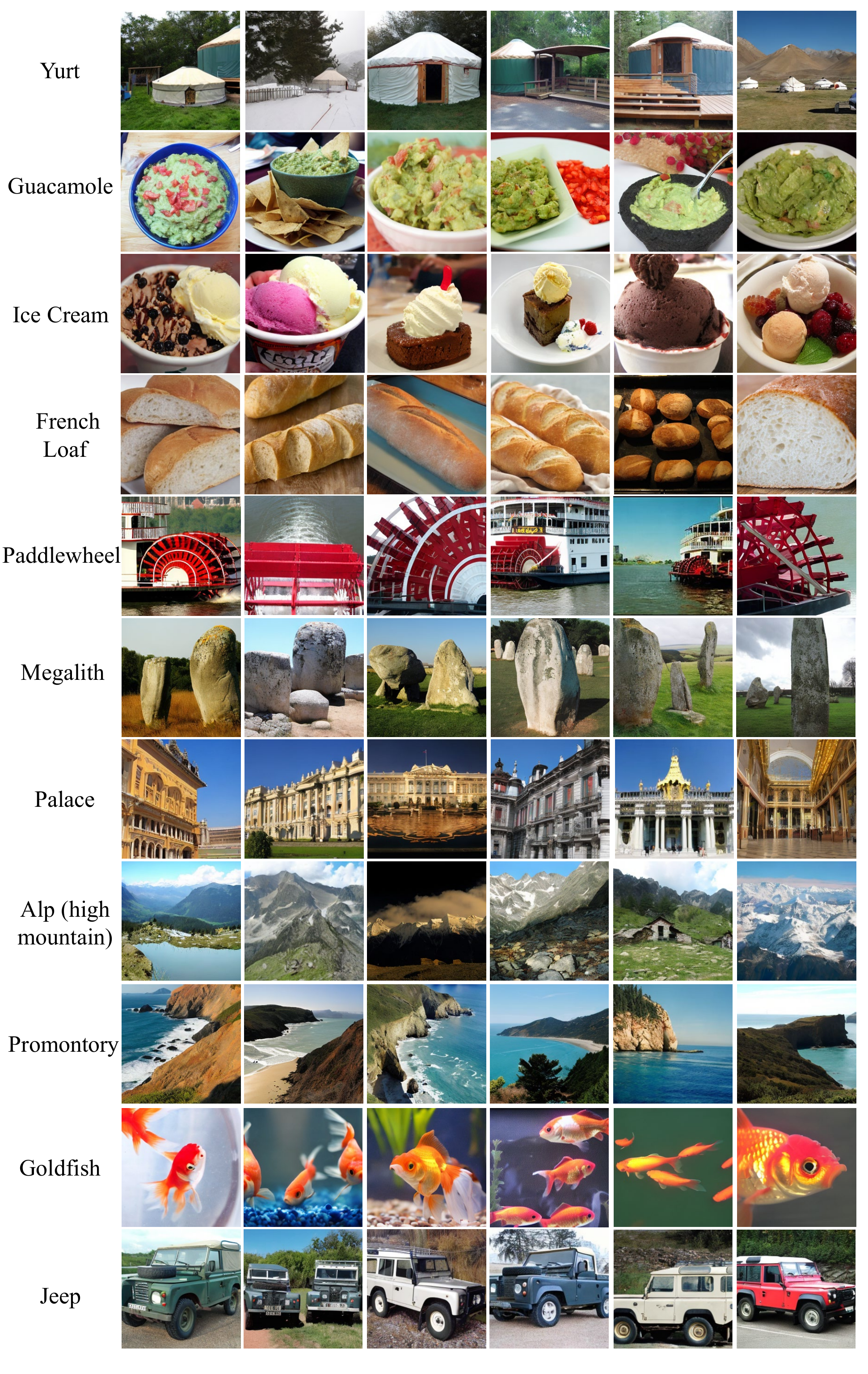}
    \caption{Additional visualizations for our distilled set on ImageNet-1K~\cite{imagenet} (IPC=10).}
    \label{visualimage2}
\end{figure}

Figures~\ref{visualimage1} and~\ref{visualimage2} provide additional visual comparisons between IGD~\cite{igd} and our method, as well as standalone visualizations of our distilled dataset.
Our method effectively captures the structural information of the real data distribution, resulting in high-fidelity samples with semantic diversity that faithfully reflects the underlying real-world distribution.

\section{Limitations}
\label{a12}
Our current framework inherits the trajectory influence guidance mechanism from IGD~\cite{igd}, which, while effective for improving the general and global alignment of sampled data, introduces substantial computational overhead. Specifically, the additional sampling steps required to maintain trajectory consistency significantly slow down the generation process compared to the vanilla DiT~\cite{dit}, which operates without such constraints. In future work, we aim to reformulate the guidance process to retain benefits while reducing the reliance on explicit trajectory tracking, thereby enabling faster and more scalable sampling.

\section{Broader Impact}
\label{a13}
Our work aims to reduce dataset size while maintaining performance, enabling model training with significantly lower computational and storage costs. This can lower the entry barrier for institutions with limited resources and promote environmentally sustainable AI development~\cite{cui2021heredity}. Moreover, our distilled datasets have the potential to facilitate efficient learning in federated and continual learning scenarios, thereby enhancing data privacy and supporting model adaptation across distributed systems. However, as with most data-driven approaches, there exists a risk that the distilled data may retain or amplify biases present in the original datasets. This could lead to unintended consequences, particularly in sensitive applications. Additionally, by accelerating the deployment of compact models, our method may inadvertently contribute to insufficiently audited systems being widely adopted. We emphasize the importance of responsible deployment, including bias auditing, fairness-aware design, and transparency, and encourage future work to explore these aspects more thoroughly.

\clearpage

\section*{NeurIPS Paper Checklist}

\begin{enumerate}

\item {\bf Claims}
    \item[] Question: Do the main claims made in the abstract and introduction accurately reflect the paper's contributions and scope?
    \item[] Answer: \answerYes{} % Replace by \answerYes{}, \answerNo{}, or \answerNA{}.
    \item[] Justification:  Our abstract and introduction accurately reflect the paper's contributions to generative large-scale dataset distillation.
    % \justificationTODO{}

\item {\bf Limitations}
    \item[] Question: Does the paper discuss the limitations of the work performed by the authors?
    \item[] Answer: \answerYes{}
    %\answerTODO{} % Replace by \answerYes{}, \answerNo{}, or \answerNA{}.
    \item[] Justification: Yes, we discuss this in Appendix~\ref{a12}.
   % \justificationTODO{}

\item {\bf Theory assumptions and proofs}
    \item[] Question: For each theoretical result, does the paper provide the full set of assumptions and a complete (and correct) proof?
    \item[] Answer: \answerNA{} % Replace by \answerYes{}, \answerNo{}, or \answerNA{}.
    \item[] Justification: This paper does not include theorems or lemmas.
    %\justificationTODO{}

    \item {\bf Experimental result reproducibility}
    \item[] Question: Does the paper fully disclose all the information needed to reproduce the main experimental results of the paper to the extent that it affects the main claims and/or conclusions of the paper (regardless of whether the code and data are provided or not)?
    \item[] Answer:\answerYes{} % Replace by \answerYes{}, \answerNo{}, or \answerNA{}.
    \item[] Justification: Yes, we provide sufficient implementation details in in Section \ref{setting} and Appendix \ref{a5}.
    %\justificationTODO{}

\item {\bf Open access to data and code}
    \item[] Question: Does the paper provide open access to the data and code, with sufficient instructions to faithfully reproduce the main experimental results, as described in supplemental material?
    \item[] Answer: \answerYes{} % Replace by \answerYes{}, \answerNo{}, or \answerNA{}.
    \item[] Justification: We provide code in the supplementary materials,
    and provide anonymous Github link.

\item {\bf Experimental setting/details}
    \item[] Question: Does the paper specify all the training and test details (e.g., data splits, hyperparameters, how they were chosen, type of optimizer, etc.) necessary to understand the results?
    \item[] Answer: \answerYes{} % Replace by \answerYes{}, \answerNo{}, or \answerNA{}.
    \item[] Justification: Experimental setting is described in Section \ref{setting} and Appendix \ref{a5}.

\item {\bf Experiment statistical significance}
    \item[] Question: Does the paper report error bars suitably and correctly defined or other appropriate information about the statistical significance of the experiments?
    \item[] Answer: \answerYes{} % Replace by \answerYes{}, \answerNo{}, or \answerNA{}.
    \item[] Justification: 
    We report error bars in Tables 1, 2, 3, 4, 5, 6.
    %\justificationTODO{}

\item {\bf Experiments compute resources}
    \item[] Question: For each experiment, does the paper provide sufficient information on the computer resources (type of compute workers, memory, time of execution) needed to reproduce the experiments?
    \item[] Answer: \answerYes{} % Replace by \answerYes{}, \answerNo{}, or \answerNA{}.
    \item[] Justification: 
    We provide the computer resouces in Section \ref{setting}.
    % \justificationTODO{}
    
\item {\bf Code of ethics}
    \item[] Question: Does the research conducted in the paper conform, in every respect, with the NeurIPS Code of Ethics \url{https://neurips.cc/public/EthicsGuidelines}?
    \item[] Answer: \answerYes{} % Replace by \answerYes{}, \answerNo{}, or \answerNA{}.
    \item[] Justification: 
    Our research conform with the NeurIPS Code of Ethics.
    %\justificationTODO{}

\item {\bf Broader impacts}
    \item[] Question: Does the paper discuss both potential positive societal impacts and negative societal impacts of the work performed?
    \item[] Answer: \answerYes{} % Replace by \answerYes{}, \answerNo{}, or \answerNA{}.
    \item[] Justification: We discuss the broader impact in Appendix \ref{a13}.
    %\justificationTODO{}
    
\item {\bf Safeguards}
    \item[] Question: Does the paper describe safeguards that have been put in place for responsible release of data or models that have a high risk for misuse (e.g., pretrained language models, image generators, or scraped datasets)?
    \item[] Answer: \answerNA{} % Replace by \answerYes{}, \answerNo{}, or \answerNA{}.
    \item[] Justification: The paper poses no such risks. %\justificationTODO{}

\item {\bf Licenses for existing assets}
    \item[] Question: Are the creators or original owners of assets (e.g., code, data, models), used in the paper, properly credited and are the license and terms of use explicitly mentioned and properly respected?
    \item[] Answer: \answerYes{} % Replace by \answerYes{}, \answerNo{}, or \answerNA{}.
    \item[] Justification: 
    All assets used in the paper, including datasets, are publicly available. Proper credits are given to the creators or original owners of these datasets where applicable. The licenses and terms of use for these datasets are explicitly mentioned and respected in accordance with their respective guidelines.
    %\justificationTODO{}

\item {\bf New assets}
    \item[] Question: Are new assets introduced in the paper well documented and is the documentation provided alongside the assets?
    \item[] Answer: \answerYes{} % Replace by \answerYes{}, \answerNo{}, or \answerNA{}.
    \item[] Justification: 
    We have attached our code and user instructions in the supplementary materials
    %\justificationTODO{}

\item {\bf Crowdsourcing and research with human subjects}
    \item[] Question: For crowdsourcing experiments and research with human subjects, does the paper include the full text of instructions given to participants and screenshots, if applicable, as well as details about compensation (if any)? 
    \item[] Answer: \answerNA{} % Replace by \answerYes{}, \answerNo{}, or \answerNA{}.
    \item[] Justification: 
    The paper does not involve crowdsourcing nor research with human subjects.
    %\justificationTODO{}

\item {\bf Institutional review board (IRB) approvals or equivalent for research with human subjects}
    \item[] Question: Does the paper describe potential risks incurred by study participants, whether such risks were disclosed to the subjects, and whether Institutional Review Board (IRB) approvals (or an equivalent approval/review based on the requirements of your country or institution) were obtained?
    \item[] Answer: \answerNA{} % Replace by \answerYes{}, \answerNo{}, or \answerNA{}.
    \item[] Justification: 
    The paper does not involve crowdsourcing nor research with human subjects.
    %\justificationTODO{}

\item {\bf Declaration of LLM usage}
    \item[] Question: Does the paper describe the usage of LLMs if it is an important, original, or non-standard component of the core methods in this research? Note that if the LLM is used only for writing, editing, or formatting purposes and does not impact the core methodology, scientific rigorousness, or originality of the research, declaration is not required.
    %this research? 
    \item[] Answer: \answerNA{} % Replace by \answerYes{}, \answerNo{}, or \answerNA{}.
    \item[] Justification: The core method development in this research does not involve LLMs as any important, original, or non-standard components.

\end{enumerate}

\end{document}